%% file: main.tex
\pdfoutput=1

\documentclass{article}

\usepackage{amsmath, amsthm, bbm, amssymb}
\usepackage{graphicx}
\usepackage{verbatim}
\usepackage{natbib}
\usepackage{caption}
\usepackage{subcaption}
\usepackage{fancyvrb}
\usepackage{enumerate}
\usepackage{relsize}
\usepackage{mathtools}
\usepackage{xcolor}
\usepackage{soul}
\usepackage{esvect}

\usepackage{hyperref}
\usepackage[margin=1.5in]{geometry}
\hypersetup{colorlinks,citecolor=blue,urlcolor=blue,linkcolor=blue}

\usepackage{stefan_tex}
\graphicspath{{./plots/}}

\theoremstyle{plain}
\newtheorem{prop}{Proposition}

\newtheorem{lemma}[prop]{Lemma}
\newtheorem{theorem}[prop]{Theorem}

\theoremstyle{definition}
\newtheorem{exam}{Example}
\newtheorem{definition}[exam]{Definition}
\newtheorem{assumption}{Assumption}

\theoremstyle{remark}

\newtheorem{step}{Step}

\DeclarePairedDelimiter\ceil{\lceil}{\rceil}
\DeclarePairedDelimiter\floor{\lfloor}{\rfloor}

\author{Xinkun Nie \\ \texttt{xinkun@stanford.edu}
\and
Stefan Wager \\ \texttt{swager@stanford.edu}}

\date{Draft version \ifcase\month\or
January\or February\or March\or April\or May\or June\or
July\or August\or September\or October\or November\or December\fi \ \number%
\year\ \  }

\title{Quasi-Oracle Estimation of \\
Heterogeneous Treatment Effects}

\begin{document}

\maketitle

\begin{abstract}
Flexible estimation of heterogeneous treatment effects lies at the heart of many statistical challenges, such as personalized medicine and optimal resource allocation. In this paper, we develop a general class of two-step algorithms for heterogeneous treatment effect estimation in observational studies. We first estimate marginal effects and treatment propensities in order to form an objective function that isolates the causal component of the signal. Then, we optimize this data-adaptive objective function. Our approach has several advantages over existing methods. From a practical perspective, our method is flexible and easy to use: In both steps, we can use any loss-minimization method, e.g., penalized regression, deep neural networks, or boosting; moreover, these methods can be fine-tuned by cross validation. Meanwhile, in the case of penalized kernel regression, we show that our method has a quasi-oracle property: Even if the pilot estimates for marginal effects and treatment propensities are not particularly accurate, we achieve the same error bounds as an oracle who has a priori knowledge of these two nuisance components. We implement variants of our approach based on penalized regression, kernel ridge regression, and boosting in a variety of simulation setups, and find promising performance relative to existing baselines.
\end{abstract}

\textbf{Keywords:} Boosting; Causal inference; Empirical risk minimization; Kernel regression; Penalized regression.
\section{Introduction}

The problem of heterogeneous treatment effect estimation in observational studies arises
in a wide variety application areas \citep{athey2017beyond}, ranging from personalized medicine \citep{obermeyer2016predicting}
to offline evaluation of bandits \citep*{dudik2011doubly}, and is also a key component of several proposals
for learning decision rules \citep{athey2017efficient,hirano2009asymptotics}.
There has been considerable interest in developing
flexible and performant methods for heterogeneous treatment effect estimation. Some notable
recent advances include proposals based on the
lasso \citep{imai2013estimating},
recursive partitioning \citep*{athey2015machine,su2009subgroup},
BART \citep*{hahn2020bayesian,hill2011bayesian},
random forests \citep*{wager2017estimation},
boosting \citep{powers2018some},
neural networks \citep*{shalit2017estimating},
etc., as well as combinations thereof \citep*{kunzel2019metalearners};
see \citet{dorie2019automated} for a recent survey and comparisons.

However, although this line of work has led to many promising methods, the literature has not yet settled on
a comprehensive answer as to how machine learning methods should be adapted for treatment effect estimation in observational studies.
The process of developing causal variants of machine learning methods is in practice a labor intensive process, effectively requiring the involvement of specialized researchers.
Moreover, with some exceptions, the above methods are mostly justified via numerical experiments, and come with
no formal convergence guarantees or error bounds proving that the methods in fact succeed in isolating
causal effects better than a simple non-parametric regression-based approach would.

In this paper, we discuss a new approach to estimating heterogeneous treatment effects that addresses
both of these concerns. Our framework allows for fully automatic specification of heterogeneous treatment
effect estimators in terms of arbitrary loss minimization procedures. Moreover, we show how the resulting
methods can achieve comparable error bounds to oracle methods that know everything about
the data-generating distribution except the treatment effects.
Conceptually, our approach fits into a research program---outlined by
\citet{van2003cross} and later developed by
\citet*{chernozhukov2017double},
\citet{luedtke2016super}
and references therein---whereby
we pair ideas on doubly robust estimation with oracle inequalities and cross-validation
to develop loss functions that can be used for principled statistical estimation using generic machine
learning tools.

\section{A Loss Function for Treatment Effect Estimation}

We formalize our problem in terms of the potential outcomes framework \citep{neyman1923applications,rubin1974estimating}.
The analyst has access to $n$ independent and identically distributed examples $(X_i, \, Y_i, \, W_i)$,
$i = 1, \, ..., \, n$, where $X_i \in \xx$ denotes per-person features, $Y_i \in \RR$ is the observed outcome, and
$W_i \in \cb{0, \, 1}$ is the treatment assignment. We posit the existence of potential outcomes
$\cb{Y_i(0), \, Y_i(1)}$ corresponding to the outcome we would have observed given the treatment assignment
$W_i = 0$ or $1$ respectively, such that $Y_i = Y_i(W_i)$, and seek to estimate the conditional average treatment
effect (CATE) function 
$\tau^*(x) = E\{Y(1) - Y(0) \cond X = x\}.$
In order to identify $\tau^*(x)$, we assume unconfoundedness,
i.e., the treatment assignment is randomized
once we control for the features $X_i$ \citep{rosenbaum1983central}. 

\begin{assumption}
	\label{assu:unconf}
	The treatment assignment $W_i$ is unconfounded, $\cb{Y_i(0), \, Y_i(1)} \indep W_i \cond X_i$.
\end{assumption}

We write the treatment propensity as \smash{$e^*(x) = \PP{W = 1 \cond X = x}$}
and the conditional response surfaces as \smash{$\mu^*_{(w)}(x) = E\{Y(w) \cond X = x\}$} for $w\in \{0,1\}$;
throughout this paper, we use $*$-superscripts to denote unknown population quantities.
Then, under unconfoundedness,
\begin{equation*}
E\{\varepsilon_i(W_i) \mid X_i, \, W_i\} = 0,  \text{ where } \varepsilon_i(w) := Y_i(w) - \{\mu_{(0)}^*(X_i) + w \tau^*(X_i)\}.
\end{equation*}
Given this setup, it is helpful to re-write the CATE function $\tau^*(x)$ in terms of the conditional
mean outcome \smash{$m^*(x) = \EE{Y \cond X = x} = \mu_{(0)}^*(X_i) + e^*(X_i) \tau^*(X_i)$} as follows, with the shorthand $\varepsilon_i := \varepsilon_i(W_i)$,
\begin{equation}
\label{eq:robinson}
Y_i - m^*(X_i) = \{W_i - e^*(X_i)\} \, \tau^*\p{X_i} + \varepsilon_i.
\end{equation}
This decomposition was originally used by \citet{robinson1988root} to estimate parametric components
in partially linear models, and has received considerable attention
in recent years. \citet*{athey2018generalized} rely on it to grow a causal forest that is robust to confounding,
\citet{robins2004optimal} builds on it in developing $G$-estimation for sequential trials,
and \citet{chernozhukov2017double} present it as a leading example on how machine learning methods can be put
to good use in estimating nuisance components for semiparametric inference.
All these results, however, consider estimating parametric models for $\tau(\cdot)$ or, 
in the case of \citet{athey2018generalized}, local parametric modeling.

The goal of this paper is to study how we can use the Robinson's transfomation \eqref{eq:robinson} for flexible
treatment effect estimation that builds on modern machine learning approaches such as boosting
or deep learning. Our main result is that we can use this representation to construct a loss function that
captures heterogeneous treatment effects, and that we can then accurately estimate treatment
effects---both in terms of empirical performance and asymptotic guarantees---by
finding regularized minimizers of this loss function.

As motivation for our approach, note that \eqref{eq:robinson} can equivalently be expressed as \citep{robins2004optimal}
\begin{equation}
\label{eq:oracle_loss}
\tau^*(\cdot) = \argmin_{\tau} \cb{\EE{\Bigg[\{Y_i - m^*(X_i)\} - \{W_i - e^*(X_i)\} \, \tau(X_i) \Bigg]^2}},
\end{equation}
and so an oracle who knew both the functions $m^*(x)$ and $e^*(x)$
a priori could estimate the heterogeneous treatment effect function $\tau^*(\cdot)$ by empirical loss minimization,
\begin{equation}
\label{eq:oracle}
\ttau(\cdot) = \argmin_{\tau} \Bigg(\frac{1}{n} \sum_{i = 1}^n \Bigg[\{Y_i - m^*(X_i)\} - \{W_i - e^*(X_i)\} \, \tau(X_i) \Bigg]^2 + \Lambda_n\{\tau(\cdot)\}\Bigg),
\end{equation}
where the term $\Lambda_n\p{\tau(\cdot)}$ is interpreted as a regularizer on the complexity of the $\tau(\cdot)$ function.
This regularization could be explicit as in penalized regression, or implicit,
e.g., as provided by a carefully designed deep neural network.
The difficulty, however, is that in practice we never know the weighted main effect function $m^*(x)$ and usually don't know the
treatment propensities $e^*(x)$ either, and so the estimator \eqref{eq:oracle} is not feasible.

Given these preliminaries, we here study the following class of two-step estimators using cross-fitting \citep{chernozhukov2017double, schick1986asymptotically} motivated by the above oracle procedure:

\begin{step}
	
	Divide up the data into $Q$ (typically set to 5 or 10) evenly sized folds. Let $q(\cdot)$ be a mapping from the $i=1, \ldots, n$ sample indices to $Q$ evenly sized data folds, and fit \smash{$\hat{m}$} and \smash{$\he$} with cross-fitting over the $Q$ folds  via methods
	tuned for optimal predictive accuracy, then
\end{step}
\begin{step}
	Estimate treatment effects via a plug-in version of \eqref{eq:oracle}, where \smash{$\he^{(-q(i))}(X_i)$}, etc., denote
	predictions made  without using the data fold the $i$-th training
	example belongs to,
	\begin{equation}
	\label{eq:main}
	\begin{split}
	&\htau(\cdot) = \argmin_{\tau} \Bigg[\hL_n\{\tau(\cdot)\} + \Lambda_n\{\tau(\cdot)\}\Bigg], \\
	&\hL_n\{\tau(\cdot)\} = \frac{1}{n} \sum_{i = 1}^n \Bigg[\{Y_i - \hatm^{(-q(i))}(X_i)\} - \{W_i - \he^{(-q(i))}(X_i)\} \, \tau(X_i)\Bigg]^2.
	\end{split}
	\end{equation}
\end{step}
In other words, the first step learns an approximation for the oracle objective, and the second step optimizes
it. We refer to this approach as the $R$-learner in recognition of the work of \citet{robinson1988root} and
to emphasize the role of residualization. We will also refer to the squared loss \smash{$\hL_n\{\tau(\cdot)\}$}
as the $R$-loss.

This paper makes the following contributions.
First, we implement variants of our method based on penalized regression, kernel ridge regression, and boosting. In each
case, we find that the $R$-learner exhibits promising performance relative to existing proposals.
Second, we prove that---in the case of penalized kernel regression---error bounds for
the feasible estimator for \smash{$\htau(\cdot)$} asymptotically match the best available bounds
for the oracle method \smash{$\ttau(\cdot)$}.
The main point here is that, heuristically, the rate of convergence of \smash{$\htau(\cdot)$} depends
only on the functional complexity of $\tau^*(\cdot)$, and not on the functional complexity of $m^*(\cdot)$ and $e^*(\cdot)$. More formally, provided we estimate $m^*(\cdot)$ and
$e^*(\cdot)$ at $o(n^{-1/4})$ rates in root-mean squared error, we show that we can achieve considerably faster rates
of convergence for \smash{$\htau(\cdot)$}---and these rates only depend on the complexity of
$\tau^*(\cdot)$. We note that the oracle version \eqref{eq:oracle_loss} of our loss function
is a member of a class of loss functions for heterogeneous treatment effect estimation considered in
\citet{luedtke2016super}, and that results in that paper immediately imply large-sample consistency of the minimizer of this oracle loss.
Our contribution is the result on rates---specifically, that the estimation error in nuisance components does not
affect our excess loss bounds for  \smash{$\htau(\cdot)$}.

The $R$-learning approach has several practical advantages over existing, more ad hoc proposals.
Any good heterogeneous treatment effect estimator needs to achieve two goals:
First, it needs to eliminate spurious effects by controlling for correlations
between $e^*(X)$ and $m^*(X)$, and then it needs to accurately express $\tau^*(\cdot)$. Most existing machine learning
approaches to treatment effect estimation seek to provide an algorithm that accomplishes both tasks at once
\citep[see, e.g.,][]{powers2018some,shalit2017estimating,wager2017estimation}. In contrast, the $R$-learner
cleanly separates these two tasks: We eliminate spurious correlations via the structure of the loss function
\smash{$\hL_n$}, while we can induce a representation for \smash{$\htau(\cdot)$} by choosing the
method by which we optimize \eqref{eq:main}.

This separation of tasks allows for considerable algorithmic flexibility:
Optimizing \eqref{eq:main} is an empirical minimization problem, and so can be efficiently solved
via off-the-shelf software such as
\texttt{glmnet} for high-dimensional regression \citep*{friedman2010regularization},
\texttt{XGboost} for boosting \citep{chen2016xgboost}, or
\texttt{TensorFlow} for deep learning \citep{tensorflow}.
Furthermore, we can tune any of these methods by cross validating on the loss
\smash{$\hL_n$}, which avoids the use of more sophisticated model-assisted
cross-validation procedures as developed in \citet{athey2015machine} or \citet{powers2018some}.
Relatedly, the machine learning method used to optimize \eqref{eq:main} only needs
to find a generalizable minimizer of \smash{$\hL_n$} rather than to also control
for spurious correlations, and thus we can confidently use black-box methods without
auditing their internal state to check that they properly control for confounding. Instead, we only need
to verify that they in fact find good minimizers of \smash{$\hL_n$} on holdout data.

\section{Related Work}
\label{sec:relworks}

Under unconfoundedness (Assumption \ref{assu:unconf}), the CATE function
can be written as 
$\tau^*(x) = \mu_{(1)}^*(x) - \mu_{(0)}^*(x)$, with $\mu_{(w)}^*(x) = \EE{Y \cond X = x, \, W = w}$.
As a consequence of this representation, it may be tempting to first estimate
\smash{$\hmu_{(w)}(x)$} on the treated and control samples separately, and then set
\smash{$\htau(x) = \hmu_{(1)}(x) - \hmu_{(0)}(x)$}. This approach, however, is often not
robust: Because \smash{$\hmu_{(1)}(x)$} and \smash{$\hmu_{(0)}(x)$} are not trained together,
their difference may be unstable. As an example, consider fitting the lasso \citep{tibshirani1996regression} to estimate $\hat{\mu}_{(1)}(x)$ and $\hat{\mu}_{(0)}(x)$ in the following high-dimensional linear model, \smash{$Y_i(w) = X_i^\top \beta^*_{(w)} + \varepsilon_i(w)$}
with \smash{$X_i, \beta^*_{(w)} \in \mathbb{R}^d$}, and \smash{$\EE{\varepsilon_i(w) \cond X_i} = 0$}. A naive approach would fit two separate lassos to the treated
and control samples,
\begin{equation}
\label{eq:t-lasso}
\hbeta_{(w)} = \argmin_{\beta_{(w)}} \cb{\sum_{\cb{i : W_i = w}} \bigg(Y_i - X_i^\top \beta_{(w)}\bigg)^2 + \lambda_{(w)} \Norm{\beta_{(w)}}_1},
\end{equation}
and then use it to deduce a treatment effect function,
\smash{$\htau(x) = x^\top (\hbeta_{(1)} - \hbeta_{(0)})$}.
However, the fact that both \smash{$\hbeta_{(0)}$} and \smash{$\hbeta_{(1)}$} are regularized towards
0 separately may inadvertently regularize the treatment effect estimate \smash{$\hbeta_{(1)} - \hbeta_{(0)}$}
away from 0, even when $\tau^*(x) = 0$ everywhere. This problem is especially acute when the treated and control samples
are of different sizes; see \citet*{kunzel2019metalearners} for some striking examples.

The recent literature on heterogeneous treatment effect estimation has proposed several ideas on how to avoid such
regularization bias. Some recent papers have proposed structural changes to various machine learning
methods aimed at focusing on accurate estimation of $\tau(\cdot)$
\citep{athey2015machine,hahn2020bayesian,imai2013estimating,powers2018some,shalit2017estimating,
	su2009subgroup,wager2017estimation}.
For example, with the lasso, \citet{imai2013estimating} advocate replacing \eqref{eq:t-lasso} with a single lasso as follows,
\begin{equation}
\label{eq:joint-lasso}
\p{\hb, \, \hdelta} = \argmin_{b, \, \delta} \Bigg[\sum_{i = 1}^n \Bigg\{Y_i - X_i^\top b + (W_i - 0.5)X_i^\top\delta\Bigg\}^2 + \lambda_b \Norm{b}_1 + \lambda_\delta\Norm{\delta}_1\Bigg],
\end{equation}
where then \smash{$\htau(x) = x^\top \hdelta$}.
This approach always correctly regularizes towards a sparse $\delta$-vector for treatment
heterogeneity. The other approaches cited above present variants and improvements of similar ideas in the context
of more sophisticated machine learning methods; see, for example, Figure 1 of \citet*{shalit2017estimating} for
a neural network architecture designed to highlight treatment effect heterogeneity without being affected
by confounders.

Here, instead of trying to modify the algorithms underlying different machine learning tools to improve
their performance as treatment effect estimators, we focus on modifying the loss function used to
training generic machine learning methods. In doing so, we build on the research program developed in
\citet{van2003cross}, \citet{van2006targeted} and \citet*{van2007super}, and later fleshed out for the
context of individualized treatment rules by \citet{luedtke2016optimal,luedtke2016statistical,luedtke2016super}.
In an early technical report, \citet{van2003cross} discuss choosing
the best among a potentially growing set of generic statistical rules by cross-validating on a doubly
robust objective. In the case without nuisance components, an $\varepsilon$-net version of this procedure
was shown to have good asymptotic properties \citep*{van2003cross,van2006cross}. Meanwhile, \citet{luedtke2016super}
discuss a class of valid objectives for learning either individualized treatment rules or heterogeneous treatment
effects---the oracle version \eqref{eq:oracle_loss} of our loss function fits within this class---and discuss properties of model
averaging and cross-validation with these objectives. Our contributions with respect to this line of work include using the
$R$-loss for treatment effect estimation via generic machine learning and developing strong excess loss bounds \smash{$\htau(\cdot)$}
that hold for a computationally tractable and widely used approach to non-parametric estimation, namely
penalized regression over a reproducing kernel Hilbert space.

Another closely related trend in the literature has focused on meta-learning approaches that
are not closely tied to any specific machine learning method. \citet*{kunzel2019metalearners} proposed two
approaches to heterogeneous treatment effect estimation via generic machine learning methods.
One, called the $X$-learner, first estimates $\hmu_{(w)}(x)$ via appropriate non-parametric regression
methods. Then, on the treated observations, it defines pseudo-effects \smash{$D_i = Y_i - \hmu_{(0)}^{(-i)}(X_i)$},
and uses them to fit \smash{$\htau_{(1)}(X_i)$} via a non-parametric regression. Another estimator
\smash{$\htau_{(0)}(X_i)$} is obtained analogously, and the
two treatment effect estimators are aggregated as
\begin{equation}
\label{eq:Xlearn}
\htau(x) = \{1 - \he(x)\} \, \htau_{(1)}(x) + \he(x)\htau_{(0)}(x).
\end{equation}
Another method, called the $U$-learner, starts by noticing that
\begin{equation*}
\EE{U_i \cond X_i = x} = \tau(x), \ \ U_i = \frac{Y_i - m^*(X_i)}{W_i - e^*(X_i)},
\end{equation*}
and then fitting $U_i$ on $X_i$ using any off-the-shelf method. Relatedly,
\citet{athey2015machine} and \cite*{tian2014simple} develop methods for heterogeneous treatment effect estimation based on weighting the outcomes or the covariates with the propensity score; for example, we can estimate $\tau^*(\cdot)$ by regressing $Y_i \{W_i - e^*(X_i)\} / \{e^*(X_i)(1 - e^*(X_i))\}$ on $X_i$.
In our experiments, we compare our method at length to those of \citet{kunzel2019metalearners}.
Again, relative to this line of work, our main contribution is our method, the $R$-learner, which provides meaningful
improvements over baselines in a variety of settings, and our analysis, which provides a quasi-oracle
error bound for the conditional average treatment effect function, i.e., where the error of \smash{$\htau$} may
decay faster than that of \smash{$\he$} or \smash{$\hatm$}.

The closest result to us in this line of work is from \citet*{zhao2017selective}, who combine Robinson's
transformation with the lasso to provide valid post-selection inference on effect modification in the
high-dimensional linear model.
To our knowledge, our paper is the first to use Robinson's transformation to motivate
a loss function that is used in a general machine learning context.

Our formal results draw from the literature on semiparametric efficiency and
constructions of orthogonal moments including \citet{robinson1988root}
and, more broadly, \citet*{belloni2017program}, \citet*{bickel1993efficient}, \citet*{chernozhukov2017double}, \citet{newey1994asymptotic}, \citet{robins2004optimal},
\citet{robins1}, \citet*{robins2017minimax}, \citet{tsiatis2007semiparametric}, \citet{van2011targeted}, etc., that aim at $\sqrt{n}$-rate
estimation of a target parameter in the presence of nuisance components that cannot be estimated at a $\sqrt{n}$ rate.
Algorithmically, our approach has a close connection to targeted maximum likelihood estimation \citep*{scharfstein1999adjusting,van2006targeted}, which starts by estimating nuisance
components non-parametrically, and then uses these first stage estimates to define a likelihood function that
is optimized in a second step.
We also note that using held-out prediction for nuisance components, also known as cross-fitting,
is an increasingly popular approach for making machine learning methods usable in classical semiparametrics \citep{athey2017efficient,chernozhukov2017double,schick1986asymptotically,van2011targeted,wager2016high}.

The main difference between this literature and our results is that
existing results typically focus on estimating a single (or low-dimensional) target parameter, whereas we seek
to estimate an object $\tau^*(\cdot)$ that may also be quite complicated itself.
Another research direction that also uses ideas from semiparametrics to estimate complex objects
is centered on estimating optimal treatment allocation rules
\citep*{athey2017efficient,dudik2011doubly,laber2015tree,luedtke2016super,zhang2012estimating}. This problem is
closely related to, but subtly different from the problem of estimating $\tau^*(\cdot)$ under squared-error
loss; see \citet{kitagawa2018should}.

Finally, we note that all results presented here assume a sampling model where observations are drawn
at random from a population, and we define our target estimand $\tau(\cdot)$ in terms of moments of that
population. \citet*{ding2019decomposing} consider heterogeneous treatment effect estimation in a strict
randomization inference setting, where we the features and potential outcomes
\smash{$\cb{X_i, \, Y_i(0), \, Y_i(1)}_{i = 1}^n$} are taken as fixed and only the treatment $W_i$ is
random \citep{imbens_rubin_2015}; they then show how to estimate the projection of the realized treatment
heterogeneity $Y_i(1) - Y_i(0)$ onto the linear span of the $X_i$. It would be interesting to consider whether
it is possible to derive useful results on non-parametric (regularized) heterogeneous treatment effect estimation
under randomization inference.

\begin{figure}
	\begin{center}
		\begin{tabular}{cc}
			\includegraphics[height=0.45\textwidth]{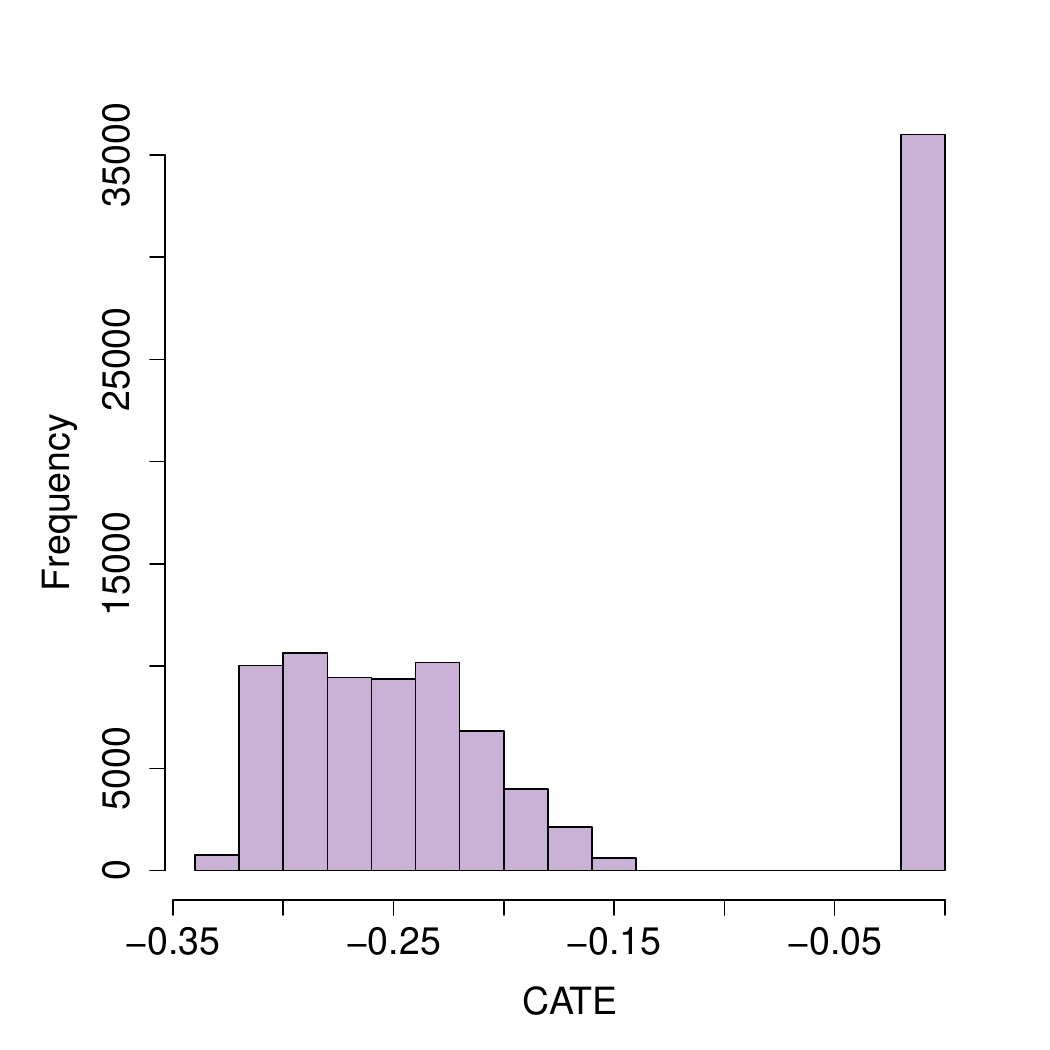} &
			\includegraphics[height=0.45\textwidth, trim = 10mm 0mm 20mm 0mm]{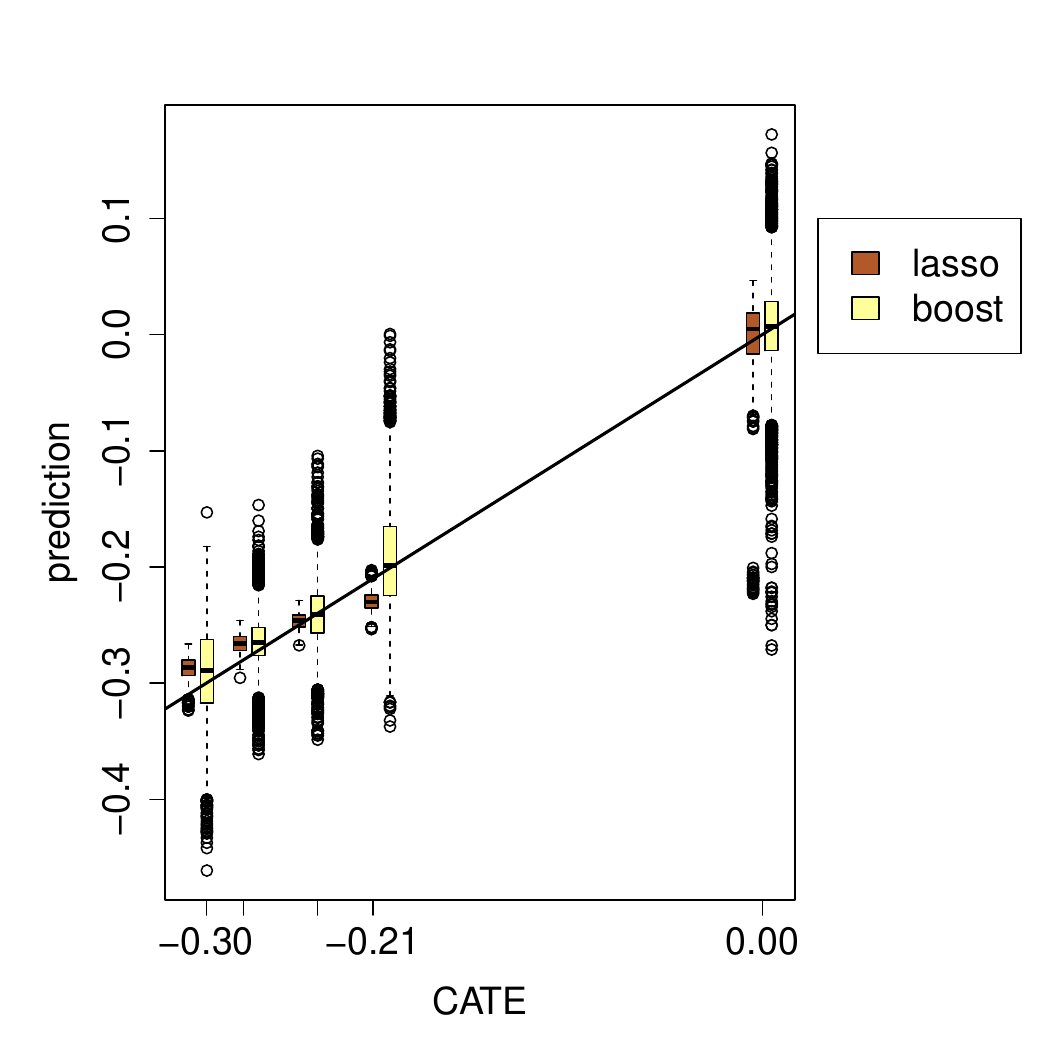} \\
			Histogram of CATE & $R$-learner estimates of CATE
		\end{tabular}
		\caption{The left panel shows the distribution of the conditional average treatment effect (CATE) function $\tau(X_i)$
			on the test set. The right panel compares the true $\tau(X_i)$ to estimates \smash{$\htau(X_i)$} obtained via the $R$-learner running the lasso and boosting respectively to minimize the $R$-loss, again on the test set. As discussed in Section \ref{sec:example}, both
			of them use nuisance components estimated via boosting.}
		\label{fig:example}
	\end{center}
\end{figure}

\section{The R-Learner in Action}


\subsection{Application to a Voting Study}
\label{sec:example}

To see how the $R$-learner works in practice, we consider an example motivated by
\citet*{arceneaux2006comparing}, who studied the effect of paid get-out-the-vote calls on voter turnout.
A common difficulty in comparing the accuracy of heterogeneous treatment effect estimators on real data
is that we do not have access to the ground truth. From this perspective, a major advantage of this application is that
\citet{arceneaux2006comparing} found no effect of get-out-the-vote calls on voter turnout, which suggests that the underlying effect is close to nonexistant. We then spike the original dataset with a synthetic treatment effect $\tau^*(\cdot)$
such as to make the task of estimating heterogeneous treatment effects non-trivial. In other words, both the baseline
signal and propensity scores are from real data; however, $\tau^*(\cdot)$ is chosen by us, and so we can check whether
different methods in fact succeed in recovering it.

The design of \citet{arceneaux2006comparing} was randomized separately by state and competitiveness of the election,
and accounting for varying treatment propensities is necessary for obtaining correct causal effects: A naive
analysis ignoring variable treatment propensities estimates the average effect of a single
get-out-the-vote call on turnout as 4\%, whereas an appropriate analysis finds with high confidence that any treatment
effect must be smaller than 1\% in absolute value.
Although the randomization probabilities were known to the experimenters, we here hide them from our algorithm, and require it to
learn a model \smash{$\he(\cdot)$} for the treatment propensities. We also note that, in the original data, not all voters
assigned to be contacted could in fact answer the phone call, meaning that all effects should be interpreted as
intent to treat effects.
We focus on $d = 11$ covariates (including state, county, age, gender, etc.). Both the outcome $Y$ and the treatment $W$ are binary.
The full sample has $1,895,468$ observations, of which $59,264$ were assigned treatment. 
For our analysis, we focused on a subset of
$148,160$ samples containing all the treated units and a random subset of the controls; thus, 2/5 of our
analysis sample was treated. We further divided this sample into a training set of size 100,000, a test set of size
25,000, and a holdout set with the rest.

As discussed above, for the purpose of this evaluation, we assume that the treatment effect in the original data is 0, and spike in a synthetic treatment
effect $\tau^*(X_i) = - \text{VOTE00}_i / (2 + 100 / \text{AGE}_i)$,
where $\text{VOTE00}_i$ indicates whether
the $i$-th unit voted in the year 2000, and $\text{AGE}_i$ is their age. Because
the outcomes are binary, we add in the synthetic treatment effect by
strategically flipping some outcome labels. Denote the original unflipped outcomes as $Y_i^*$. To add in a treatment
effect $\tau^*(\cdot)$, we first draw Bernoulli random variables $R_i$ with probability $\abs{\tau^*(X_i)}$.
Then, if $R_i = 0$, we set $Y_i(0) = Y_i(1) = Y_i^*$, whereas if $R_i = 1$, we set $\{Y_i(0), \, Y_i(1)\}$ to $(0, \, 1)$
or $(1, \, 0)$ depending on whether $\tau^*(X_i) > 0$ or $\tau^*(X_i) < 0$ respectively. Finally, we set
$Y_i = Y_i(W_i)$. As is typical in causal inference applications, the treatment
heterogeneity here is quite subtle, with $\textrm{var}\{\tau^*(X)\} = 0.016$, and so a large sample size is needed in order to
reject a null hypothesis of no treatment heterogeneity.

To use the $R$-learner, we first estimated \smash{$\he(\cdot)$} and \smash{$\hatm(\cdot)$} to form the
$R$-loss function in \eqref{eq:main}. To do so, we fit models for the nuisance components via both boosting and the
lasso with tuning parameters selected via cross-validation. Then, 
we chose the model that minimized cross-validated error. 
This criterion lead us to pick boosting for both \smash{$\he(\cdot)$} and \smash{$\hatm(\cdot)$}. Another option would
have been to combine predictions from the lasso and boosting models, as advocated by \citet*{van2007super}.

Next, we optimized the $R$-loss function. We again tried methods based on both the lasso and boosting. This
time, the lasso achieved a slightly lower training set cross-validated $R$-loss than boosting, namely 0.1816 versus 0.1818.
Because treatment effects are so weak and so there is potential to overfit even in cross-validation, we also
examined $R$-loss on the holdout set. The lasso again came out ahead, and the improvement
in $R$-loss is stable, 0.1781 versus 0.1783. We thus chose the lasso-based \smash{$\htau(\cdot)$} fit as our final model for $\tau^*(\cdot)$.
As an aside, we note that although the improvement in $R$-loss is stable, the
loss itself is somewhat different between the training and holdout samples. This appears to be due to
the term \smash{$n^{-1} \sum_i \{Y_i - \mu^*_{(W_i)}(X_i)\}^2$} induced by irreducible outcome noise. This term is large
and noisy in absolute terms; however, it gets canceled out when comparing the accuracy of two models. This phenomenon
plays a key role in understanding the behavior of model selection via cross-validation \citep{wager2020cross,yang2007consistency}.

\sloppy{
	Given the constructed CATE function $\tau^*(\cdot)$ in
	our semi-synthetic data generative distribution, we can evaluate the
	oracle test set mean-squared error, \smash{$1/n_{test} \sum_{\cb{i \in test}} \{\htau(X_i) - \tau^*(X_i)\}^2$}.
	Here, it is clear that the lasso did substantially better than boosting,
	achieving a mean-squared error of $0.47 \times 10^{-3}$ versus $1.23 \times 10^{-3}$.
	The right panel of Figure \ref{fig:example} compares \smash{$\htau(\cdot)$} estimates from minimizing the $R$-loss using the lasso and boosting respectively. The lasso is somewhat biased, but boosting is noisy, and the bias-variance trade-off
	favors the lasso in this case. With a larger sample size, we'd expect boosting to achieve lower mean-squared error.}

\begin{figure}
	\centering
	\begin{tabular}{cc}
		\includegraphics[width=0.45\textwidth]{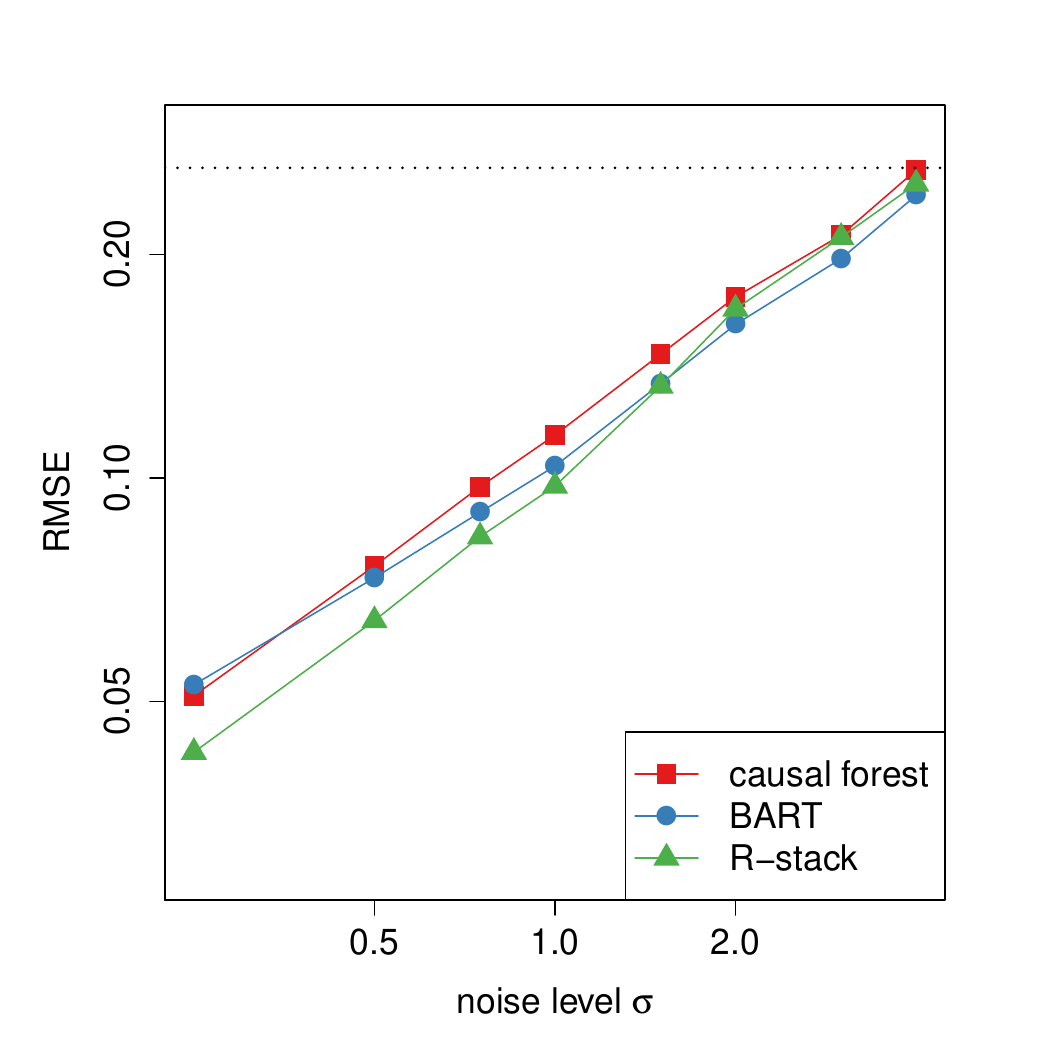} &
		\includegraphics[width=0.45\textwidth]{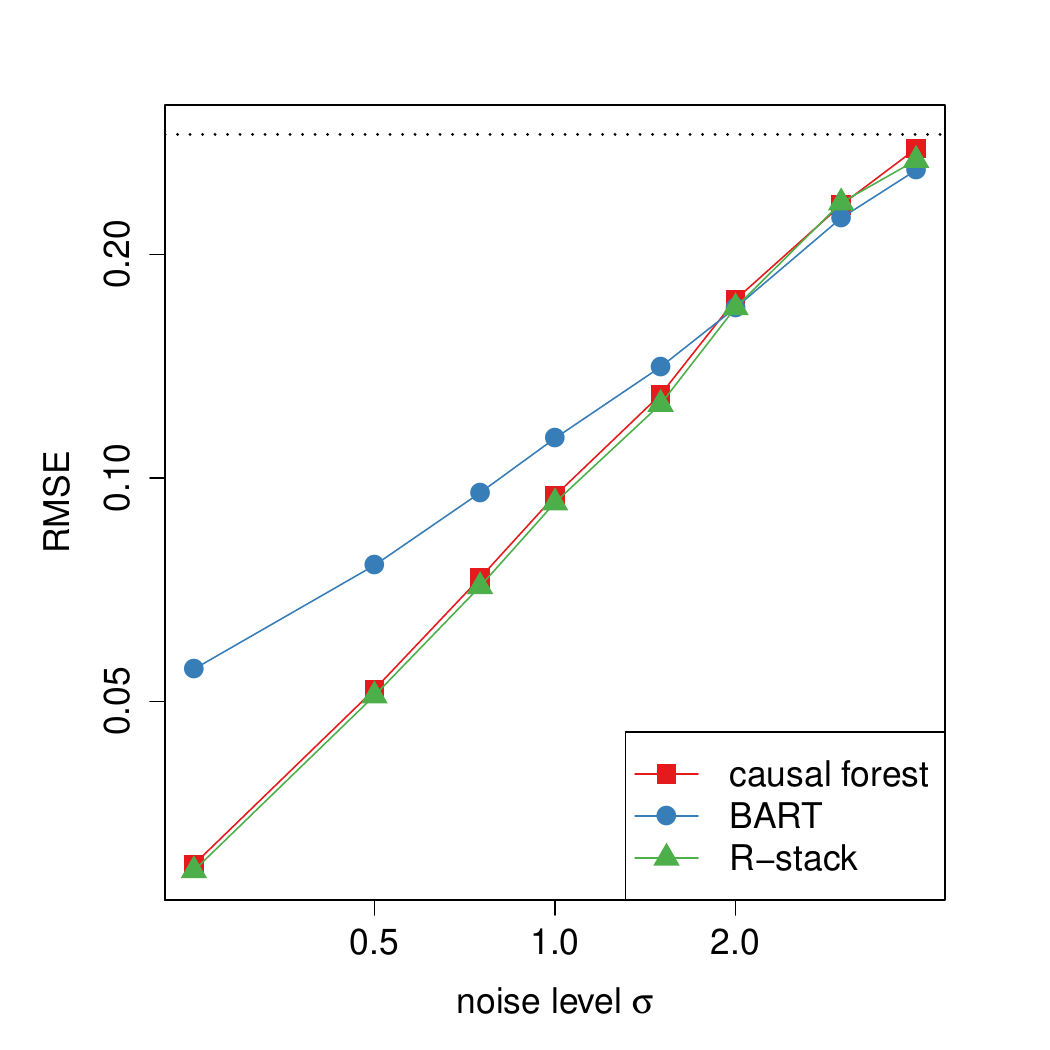} \\
		smooth $\tau^*(\cdot)$ & discontinuous $\tau^*(\cdot)$ 
	\end{tabular}
	\caption{Root-mean squared error (RMSE) on the data-generating design \eqref{eq:stack_dgp},
		for different noise levels $\sigma$. For reference, 
		the RMSE of the optimal constant predictor $\tau^*(X_i)$ is shown as a dotted line.
		All results are aggregated over 50 replications.}
	\label{fig:stack}
\end{figure}

We also compared our approach to both the single lasso approach \eqref{eq:joint-lasso}, and a popular
non-parametric approach to heterogeneous treatment effect estimation via BART \citep{hill2011bayesian}, with
the estimated propensity score added in as a feature following the recommendation of \citet*{hahn2020bayesian}.
The single lasso got an oracle test set error of $0.61 \times 10^{-3}$, whereas BART got $4.05 \times 10^{-3}$.
It thus appears that, in this example, there is value in using a non-parametric method for estimating  \smash{$\he(\cdot)$}
and \smash{$\hatm(\cdot)$}, but then using the simpler lasso for \smash{$\htau(\cdot)$}. In contrast, the single lasso
approach uses linear modeling everywhere (thus leading to potential model misspecification and confounding), whereas BART uses
non-parametric modeling everywhere, which can make  it difficult to obtain a stable $\tau(\cdot)$ fit.
Section \ref{sec:simu} has a more comprehensive simulation evaluation of the $R$-learner relative to several
baselines, including the meta-learners of \citet*{kunzel2019metalearners}.

\subsection{Model Averaging with the R-Learner}
\label{sec:stacking}

In the previous section, we considered an example application where we were willing to carefully consider
the estimation strategies used in each step of the $R$-learner. In other cases, however, a practitioner may
prefer to use some off-the-shelf treatment effect estimators as the starting point for their analysis.
Here, we discuss how to use the $R$-learning approach to build a consensus treatment effect estimate
via a variant of stacking \citep*{breiman1996stacked,luedtke2016super,van2007super,wolpert1992stacked}.

Suppose we start with $k = 1, \, ..., \, K$ different treatment effect estimators
\smash{$\htau_k$}, and that we have access to out-of-fold estimates \smash{$\htau_k^{(-i)}(X_i)$} on
our training set. Suppose, moreover, that we have trusted out-of-fold estimates $\he^{(-i)}(X_i)$ and
$\hatm^{(-i)}(X_i)$ for the propensity score and main effect respectively. Then, we propose building a
consensus estimate \smash{$\htau(\cdot)$} by taking the best positive linear combination
of the \smash{$\htau_k(\cdot)$} according to the $R$-loss:
\begin{equation}
\label{eq:stack}
\begin{split}
&\htau(x) = \hc + \sum_{k = 1}^K \alpha_k \htau_k(x), \ \
\p{\hat{b}, \hc, \halpha} = \argmin_{b, \, c, \, \alpha}\Bigg\{\sum_{i = 1}^n \Bigg[\Bigg\{Y_i - \hatm^{(-i)}(X_i)\Bigg\} - b - \\
&\ \ \ \ \ \ \ \ \ \ \ \ \ \ \ \ \ \ \Bigg\{c + \sum_{k = 1}^K \alpha_k \htau^{(-i)}(X_i)\Bigg\}\Bigg\{W_i - \he^{(-i)}(X_i)\Bigg\}\Bigg]^2 : \alpha \geq 0\Bigg\}.
\end{split}
\end{equation}
For flexibility, we also allow the stacking step \eqref{eq:stack} to freely adjust a constant treatment effect
term $c$, and we add an intercept $b$ that can be used to absorb any potential bias of \smash{$\hatm$}.

We test this approach on the following data-generation distributions. In both cases, we
drew $n = 10,000$ i.i.d. samples from a randomized study design,
$X_i \sim \nn\p{0, \,I_{d \times d}}$
\begin{equation}
\label{eq:stack_dgp}
\begin{split}
W_i \sim \text{Bernoulli}(0.5), \ \ 
Y_i \cond X_i, \, W_i \sim \nn\Bigg\{\frac{3}{1 + e^{X_{i3} - X_{i2}}} + (W_i - 0.5) \, \tau^*(X_i), \, \sigma^2\Bigg\}, 
\end{split}
\end{equation}
for different choices of $\tau^*(\cdot)$ and $\sigma$, and with $d = 10$.
We consider both a smooth treatment effect
function $\tau^*(X_i) = 1/(1 + e^{X_{i1} - X_{i2}})$, and a discontinuous
$\tau^*(X_i) = \mathbbm{1}\{(X_{i1} > 0)\}/(1 + e^{- X_{i2}})$.
Given this data-generating process, we tried estimating $\tau(\cdot)$ via
BART \citep*{hahn2020bayesian,hill2011bayesian},
causal forests \citep*{athey2018generalized,wager2017estimation},
and a stacked combination of the two using \eqref{eq:stack}. We assume that
the experimenter knows that the data was randomized, and used \smash{$\he(x) = 0.5$}
in any place a propensity score was needed. For stacking, we estimated \smash{$\hatm(\cdot)$}
using a random forest.

Results are shown in Figure \ref{fig:stack}. In the example with a smooth $\tau^*(\cdot)$, BART
slightly out-performs causal forests, while stacking does better than either on its own until the
noise level $\sigma$ gets very large---in which case none of the methods do much better than
a constant treatment effect estimator. Meanwhile, the setting with the discontinuous $\tau^*(\cdot)$
appears to be particularly favorable to causal forests, at least for lower noise levels. Here, stacking
is able to automatically match the performance of the more accurate base
learner.

\section{A Quasi-Oracle Error Bound}
\label{sec:rkhs}

As discussed in the introduction, the high-level goal of our formal analysis is to establish error bounds for $R$-learning that
only depend on the complexity of $\tau^*(\cdot)$, and that match the error bounds we could achieve if we knew
$m^*(\cdot)$ and $e^*(\cdot)$ a-priori. In order to do so, we focus on a variant of the $R$-learner based on
penalized kernel regression. The problem of regularized kernel learning covers a broad class of methods
that have been thoroughly studied in the statistical learning literature
\citep[see, e.g.,][]{bartlett2006empirical,caponnetto2007optimal,cucker2002mathematical,mendelson2010regularization,
	steinwart2008support}, and thus provides an ideal case study
for examining the asymptotic behavior of the $R$-learner.

We study \smash{$\Norm{\cdot}_\hh$}-penalized kernel regression,
where $\hh$ is a reproducing kernel Hilbert space (RKHS) with a continuous, positive semi-definite kernel function $\mathcal{K}$.
Let $\pp$ be a non-negative measure over the compact metric space $\xx \subset \mathbb{R}^d$,
and let $\mathcal{K}$ be a kernel with respect to $\pp$. Let \smash{$T_\mathcal{K}: L_2(\pp)\to L_2(\pp)$} be defined as \smash{$T_\mathcal{K}(f)(\cdot) = E\{\mathcal{K}(\cdot, \, X) f(X)\}$}. By Mercer's theorem \citep{cucker2002mathematical}, there is an orthonormal basis of eigenfunctions $(\psi_j)_{j=1}^\infty$ of $T_\mathcal{K}$ with corresponding eigenvalues $(\sigma_j)_{j=1}^\infty$ such that
$\mathcal{K}(x,y) = \sum_{j=1}^\infty \sigma_j \psi_j(x) \psi_j(y)$.
Consider the function \smash{$\phi: \xx \to l_2$} defined by \smash{$\phi(x) = (\sqrt{\sigma_j}\psi_j(x))_{j=1}^\infty$}. Following \citet{mendelson2010regularization}, we define the RKHS $\hh$ to be the image of $l_2$: For every $t\in l_2$, define the corresponding element in $\hh$ by \smash{$f_t(x) = \langle \phi(x), t \rangle$} with the induced inner product \smash{$\langle f_s, f_t \rangle_\hh = \langle t, s \rangle$}. 

\begin{assumption}
	\label{assu:rkhs}
	Without loss of generality, we assume \smash{$\mathcal{K}(x, \, x) \leq 1$} for all \smash{$x \in \xx$}. We assume that for $0 < p < 1$, the eigenvalues \smash{$\sigma_j$} satisfy \smash{$G = \sup_{j\geq 1} j^{1/p}\sigma_j$} for some constant \smash{$G < \infty$},
	and that the orthonormal eigenfunctions $\psi_j(\cdot)$  with \smash{$\Norm{\psi_j}_{L_2(\mathcal{P})} = 1$} are uniformly bounded,
	i.e., \smash{$\sup_j \Norm{\psi_j}_\infty \leq A < \infty$}.
	Finally, we assume that the outcomes $Y_i$ are almost surely bounded, $\abs{Y_i} \leq M$.
\end{assumption}

\begin{assumption}
	\label{assu:approx}
	The true CATE function \smash{$\tau^*(x) = E\{Y_i(1) - Y_i(0) \cond X_i = x\}$}
	satisfies \smash{$\Norm{T_\mathcal{K}^\alpha \{\tau^*(\cdot)\}}_\hh < \infty$} for some
	\smash{$0 < \alpha < 1/2$}. 
\end{assumption}

To interpret the assumption above, note that we do not assume
that $\tau^*(\cdot)$ has a finite $\hh$-norm; rather, we only assume that 
we can make it have a finite $\hh$-norm after a sufficient amount of smoothing.
More concretely, with \smash{$\alpha = 0$}, \smash{$T_\mathcal{K}^\alpha$} would be the identity operator, and so this assumption would be
equivalent to the strongest possible assumption that \smash{$\Norm{\tau^*(\cdot)}_\hh < \infty$} itself.
Then, as \smash{$\alpha$} grows, this assumption gets progressively weaker, and at $\alpha = 1/2$ it would devolve
to simply asking that \smash{$\tau^*(\cdot)$} belong to the space \smash{$L_2(\mathcal{P})$} of square-integrable
functions.

We study oracle penalized regressions $\ttau(\cdot)$ that minimize the following objective,
\begin{equation}
\label{eq:oracle_kernel}
\begin{split}
&\ttau(\cdot) = \argmin\bigg(\frac{1}{n} \sum_{i = 1}^n \bigg[\bigg\{Y_i - m^*(X_i)\bigg\} - \bigg\{W_i - e^*(X_i)\bigg\} \tau(X_i)\bigg]^2 \\
&\ \ \ \ \ \ \ \  + \Lambda_n\p{\Norm{\tau}_\hh} : \Norm{\tau}_\infty \leq 2M\bigg),
\end{split}
\end{equation}
as well as feasible analogues obtained by cross-fitting \citep{chernozhukov2017double,schick1986asymptotically}:
\begin{equation}
\label{eq:main_kernel}
\begin{split}
&\htau(\cdot) = \argmin_{\tau \in \hh} \Bigg(\frac{1}{n} \sum_{i = 1}^n \Bigg[\Big\{Y_i - \hatm^{(-q(i))}(X_i)\Big\} \\
&\ \ \ \ \ \ \ \  - \Big\{W_i - \he^{(-q(i))}(X_i)\Big\} \tau(X_i)\Bigg]^2 +  \Lambda_n\p{\Norm{\tau}_\hh} : \Norm{\tau}_\infty \leq 2M \Bigg),
\end{split}
\end{equation}
Adding the upper bound
\smash{$\Norm{\tau}_\infty \leq 2M$} (or, in fact, any finite upper bound on $\tau$)
enables us to rule out some pathological behaviors.

We seek to characterize the accuracy of our estimator \smash{$\htau(\cdot)$}
by bounding its regret \smash{$R(\htau)$},
\begin{equation*}
R(\tau) = L(\tau) - L(\tau^*), \ \ L(\tau) = \EE{\Bigg[\Big\{Y_i - m^*(X_i)\Big\} - \tau(X_i) \Big\{W_i - e^*(X_i)\Big\}\Bigg]^2}.
\end{equation*}
Recall that, by the expansion \eqref{eq:robinson}, we have
\smash{$E\{Y_i - m^*(X_i) \cond X_i, \, W_i\} = \tau^*(X_i)\{W_i - e^*(X_i)\}$}, implying that
$$ L(\tau) = E[\textrm{var}\{Y_i - m^*(X_i) \cond X_i, \, W_i\}] + E[ \{\tau(X_i)-\tau^*(X_i)\}^2\p{W_i-e^*(X_i)}^2], $$
and \smash{$R(\tau) = E[ \{\tau(X_i)-\tau^*(X_i)\}^2\{W_i-e^*(X_i)\}^2]$}. Thus
if we have overlap, i.e., there is an $\eta > 0$ such that $\eta < e^*(x) < 1 - \eta$ for all $x \in \xx$,
then
\begin{equation}
\label{eq:R_tau_l2p}
(1-\eta)^{-2} R(\tau) < E[\{\tau(X_i) - \tau^*(X_i)\}^2] < \eta^{-2} R(\tau),
\end{equation}
meaning that regret bounds translate into squared-error loss bounds for $\tau(\cdot)$, and vice-versa. We note that when the overlap parameter $\eta$ gets close to 0, the coupling \eqref{eq:R_tau_l2p}
gets fairly loose.

The sharpest regret bounds for the oracle learner \eqref{eq:oracle_kernel} under
Assumptions \ref{assu:rkhs} and \ref{assu:approx}
are due to \citet{mendelson2010regularization} (see also \citet*{steinwart2009optimal}),
and scale as
\begin{equation}
\label{eq:oracle_regret}
R\p{\ttau} = \too_P\p{n^{-\frac{1 - 2\alpha}{p + (1 - 2\alpha)}}},
\end{equation}
where the \smash{$\too_P$}-notation hides logarithmic factors. In the case $\alpha = 0$ where
$\tau^*$ is within the RKHS used for penalization, we recover the more familiar $n^{-1/(1+p)}$
rate established by \citet{caponnetto2007optimal}.
Again, our goal is to establish excess loss bounds for our feasible estimator \smash{$\htau$} that
match the bound \eqref{eq:oracle_regret} available to the oracle that knows $m^*(\cdot)$ and $e^*(\cdot)$
a-priori.


In order to do so, we first need to briefly review the proof techniques
underlying \eqref{eq:oracle_regret}. The argument of \citet{mendelson2010regularization} relies on the following
quasi-isomorphic coordinate projection lemma of \citet{bartlett2008fast}.
To state this result, write
\begin{equation}
\label{eq:Hc}
\hh_c = \cb{\tau : \Norm{\tau}_\hh \leq c,\  \Norm{\tau}_\infty \leq 2M}
\end{equation}
for the radius-$c$ ball of $\hh$ capped by $2M$, let
\smash{$\tau^*_c = \argmin\cb{L(\tau) : \tau \in \hh_c}$} denote the best approximation to $\tau^*$ within $\hh_c$,
and define $c$-regret \smash{$R(\tau; \, c) = L(\tau) - L(\tau^*_c)$} over $\tau \in \hh_c$.
We also define the estimated and oracle $c$-regret functions $\hR_n$ and $\tR_n$ written
in terms of the estimated and oracle losses $\hL_n$ and $\tL_n$:
\begin{align*}
&\hR_n(\tau; \, c) = \hL_n(\tau) - \hL_n(\tau^*_c), \ \ \ \ \tR_n(\tau; \, c) = \tL_n(\tau) - \tL_n(\tau^*_c), \\
&\tL_n(\tau) = \frac{1}{n} \sum_{i = 1}^n \Big[Y_i - m^*(X_i) - \tau(X_i)\Big\{W_i - e^*(X_i)\Big\}\Big]^2, \\
&\hL_n(\tau) = \frac{1}{n} \sum_{i = 1}^n \Big[Y_i - \hat{m}^{(-q(i))}(X_i) - \tau(X_i)\Big\{W_i - \hat{e}^{(-q(i))}(X_i)\Big\}\Big]^2.
\end{align*}
\smash{$\hat{R}_n(\tau; \, c)$} is not observable as it depends on $\tau^*_c$;
however, this does not hinder us from establishing high-probability bounds for it.
The lemma below is adapted from \citet{bartlett2008fast}.

\begin{lemma}
	\label{lemm:bmn}
	Let \smash{$\check{L}_n(\tau)$} be any loss function, and
	\smash{$\check{R}_n(\tau; \, c) = \check{L}_n(\tau) - \check{L}_n(\tau^*_c)$} be the associated regret. Let $\rho_n(c)$ be a continuous positive function that is increasing in $c$. 
	Suppose that, for every $1 \leq c \leq C$ and some $k > 1$, the following inequality holds:
	\begin{equation}
	\label{eq:isomorphism}
	\frac{1}{k} \check{R}_{n}(\tau; \, c) - \rho_n(c) \leq R(\tau; \, c) \leq k \check{R}_n(\tau; \, c) + \rho_n(c)
	\ \text{ for all } \ \tau \in \hh_c.
	\end{equation}
	Then, writing $\kappa_1 = 2k + \frac{1}{k}$ and $\kappa_2 = 2k^2 + 3$,
	any solution to the empirical minimization problem with regularizer \smash{$\Lambda_n(c) \geq \rho_n(c)$},
	\begin{equation*}
	\begin{split}
	&\check{\tau} \in \argmin_{\tau \in \hh_C} \cb{\check{L}(\tau) + \kappa_1 \Lambda_n\p{\Norm{\tau}_\hh}},
	\end{split}
	\end{equation*}
	also satisfies the following risk bound:
	\begin{equation*}
	L\p{\check{\tau}} \leq \inf_{\tau\in \hh_C} \cb{L(\tau) + \kappa_2 \Lambda_n \p{\Norm{\tau}_\hh} }.
	\end{equation*}
\end{lemma}

In other words, the above lemma reduces the problem of deriving regret bounds 
to establishing quasi-isomorphisms as in \eqref{eq:isomorphism}, and any with-high-probability
quasi-isomorphism guarantee yields a with-high-probability regret bound. In particular, we can
use this approach to prove the regret bound \eqref{eq:oracle_regret} for the oracle learner as follows.
We first need a with-high-probability quasi-isomorphism of the following form,
\begin{equation}
\label{eq:oracle_isomoprhism}
\frac{1}{k} \tR_{n}(\tau; \, c) - \rho_n(c) \leq R(\tau; \, c) \leq k \tR_n(\tau; \, c) + \rho_n(c).
\end{equation}
\citet{mendelson2010regularization} provide such a bound for $\rho_n(c)$ that scales as
\begin{equation}
\label{eq:MNrho}
\rho_n(c) \sim \{1 + \log(n) + \log\log\p{c + e}\} \p{\frac{\p{c + 1}^p \log(n)}{\sqrt{n}}}^{2/(1 + p)}.
\end{equation}
Lemma \ref{lemm:bmn} then immediately implies that penalized regression over $\hh_C$ with the oracle
loss function \smash{$\tL(\cdot)$} and regularizer $\kappa_1\rho_n(c)$ satisfies the bound below with high probability:
\begin{equation*}
R(\tilde{\tau}) = L\p{\ttau} - L\p{\tau^*} \leq \inf_{\tau\in \hh_C} \cb{L(\tau) + \kappa_2 \rho_n \p{\Norm{\tau}_\hh}} - L\p{\tau^*}.
\end{equation*}
Furthermore, following Corollary 2.7 in \citet{mendelson2010regularization}, for any $1 \leq c \leq C$,
we also have 
\begin{equation}
\label{eq:approx_err}
\inf_{\tau\in \hh_C} \cb{L(\tau) + \kappa_2 \rho_n \p{\Norm{\tau}_\hh}}
\leq L\p{\tau^*} + \{L\p{\tau^*_c} - L\p{\tau^*}\} + \kappa_2 \rho_n(c).
\end{equation}
Note that \citet{mendelson2010regularization} considers
the case where $C = \infty$; here, instead, we only take $C$ to be large enough for our
argument; see the proof for details. Finally, 
noting the scaling of $\rho_n(c)$ in \eqref{eq:MNrho} and the approximation error bound
\begin{equation} 
\label{eq:approx}
L\p{\tau^*_c} - L\p{\tau^*} \leq c^{(2\alpha - 1)/\alpha} \Norm{T_\mathcal{K}^\alpha \{\tau^*(\cdot)\}}_\hh^{1/\alpha}
\end{equation}
established by \citet{smale2003estimating} under the setting of Assumption \ref{assu:approx},
we achieve a practical regret bound by choosing $c = c_n$ to optimize the right-hand side
of \eqref{eq:approx_err}.
The specific rate in \eqref{eq:oracle_regret} arises by setting \smash{$c_n = n^{{\alpha}/(p + (1 - 2\alpha))}$}.

For our purposes, the upshot is that if we can match the strength of the quasi-isomorphism bounds
\eqref{eq:oracle_isomoprhism} with our feasible loss function, i.e., get an analogous bound in terms of 
\smash{$\hR_n$} as opposed to \smash{$\tR_n$}, then we can also match the rate of
any regret bounds proved using the above argument.
The proof of the following result relies several concentration results, including Talagrand's inequality
and generic chaining \citep{talagrand2006generic}, and makes heavy use
of cross-fitting style arguments \citep{chernozhukov2017double,schick1986asymptotically,van2011targeted}.

\begin{lemma}
	\label{lemm:cross-fit-shortened}
	Given the conditions in Lemma \ref{lemm:bmn}, suppose that the propensity estimate $\he(x)$ is uniformly consistent, 
	$\xi_n := \sup_{x \in \mathcal{X}} \abs{\hat{e}(x) - e^*(x)} \to_p 0$,
	and the $L_2$ errors converge at rate
	\begin{equation*}
	E[\{\hat{m}(X) - m^*(X)\}^2], \ E[\{\hat{e}(X) - e^*(X)\}^2] = \oo\p{a_n^2}
	\end{equation*}
	for some sequence $a_n$ such that 
	$a_n = \oo\p{n^{-\kappa}} \ with \ \kappa > \frac{1}{4}$. 
	Suppose, moreover, that we have overlap, i.e.,
	$\eta < e^*(x) < 1 - \eta$ for some $\eta > 0$, and that
	Assumptions \ref{assu:rkhs} and \ref{assu:approx} hold.
	\begin{align}
	\abs{\hR_n(\tau; \, c) - \tR_n(\tau; \, c)} \leq 0.125 R(\tau; \, c) + o(\rho_n(c)),\label{eq:R_h_R_t_inequality}
	\end{align}
	with probability at least $1-\varepsilon$, for all $\tau \in \hh_c, 1\leq c \leq c_n \log(n)$ with $c_n = n^{\alpha/(p+1-2\alpha)}$ for large enough $n$.
\end{lemma}

This result implies that we can turn any quasi-isomorphism for the oracle learner \eqref{eq:oracle_isomoprhism} with error
$\rho_n(c)$ into a quasi-isomoprhism bound for \smash{$\hR(\tau)$} with error inflated by the right hand side of \eqref{eq:R_h_R_t_inequality}.
Thus, given any regret bound for the oracle learner built using Lemma \ref{lemm:bmn}, we
can also get an analogous regret bound for the feasible learner provided we regularize just
a little bit more. The following result makes this formal.

\begin{theorem}
	\label{theo:rlearn}
	Given the conditions of Lemma \ref{lemm:cross-fit-shortened} and that $2\alpha < 1-p$, suppose 
	that we obtain $\htau(\cdot)$ via a penalized kernel regression variant of the $R$-learner \eqref{eq:main_kernel}, with a properly chosen
	penalty of the form \smash{$\Lambda_n(\Norm{\htau}_\hh)$} specified in the proof.
	Then \smash{$\htau(\cdot)$} satisfies the same regret bound \eqref{eq:oracle_regret} as \smash{$\ttau(\cdot)$}, i.e., 
	$R\p{\htau} = \too_P\p{n^{-(1 - 2\alpha)/\{p + (1 - 2\alpha)\}}}$.
\end{theorem}

In other words, we have found that with penalized kernel regression, the $R$-learner can match the
best available performance guarantees available for the oracle learner \eqref{eq:oracle_kernel} that knows
everything about the data generating distribution except the true treatment effect function---both
the feasible and the oracle learner satisfy
\begin{equation}
\label{eq:rn}
R(\htau), \, R(\ttau) = \too_P(r^2_n), \ \with \ r_n = n^{-(1 - 2\alpha)/[2\{p + (1 - 2\alpha)\}]}.
\end{equation}
As we approach the semiparametric case, i.e., $\alpha, \, p \rightarrow 0$,
we recover the well-known result from the semiparametric inference literature that,
in order to get $1/\sqrt{n}$-consistent inference for a single
target parameter, we need 4-th root consistent nuisance parameter estimates; see
\citet{chernozhukov2017double} for a review and references.
We also note that after we disseminated a first draft of our paper, several authors have established
further quasi-oracle type results for the $R$-learner and related methods; see in particular \citet{foster2019orthogonal}
and \citet{kennedy2020optimal}.

We emphasize that our quasi-oracle result depends on a local robustness property of the $R$-loss function, and
does not hold for general meta-learners; for example, it does not hold for the $X$-learner of \citet*{kunzel2019metalearners}.
To see this, we argue by contradiction: We show that it is possible to make $o(n^{-1/4})$-changes to the nuisance
components \smash{$\hmu_{(w)}(x)$} used by the $X$-learner that induce changes in the $X$-learner's \smash{$\htau(\cdot)$}
estimates that dominate the error scale in \eqref{eq:rn}. Thus, there must be some choices of $o(n^{-1/4})$-consistent
\smash{$\hmu_{(w)}(x)$} with which the $X$-learner does not converge at the rate \eqref{eq:rn}.
The contradiction arises as follows: Pick $\xi > 0$ such that $0.25 + \xi <  (1 - 2\alpha)/[2\{p + (1 - 2\alpha)\}]$, and
modify the nuisance components used to form the $X$-learner in \eqref{eq:Xlearn}
such that \smash{$\hmu_{(0)}(x) \leftarrow \hmu_{(0)}(x) - c / n^{0.25 + \xi}$} and
\smash{$\hmu_{(1)}(x) \leftarrow \hmu_{(1)}(x) + c / n^{0.25 + \xi}$}.
Recall that the $X$-learner fits \smash{$\hat{\tau}_{(1)}(\cdot)$} by minimizing
${n_1^{-1}} \sum_{W_i = 1} \{Y_i -  \hat{\mu}_{(0)}^{(-i)}(X_i) - \tau_{(1)}(X_i)\}^2$,
and fits \smash{$\hat{\tau}_{(0)}(\cdot)$} by solving an analogous problem on the controlled units.
Combining the $\hat{\tau}_{(w)}$ estimates from these two loss functions, we see by inspection that its final estimate of the treatment effect is also shifted by \smash{$\htau(x) \leftarrow \htau(x) + c / n^{0.25 + \xi}$}.
The perturbations \smash{$c / n^{0.25 + \xi}$} are vanishingly small on the $n^{-1/4}$ scale,
and so would not affect conditions analogous to those of Theorem \ref{theo:rlearn}; yet they have a big enough effect
on $\htau(x)$ to break any convergence results on the scale of \eqref{eq:rn}. We note that \citet{kunzel2019metalearners} 
do have some quasi-oracle type results; however, they only
focus on the case where the number of control units
$\abs{\cb{W_i = 0}}$ grows much faster than the number of treated units $\abs{\cb{W_i = 1}}$.
In this case, they show that the $X$-learner performs as well as an oracle who already knew the mean
response function for the controls, \smash{$\mu^*_{(0)}(x) = \EE{Y_i(0) \cond X_i = x}$}. Intriguingly, in
this special case, we have \smash{$m^*(x) \approx \mu^*_{(0)}(x)$} and $e^*(x) \approx 0$, and so the $R$-learner as in \eqref{eq:main_kernel}
is roughly equivalent to the $X$-learner procedure \eqref{eq:Xlearn}. Thus, at least qualitatively,
we can interpret the result of \citet{kunzel2019metalearners} as a special case of our result in the case where the number
of controls dominates the number of treated units, or vice-versa.

\section{Simulation Experiments}
\label{sec:simu}
\subsection{Baseline methods and simulation setups}
Our approach to heterogeneous treatment effect estimation via learning objectives can
be implemented using any method that is framed as a loss minimization problem, such as
boosting, decision trees, etc. In this section, we focus on simulation experiments using the
$R$-learner,  a direct implementation of \eqref{eq:main} based on lasso, kernel ridge regression, and boosting.
We follow the nomenclature of \citet{kunzel2019metalearners} and consider the following 
methods for heterogeneous treatment effect estimation as baselines.
The $S$-learner fits a single model for \smash{$f(x, \, w) = \EE{Y \cond X = x, \, W = w}$},
and then estimates \smash{$\htau(x) = \hf(x, 1) - \hf(x, \, 0)$};
the $T$-learner fits the functions \smash{$\mu^*_{(w)}(x) = \EE{Y \cond X = x, \, W = w}$} separately for $w\in\{0,1\}$,
and then estimates \smash{$\htau(x) = \hmu_{(1)}(x) - \hmu_{(0)}(x)$}; 
the $X$-learner and $U$-learner are as described in Section \ref{sec:relworks}. In addition, for the boosting-based experiments, we consider the causal boosting algorithm (denoted by $CB$ in Section \ref{subsec:boost}) proposed by \citet{powers2018some}.

Finally, for the lasso-based experiments, we consider an additional variant of our method,
the $RS$-learner, that combines the spirit of $R$- and $S$-learners by
adding an additional term in the loss function and then separately penalizes the main and treatment
effect terms as in \citet{imai2013estimating}. Specifically, we use \smash{$\htau(x) = x^\top \hdelta$},
where \smash{$\hb$} and \smash{$\hdelta$} minimize
\begin{equation*}
\begin{split}
\frac{1}{n} \sum_{i = 1}^n \Big[Y_i - \hatm^{(-i)}(X_i) - X_i^\top b 
- \{W_i - \he^{(-i)}(X_i)\}  X_i^\top\delta\Big]^2 + \lambda \p{ \Norm{b}_1 +  \Norm{\delta}_1}. 
\end{split}
\end{equation*}
Heuristically, one may hope that the $RS$-learner may be more robust, as it has an additional term to
eliminate confounders. 

In all simulations, we generate data as follows:
for different choices of $X$-distribution $P_d$ indexed by dimension $d$, noise level $\sigma$,
propensity function $e^*(\cdot)$, baseline main effect $b^*(\cdot)$, and treatment effect function $\tau^*(\cdot)$:
\begin{equation*}
\begin{split}
&X_i \sim P_d, \ \ W_i \mid X_i \sim \textrm{Bernoulli}(e^*(X_i)), \ \  \varepsilon_i\mid X_i \sim \mathcal{N}(0,1), \\ 
&Y_i = b^*(X_i)+(W_i-0.5)\tau^*(X_i) + \sigma \varepsilon_i.
\end{split}
\end{equation*}
We consider the following specific simulation designs.
Setup A has difficult nuisance components and an easy treatment effect function. 
We use the scaled \citet{friedman1991multivariate} function  for the baseline main effect
\smash{$b^*(X_i) = \sin(\pi X_{i1} X_{i2})+ 2(X_{i3}-0.5)^2+ X_{i4}+0.5 X_{i5}$},
along with \smash{$X_i \sim \textrm{Unif}(0,1)^d$},
\smash{$e^*(X_i)=\textrm{trim}_{0.1} \{\sin(\pi X_{i1} X_{i2})\}$}
and $\tau^*(X_i) = (X_{i1}+X_{i2})/2$,
where $\textrm{trim}_\eta(x) = \max\{\eta, \, \min(x, 1-\eta)\}$.
Setup B employs a randomized trial. 
Here, $e^*(x) = 1/2$ for all $x \in \RR^d$, so it is possible to be accurate without explicitly controlling
for confounding. We take
\smash{$X_i \sim \mathcal{N}(0,I_{d\times d})$},
\smash{$\tau^*(X_i)=X_{i1} + \log(1+e^{X_{i2}})$},
and \smash{$b^*(X_i) = \max\{X_{i1}+X_{i2}, X_{i3},0\} + \max\{X_{i4}+X_{i5},0\}$}.
Setup C has an easy propensity score and a difficult baseline. 
In this setup, there is strong confounding, but the propensity score is much easier to estimate
than the baseline:
\smash{$X_i \sim \mathcal{N}(0,I_{d\times d})$},
\smash{$e^*(X_i) = {1}/(1+e^{X_{i2}+X_{i3}})$},
\smash{$b^*(X_i) = 2\log(1+e^{X_{i1}+X_{i2}+X_{i3}})$},
and the treatment effect is constant, $\tau^*(X_i)= 1$.
Setup D has unrelated treatment and control arms, with data generated as
\smash{$X_i \sim \mathcal{N}(0,I_{d\times d})$, $e^*(X_i) = 1/(1+e^{-X_{i1}}+e^{-X_{i2}})$},
$\smash{\tau^*(X_i)=  \max\{X_{i1}+X_{i2}+X_{i3},0\}} - \smash{\max\{X_{i4}+X_{i5},0\}}$, and
\smash{$b^*(X_i) = \p{\max\{X_{i1}+X_{i2}+X_{i3},0\} + \max\{X_{i4}+X_{i5},0\}}/2$}.
Here, $\mu^*_{(0)}(X)$ and $\mu^*_{(1)}(X)$ are uncorrelated, and so there is no upside to learning
them jointly.

\subsection{Lasso-based experiments}
\label{subsec:lasso}
In this section, we compare $S$-, $T$-, $X$-, $U$-, and our $R$- and $RS$-learners implemented via the lasso on
simulated designs. For the $S$-learner, we follow \citet{imai2013estimating} using
\eqref{eq:joint-lasso}, while for the $T$-learner, we use \eqref{eq:t-lasso}.
For the $X$-, $R$-, and $RS$-learners, we use $L_1$-penalized logistic regression to estimate propensity $\hat{e}$,
and the lasso for all other regression estimates.

For all estimators, we run the lasso on the pairwise interactions of a natural spline basis expansion with 7 degrees of freedom on $X_i$. We generate $n$ data points as the training set and generate a separate test set also with $n$ data points, and the reported mean-squared error is on the test set. The penalty parameter is chosen by 10-fold cross validation.
For the $R$- and $RS$-learners, we use 10-fold cross-fitting on $\hat{e}$ and $\hat{m}$ in \eqref{eq:main}. All methods are implemented via \texttt{glmnet} \citep*{friedman2010regularization}. We note that the $U$-learner suffers from high variance and instablity due to dividing by the propensity estimates.
Therefore, we set a cutoff for the propensity estimate to be at level 0.05. We have also found empirically $U$-learner achieves much lower estimation error if we choose to use the largest regularization parameter that achieves 1 standard error away from the minimum in the cross validation step. Therefore, the $U$-learner uses \texttt{lambda.1se} as its cross validation parameter; other learners use \texttt{lambda.min} in \texttt{glmnet}.

\begin{figure}[!htbp]
	\begin{center}
		\includegraphics[scale=1.2] {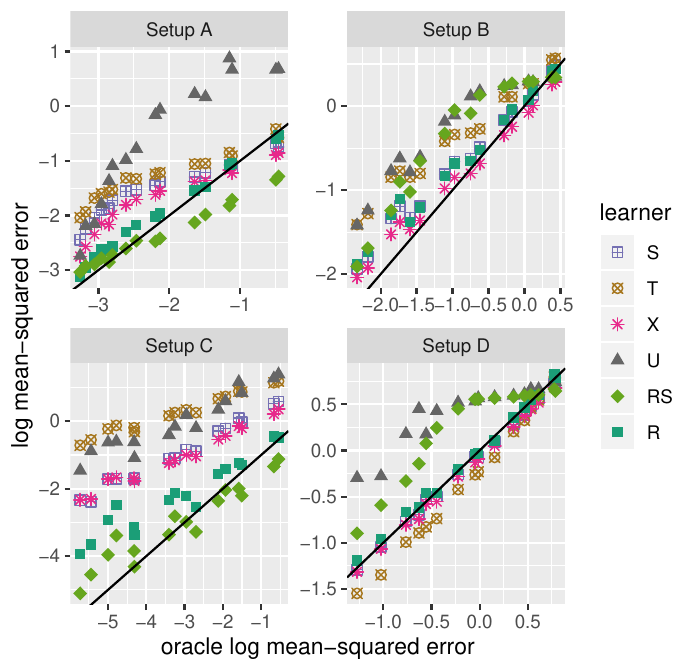} 
		\caption{Performance of lasso-based $S$-, $T$-, $X$-, $U$-, $RS$- and $R$-learners, relative
			to a lasso-based oracle learner \eqref{eq:oracle}, across simulation Setups A--D described
			in Section \ref{sec:simu}; recall that (A) has complicated nuisance components but a
			simple $\tau(\cdot)$ function, (B) is a randomized trial, (C) has a simple propensity function
			but a complicated main effect function, and (D) has unrelated treatment and control response surfaces.
			We report results for all combinations of $n \in \cb{500, \, 1000}$, $d \in \cb{6, \, 12}$
			and $\sigma \in \cb{0.5, \, 1, \, 2, \, 3}$, and each point on the plots represents the average performance of
			one learner for one of these 16 parameter specifications.
			All mean-squared error numbers are aggregated over 500 runs and reported on an independent test set,
			and are plotted on the logarithmic scale.
			Detailed simulation results are reported in Appendix \ref{sec:extrasimu}.}
		\label{fig:lasso}
	\end{center}
\end{figure}

In Figure \ref{fig:lasso}, we compare the performance of our 6 considered methods to
an oracle that runs the lasso on \eqref{eq:oracle}, for different values of sample size $n$,
dimension $d$, and noise level $\sigma$.
As is clear from these illustrations, the considered simulation settings differ vastly in difficulty, both in terms of the accuracy
of the oracle, and in terms of the ability of feasible methods to approach the oracle.
The raw numbers depicted in Figure \ref{fig:lasso}
is available in Appendix \ref{sec:extrasimu}.

In Setups $A$ and $C$, where there is complicated confounding that needs to be overcome before we can estimate
a simple treatment effect function $\tau^*(\cdot)$, the $R$- and $RS$-learners stand out.
All methods do reasonably well in the randomized trial (Setup $B$) where it was not
necessary to adjust for confounding, and the $X$-, $S$-, and $R$-learners do best. Finally,
having completely disjoint functions for the treated and control arms is unusual in practice.
However, we consider this possibility in Setup $D$, where there is no reason to model
\smash{$\mu^*_{(0)}(x)$} and \smash{$\mu^*_{(1)}(x)$}
jointly, and find that the $T$-learner---which in fact models them separately---performs well.

Overall, the $R$- and $RS$-learner consistently achieve good performance and,
in most simulation specifications, essentially match the
performance of the oracle \eqref{eq:oracle} in terms of the mean-squared error.
The $U$-learner suffers from high loss due to its instability.

\subsection{Kernel ridge regression based experiments}
\label{subsec:kernel}

We move on to compare $S$-, $T$-, $X$-, $U$-, and $R$- learners implemented via kernel ridge regression with a Gaussian kernel. We use a variant of the \texttt{KRLS} package \citep{krls}, available at \url{https://github.com/lukesonnet/KRLS}, that allows for weighted regression.
For fitting the objective in each subroutine in all methods, we run a 5-fold cross validation to search through the width of the Gaussian kernel and the ridge regularization parameter both from a grid of $10^k$s with $k$ ranging in $[-3,\,-2.5,\, -2,\,\ldots,2, \, 2.5,\, 3]$.  We experiment on the same set of setups and parameter variations, including variations on sample size $n$, dimension $d$, and noise level $\sigma$  as in Section \ref{subsec:lasso}, and include all numbers depicted in Figure \ref{fig:kernel} in Appendix \ref{sec:extrasimu}.
In Figure \ref{fig:kernel}, we again observe that the $R$-learner does particularly well in Setups $A$ and $C$,
where the treatment effect functions are relatively simple and the treatment propensity is not constant.

\begin{figure}[!htbp]
	\begin{center}
		\includegraphics[scale=1.2] {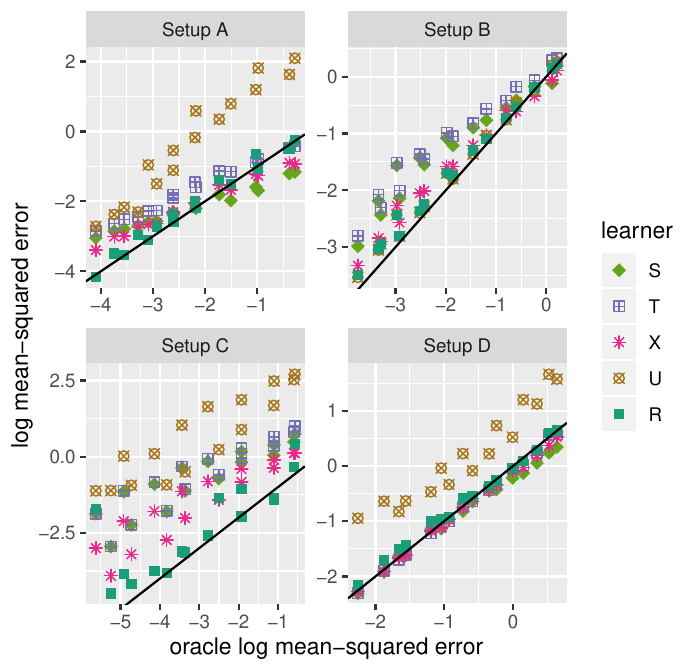} 
		\caption{Performance of $S$-, $T$-, $X$-, $U$-, $RS$- and $R$-learners relative
			to an oracle learner \eqref{eq:oracle} all based on kernel ridge regression with a Gaussian kernel, across simulation Setups A--D described
			in Section \ref{sec:simu}; recall that (A) has complicated nuisance components but a
			simple $\tau(\cdot)$ function, (B) is a randomized trial, (C) has a simple propensity function
			but a complicated main effect function, and (D) has unrelated treatment and control response surfaces.
			We report results for all combinations of $n \in \cb{500, \, 1000}$, $d \in \cb{6, \, 12}$
			and $\sigma \in \cb{0.5, \, 1, \, 2, \, 3}$, and each point on the plots represents the average performance of
			one learner for one of these 16 parameter specifications.
			All mean-squared error numbers are aggregated over 200 runs and reported on an independent test set,
			and are plotted on the logarithmic scale.
			Detailed simulation results are reported in Appendix \ref{sec:extrasimu}.}
		\label{fig:kernel}
	\end{center}
\end{figure}

\begin{figure}[!htbp]
	\begin{center}
		\includegraphics[scale=1.2]{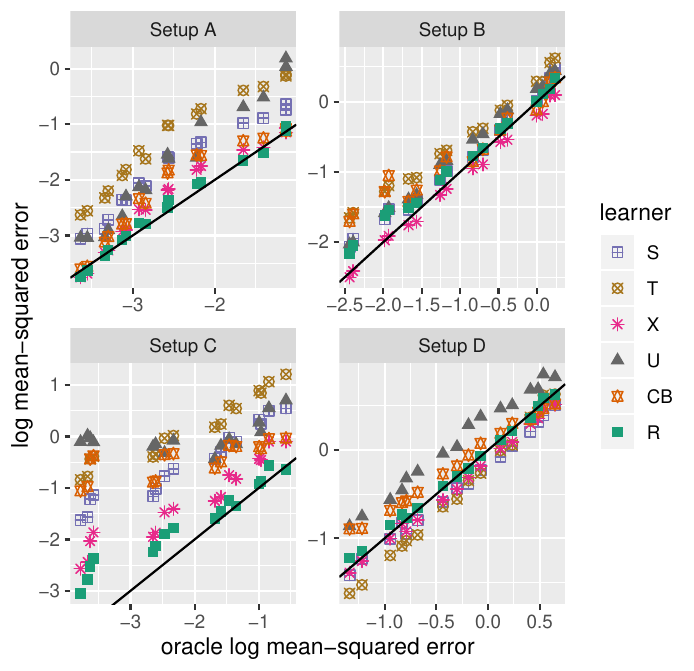} 
		\caption{Performance of boosting-based $S$-, $T$-, $X$-, $U$-, $R$-learners as well as causal boosting ($CB$), relative
			to a boosting-based oracle learner \eqref{eq:oracle}, across simulation Setups A--D described
			in Section \ref{sec:simu}; recall that (A) has complicated nuisance components but a
			simple $\tau(\cdot)$ function, (B) is a randomized trial, (C) has a simple propensity function
			but a complicated main effect function, and (D) has unrelated treatment and control response surfaces.
			We report results for all combinations of $n \in \cb{500, \, 1000}$, $d \in \cb{6, \, 12}$
			and $\sigma \in \cb{0.5, \, 1, \, 2, \, 3}$, and each point on the plots represents the average performance of
			one learner for one of these 16 parameter specifications.
			All mean-squared error numbers are aggregated over 200 runs and reported on an independent test set,
			and are plotted on the logarithmic scale.
			Detailed simulation results are reported in Appendix \ref{sec:extrasimu}.}
		\label{fig:boost}
	\end{center}
\end{figure}

\subsection{Gradient boosting-based experiments}
\label{subsec:boost}

Finally, we compare $S$-, $T$-, $X$-, $U$-, and $R$- learners implemented via gradient boosting, as well as the causal boosting ($CB$) algorithm. We use the \texttt{causalLearning R} package for $CB$, while all other methods are implemented via \texttt{XGboost} \citep{chen2016xgboost}. For fitting the objective in each subroutine in all methods, we draw a random set of 10 combinations of hyperparmaeters from the following grid: \texttt{subsample}$=[0.5,\, 0.75,\, 1]$, \texttt{colsample\_bytree}$=[0.6,\,  0.8,\,  1]$, \texttt{eta}$=[5\text{e-}3,\,  1\text{e-}2,\,  1.5\text{e-}2,\,  2.5\text{e-}2,\,  5\text{e-}2,\,  8\text{e-}2,\,  1\text{e-}1,\,  2\text{e-}1]$, \texttt{max\_depth}$=[3,\cdots, 20]$, \texttt{gamma}=\texttt{Uniform}$(0,\, 0.2)$, \texttt{min\_child\_weight}$=[1, \cdots, 20]$, \texttt{max\_delta\_step}$=[1, \cdots, 10]$, and cross validate on the number of boosted trees for each combination with an early stopping of 10 iterations. We experiment on the same set of setups and parameter variations including variations on sample size $n$, dimension $d$, and noise level $\sigma$ as in Section \ref{subsec:lasso}, and include all numbers depicted in Figure \ref{fig:boost} in Appendix \ref{sec:extrasimu}.

In Figure \ref{fig:boost}, we observe again that $R$-learner stands out in Setup $A$ and $C$, with all methods performing reasonably well in the randomized control setting of Setup $B$; in Setup $D$, the $T$-learner performs best since the the treated and control arms are generated from unrelated functions. 

Before we conclude this section, we note that in both sets of the experiments, for simplicity of illustration, we have used lasso, kernel ridge regression, and boosting respectively to learn $\hat{m}(\cdot)$ and $\hat{e}(\cdot)$. In practice, we recommend cross validating on a variety of black-box learners, e.g. lasso, random forests, neural networks, etc. that are tuned for prediction accuracy to learn these two pilot quantities.  All simulation results above can be replicated using the publicly available \texttt{rlearner} package for \texttt{R} \citep{CRAN}, available at \url{https://github.com/xnie/rlearner}.

\section{Discussion and Extensions}


A natural generalization of our setup arises when, in some applications, we need to work with multiple
treatment options. For example, in medicine, we may want to compare a control condition to multiple different
experimental treatments. If there are $k$ different treatments along with a control arm, we can encode
$W \in \cb{0, \, 1}^k$, and note that a multivariate version of Robinson's transformation suggests the
following estimator,
\begin{equation*}
\begin{split}
&\htau(\cdot) = \argmin_{\tau} \Bigg(\frac{1}{n} \sum_{i = 1}^n \Big[\Big\{Y_i - \hatm^{(-i)}(X_i)\Big\} 
- \left\langle W_i - \he^{(-i)}(X_i), \tau(X_i) \right\rangle \Big]^2 + \Lambda_n\{\tau(\cdot)\}\Bigg),
\end{split}
\end{equation*}
where the angle brackets indicate an inner product, $e(x) = \EE{W \cond X = x} \in \RR^k$ is a vector, and $\tau_l(x)$ measures the conditional average
treatment effect of the $l$-th treatment arm at $X_i = x$, for $l = 1, \, ..., \, k$. When implementing variants of this approach
in practice, different choices of $\Lambda_n\{\tau(\cdot)\}$ may be needed to reflect relationships between
the treatment effects of different arms, e.g. whether there is a natural ordering of treatment arms, or
if there are some arms that we believe a priori to have similar effects.

It would also be interesting to consider extensions of the $R$-learner to cases where the treatment assignment
$W_i$ is not unconfounded, and we need to rely on an instrument to identify causal effects.
\citet{chernozhukov2017double} discusses how Robinson's approach to the partially linear
model generalizes naturally to this case, and \citet*{athey2018generalized} adapt their causal forest to work with
instruments. The underlying estimating equations, however, cannot be interpreted as loss functions as easily
as \eqref{eq:oracle}, especially in the case where instruments may be weak, and so we leave this extension of
the $R$-learner to future work.

\section*{Acknowledgement}

We are grateful for enlightening conversations with
Susan Athey,
Emma Brunskill,
John Duchi,
Tatsunori Hashimoto,
Guido Imbens,
S\"oren K\"unzel,
Percy Liang,
Whitney Newey,
Mark van der Laan,
Alejandro Schuler,
Robert Tibshirani,
Bin Yu
and Yuchen Zhang
as well as for helpful comments and feedback from seminar participants
at several universities and workshops and from the referees. 
This research was partially supported by a grant from the National Science Foundation,
a Facebook Faculty Award, and a
2018 Stanford Human-Centered AI seed grant.
The first author was awarded a Thomas R. Ten Have Award based on this
work at the 2018 Atlantic Causal Inference Conference.

\setlength{\bibsep}{0.2pt plus 0.3ex}
\bibliographystyle{plainnat-abbrev}
\bibliography{references}
\newpage

\begin{appendix}

	\section{Appendix: Proofs}
	\label{sec:proofs}
	
	\section*{Preliminaries}
	\subsection{A useful inequality relating function norms in RKHS}
	Before beginning our proof, we present an inequality that we will use frequently.
	Under Assumption \ref{assu:rkhs}, directly following from Lemma 5.1 of
	\citet{mendelson2010regularization}, 
	there is a constant $B$ depending on $A$, $p$ and $G$
	such that for all $\tau \in \mathcal{H}$, 
	\begin{equation}
	\label{eq:MN_bound}
	\Norm{\tau}_\infty
	\leq B \Norm{\tau}_\hh^p  \Norm{\tau }_{L_2(\mathcal{P})}^{1 - p}.
	\end{equation}
	If $\eta < e^*(x) < 1 - \eta$ for some $\eta > 0$, a consequence of the above inequality is as follows: for $\tau \in \hh_{c}$,
	\begin{equation}
	\label{eq:infty_bound}
	\Norm{\tau - \tau^*_c}_\infty
	\leq B \Norm{\tau - \tau^*_c}_\hh^p  \Norm{\tau - \tau^*_c}_{L_2(\mathcal{P})}^{1 - p}
	\leq B2^p\eta^{-(1 - p)} \, c^p  R(\tau; \, c)^{(1 - p)/2}.
	\end{equation}
	We note that the second inequality in \eqref{eq:infty_bound} follows from combining \eqref{eq:R_tau_l2p} with the fact that for $\tau \in \hh_{c}$, $\Norm{\tau - \tau^*_c}_\hh \leq 2c$ by the triangle inequality. 
	
	\subsection{Talagrand's Inequalities}
	Below we state Talagrand's Concentration Inequality for an empirical process indexed by a class of uniformly bounded functions \citep{talagrand2006generic}. The version of the inequality we shall use here is due to \citet{massart2000constants}. 
	
	Let $\mathcal{F}$ be a class of functions defined on $(\Omega, \pp)$ such that for every $f\in \mathcal{F}$, $\Norm{f}_\infty \leq b$, and $\EE{f} = 0$. Let $X_1, \cdots, X_n$ be independent random variables distributed according to $\pp$ and set $\sigma^2 = \sup_{f\in\mathcal{F}} \EE{f^2}$. Define 
	$$Z = \sup_{f\in\mathcal{F}} \frac{1}{n}\sum_{i=1}^n f(X_i) \quad and \quad \bar{Z} =  \sup_{f\in\mathcal{F}} \frac{1}{n} \abs{\sum_{i=1}^n f(X_i)}.$$
	Then, there exists an absolute constant $C$ such that for every $t >0$, and every $\rho > 0$,
	\begin{align}
	\textrm{pr}\Big\{Z > (1+\rho)\EE Z + \frac{\sigma}{\sqrt{n}} \sqrt{Ct} + \frac{C}{n}(1+\rho^{-1}) bt\Big\} \leq e^{-t},\label{eq:talagrand-1}\\
	\textrm{pr}\Big\{Z < (1-\rho)\EE Z - \frac{\sigma}{\sqrt{n}} \sqrt{Ct} - \frac{C}{n}(1+\rho^{-1}) bt\Big\} \leq e^{-t},
	\end{align}
	and the same inequalities holds for $\bar{Z}$. 
	
	We will also make use of the following bound from Corollary 3.4 of \citet{talagrand1994sharper}:
	\begin{equation}
	\EE {\sup_{f\in \mathcal{F}} \sum_{i=1}^n f^2(X_i)} \leq n\sigma^2 + 8bE\Bigg\{\sup_{f\in\mathcal{F}} \abs{\sum_{i=1}^n \varepsilon_i f(X_i)}\Bigg\}, \label{eq:talagrand-squared}
	\end{equation}
	where $\varepsilon_i$ are independent Rademacher variables indepedent of the variables $X_i$. 
	
	\section*{Technical definitions and lemmas}
	\subsection{Proof of Lemma \ref{lemm:bmn}}
	\begin{proof}
		First, we note that for $1\leq c \leq C$, $\hh_c$ defined as $\{\tau \in \hh, \Norm{\tau}_\hh \leq c, \Norm{\tau}_\infty \leq 2M\}$ is an ordered set, i.e. $\hh_c \subseteq \hh_{c'}$ for $c \leq c'$. Without loss of generality, it suffices to consider $\Lambda_n(\cdot) = \rho_n(\cdot)$ because if \eqref{eq:isomorphism} holds with $\rho_n(c)$, it also holds with $\rho_n(c)$ replaced by $\Lambda_n(c) \geq \rho_n(c)$. 
		Define
		\begin{align*}
		&\tau_c^* = \argmin_{\tau \in \hh_c} L(\tau),\\
		&\check{\tau}_c = \argmin_{\tau \in \hh_c} \check{L}(\tau),\\
		&\check{m} = \Norm{\check{\tau}}_\hh.
		\end{align*}
		
		Following the proof of Theorem 4 in \citet{bartlett2008fast}, first check the following facts:
		\begin{itemize}
			\item In the event that $c \geq \check{m}$ (see Lemma 5 of \citet{bartlett2008fast}),
			$$ L(\check{\tau}) \leq L(\tau_c^*) + \max\{k\kappa_1+2, 3\}\rho_n(c). $$
			\item In the event that $c \leq \check{m}$ (see Lemma 6 of \citet{bartlett2008fast}),
			$$ L(\tau_{\check{m}}^*) \leq L(\tau_c^*) + \p{\frac{1}{k^2} - \frac{\kappa_1}{k} + 1} \rho_n(\check{m}) + \frac{\kappa_1\rho_n(c)}{k}. $$
			\item In the event that $c \leq \check{m}$ (see Lemma 7 of \citet{bartlett2008fast}),
			$$ L(\check{\tau}) \leq L(\tau_c^*) + \p{\frac{1}{k^2} - \frac{\kappa_1}{k} + 2}\rho_n(\check{m}) + \frac{\kappa_1\rho_n(c)}{k}. $$
			
			Now, choosing $\kappa_1 = \frac{1}{k}+2k$ shows that 
			\begin{align*}
			L(\check{\tau}) \leq L(\tau_c^*) +  \frac{\kappa_1\rho_n(c)}{k}.
			\end{align*}
		\end{itemize}
		Let $\kappa_2 = 2k^2 + 3$, combining the above,
		\begin{align*}
		L(\check{\tau}) \leq \inf_{1\leq c\leq C} L(\tau_c^*) + \kappa_2 \rho_n(c).
		\end{align*}
		Finally, for any \smash{$\tau = \argmin_{\tau\in\hh_C} \{L(\tau) + \kappa_2 \Lambda_n\p{\Norm{\tau}_\hh}\}$}, \smash{$\tau = \tau_{\Norm{\tau}_\hh}^*$}. Suppose not, then \smash{$L(\tau) + \kappa_2 \Lambda_n\p{\Norm{\tau}_\hh} >  L(\tau_{\Norm{\tau}_\hh}^*) + \kappa_2 \Lambda_n\p{\Norm{\tau}_\hh}$}, which is a contradiction. Thus, the claim follows.
	\end{proof}
	\subsection{Technical Definitions and Auxillary Lemmas}
	Before we proceed with the proof of Lemma \ref{lemm:cross-fit-shortened}, it is helpful to prove the following results. 
	\begin{definition} [Definition 2.4 from \citet{mendelson2010regularization}]
		Given a class of functions $F$, we say that $\{F_c: c\geq 1\}$ is an ordered, parameterized hierarchy of $F$ if the following conditions are satisfied:
		\begin{itemize}
			\item $\{F_c: c\geq 1\}$ is monotone;
			\item for every $c\geq 1$, there exists a unique element $f_c^* \in F_c$ such that $L(f_c^*) = \inf_{f\in F_c} L(f)$;
			\item the map $c \to L(f_c^*)$ is continuous;
			\item for every $c_0 \geq 1$, $\cap_{c > c_0} F_c = F_{c_0}$
			\item $\cup_{c\geq 1}F_c = F$
		\end{itemize}
	\end{definition}
	
	\begin{lemma}
		\label{lemm:hiercharchy}
		$\hh_c := \{\tau \in \hh, \Norm{\tau}_\hh \leq c, \Norm{\tau}_\infty \leq 2M\}$ is an ordered, parameterized hiercharchy of $\hh$.
		\proof
		First, we show that $\hh_1$ is compact. Let $(\tau_n)_n$ be a sequence in $\hh_1$. Following from the fact that $B_1 = \{\tau \in \hh, \Norm{\tau}_\hh \leq 1\}$ is compact with respect to $L_2$-norm, $\tau_n$ has a converging subsequence $(\tau_{n_k})_k$ with a limit $\tau \in B_1$. For any $\varepsilon > 0$, there exists $K$ such that for all $k > K$, $\Norm{\tau_{n_k} - \tau}_{L_2(\mathcal{P})} \leq \varepsilon$. Suppose $\Norm{\tau}_\infty > 2M$, then take $\tau'(x) = \min(\max(\tau(x), -2M), 2M)$, we see that $\Norm{ \tau_{n_k}(x) - \tau'(x)}_{L_2(\mathcal{P})} \leq \Norm{ \tau_{n_k}(x) - \tau(x)}_{L_2(\mathcal{P})}$ for all $k \geq K$. So the limit $\tau(x) = \tau'(x)$. Thus the subsequence converges to a limit in $\hh_1$, and so $\hh_1$ is compact.
		The proof now follows exactly the proof of Lemma 3.6 in \citet{mendelson2010regularization}.
		\endproof
	\end{lemma}

	\begin{lemma}[chaining]
		\label{lemm:chaining}
		Let $\hh$ be an RKHS with kernel $\mathcal{K}$ satisfying Assumption \ref{assu:rkhs},
		let $X_1, \, ..., \, X_n$ be $n$ independent draws from the measure $\pp$,
		and let $Z_1, \, ..., \, Z_n$ be independent mean-zero sub-Gaussian random variables
		with variance proxy $M^2$, conditionally on the $X_i$.
		Then, there is a constant $B$ such that,
		for any (potentially random) weighting function $\omega(x)$,
		\begin{equation}
		\label{eq:chain}
		\begin{split}
		&\EE{\sup_{h\in \hh_{c, \, \delta}}\cb{\frac{1}{n} \sum_{i = 1}^n Z_i \omega(X_i) h(X_i)}} 
		\leq B M c^p \delta^{1-p} E\{\omega^2(X)\}^{1/2} \frac{\log(n)}{\sqrt{n}} ,
		\end{split}
		\end{equation}
		where $\hh_{c, \, \delta} := \cb{h \in \hh : \Norm{h}_\hh \leq c, \, \Norm{h}_{L_2(\pp)} \leq \delta}$.
		\proof
		
		Our proof proceeds by generic chaining. Defining random variables
		$$ Q_h  = \frac{1}{n} \sum_{i = 1}^n Z_i \omega(X_i) h(X_i), $$
		the basic generic chaining result of \citet{talagrand2006generic} (Theorem 1.2.6)
		states that if $\{Q_h\}_{h\in \hh_{c,\delta}}$ is a sub-Gaussian process relative to some metric $d$, i.e., for every $h_1, h_2 \in \hh_{c,\delta}$ and every $u\geq 1$,
		\begin{equation}
		\label{eq:chain_delta}
		\textrm{pr}\Big\{\abs{Q_{h_1} - Q_{h_2}} \geq u d(h_1, \, h_2)\Big\} \leq 2 e^{-u^2/2},
		\end{equation}
		then for some universal constant $B$ (not the same as in \eqref{eq:chain}),
		\begin{equation}
		\label{eq:chain_sup}
		\EE{\sup_{h \in \hh_{c, \, \delta}} Q_h} \leq B \gamma_2\p{\hh_{c, \, \delta}, \, d}.
		\end{equation}
		Here, $\gamma_2$ is a measure of the complexity of the space $\hh_{c, \, \delta}$ in terms
		of the metric $d$: writing $\set_j$, $j = 1, \, 2, \, ...$, for a sequence of collections of
		elements form $\hh_{c, \, \delta}$,
		\begin{equation*}
		\gamma_2\p{\hh_{c, \, \delta}, \, d} =
		\inf_{(\set_j)}\cb{\sup_{h \in \hh_{c, \, \delta}} \cb{ \sum_{j = 0}^\infty 2^{j/2} d(h, \, \set_j)} : \abs{\set_0} = 1, \, \abs{\set_j} = 2^{2^j} \eqfor j > 0},
		\end{equation*}
		where the infimum is with respect to all sequences of collections $(\set_j)_{j=0}^\infty$, and $d(h, \, \set) = \inf_{g\in \set}d(h,g)$.
		
		To establish \eqref{eq:chain}, we start by applying generic chaining conditionally on $X_1, \, ..., \, X_n$:
		given a (possibly random) distance measure $d$ such that \eqref{eq:chain_delta} holds conditionally
		on the $X_i$, then \eqref{eq:chain_sup} also provides a uniform bound conditionally on the $X_i$.
		To this end, we study the following metric:
		\begin{align}
		&d(h_1, \, h_2) = \frac{1}{n} Md_{\infty, \, n}(h_1, \, h_2)  \sqrt{\sum_{i = 1}^n \omega^2(X_i)}, \\
		&d_{\infty, \, n}(h_1, \, h_2) = \sup\cb{\abs{h_1(X_i) - h_2(X_i)} : i = 1, \, ..., \, n}.
		\end{align}
		Conditionally on the $X_i$, $Q_{h_1} - Q_{h_2}$ is a sum of $n$ independent mean-zero sub-Gaussian random variables,
		the $i$-th of which is has its sub-Gaussian variance proxy $n^{-2}M^2d^2_{\infty, \, n}(h_1, \, h_2) \omega^2(X_i)$, so
		\eqref{eq:chain_delta} holds by elementary properties of sub-Gaussian random variables. Finally,
		noting that $d(\cdot, \, \cdot)$ is a constant multiple of $d_{\infty, \, n}(\cdot, \, \cdot)$ conditionally on $X_1, \, ..., \, X_n$, the definition of $\gamma_2$ implies that
		$$\gamma_2\p{\hh_{c, \, \delta}, \, d} = \frac{1}{n} M  \sqrt{\sum_{i = 1}^n \omega^2(X_i)} \gamma_2\p{\hh_{c, \, \delta}, \, d_{\infty, n}}.$$
		Our argument so far implies that
		\begin{equation*}
		E\Bigg[\sup_{\hh_{c, \, \delta}}\cb{\frac{1}{n} \sum_{i = 1}^n Z_i \omega(X_i) h(X_i)}  \cond X_1, \, ..., \, X_n\Bigg]
		\leq \frac{BM}{n}   \sqrt{\sum_{i = 1}^n \omega^2(X_i)} \gamma_2\p{\hh_{c, \, \delta}, \, d_{\infty, n}}.
		\end{equation*}
		It now remains to bound moments of $\gamma_2\p{\hh_{c, \, \delta}, \, d_{\infty, n}}$.
		
		Writing $\sigma_j$ for the eigenvalues of $\mathcal{K}$ and
		$A$ for the uniform bound on the eigenfunctions as in Assumption \ref{assu:rkhs}, \citet{mendelson2010regularization}
		show that for another universal constant $B$, (Theorem 4.7) 
		\begin{equation*}
		E\{\gamma_2^2\p{\hh_{c, \, \delta}, \, d_{\infty, \, n}}\}^{1/2} \leq A B \log(n) \sqrt{\sum_{j = 1}^\infty \min\cb{\delta^2, \, \sigma_j c^2/4}},
		\end{equation*}
		and that for yet another universal constant $B_p$ depending only on $p$, (Lemma 3.4)
		\begin{equation*}
		\sum_{j = 1}^\infty \min\cb{\delta^2, \, \sigma_j c^2/4} \leq B_p  \delta^{2(1-p)} c^{2p}G,
		\end{equation*}
		where $G = \sup_{j \geq 1} j^{1/p} \sigma_j$ as defined in Assumption \ref{assu:rkhs}. Thus, by Cauchy-Schwartz,
		\begin{align*}
		E\Bigg\{\gamma_2\p{\hh_{c, \, \delta}, \, d_{\infty,n}}\sqrt{\frac{1}{n}\sum_{i = 1}^n \omega^2(X_i)} \Bigg\}
		&\leq E\Big\{\gamma_2^2\p{\hh_{c, \, \delta}, \, d_{\infty, n}}\Big\}^{1/2} E\Bigg\{\frac{1}{n}\sum_{i = 1}^n \omega^2(X_i)\Bigg\}^{1/2} \\
		& = B \delta^{1-p} c^{p} E\{\omega^2(X)\}^{1/2},
		\end{align*}
		where $B$ is a (different) constant.
		The desired result then follows.
		\endproof
	\end{lemma}
	
	\begin{lemma}
		\label{lemm:tau_c_l2p}
		Suppose we have overlap, i.e., $\eta < e^*(x) < 1-\eta$ for some $\eta>0$, for $1 < c < c'$. Then, the following holds:
		\begin{equation*}
		\Norm{\tau_c^*(X_i) - \tau_{c'}^*(X_i)}_{L_2(\mathcal{P})} \leq \frac{1}{\eta}\p{1-\frac{c}{c'}}\Norm{\tau_{c'}^*}_{L_2(\mathcal{P})}.
		\end{equation*}
		\proof
		First, we note that following a similar derivation behind \eqref{eq:R_tau_l2p}, we have for any $\tau, \tau' \in \mathcal{H}$, 
		\begin{equation*}
		\eta^2 \Norm{\tau(X_i) - \tau'(X_i)}^2_{L_2(\mathcal{P})} \leq \abs{L(\tau) - L(\tau')} \leq (1-\eta)^2 \Norm{\tau(X_i) - \tau'(X_i)}^2_{L_2(\mathcal{P})}.
		\end{equation*}
		Then, we have 
		\begin{align}
		\Norm{\tau_c^*(X_i) -\frac{c}{c'} \tau_{c'}^*(X_i)}^2_{L_2(\mathcal{P})} 
		&\leq \eta^{-2}\Big\{L\p{\frac{c}{c'}\tau_{c'}^*} - L\p{\tau_c^*}\Big\} \nonumber\\
		&\leq \eta^{-2}\Big\{L\p{\frac{c}{c'}\tau_{c'}^*} - L\p{\tau_{c'}^*}\Big\} \nonumber\\
		&\leq \frac{(1-\eta)^2}{\eta^2} \Norm{\tau_{c'}^*}_{L_2(\mathcal{P})}^2\p{1-\frac{c}{c'}}^2. \label{eq:tau_c_star_final}
		\end{align} 
		Finally, by the triangle inequality, 
		\begin{align*}
		\Norm{\tau_c^*(X_i) - \tau_{c'}^*(X_i)}_{L_2(\mathcal{P})} 
		&\leq \Norm{\tau_{c'}^*(X_i) - \frac{c}{c'}\tau_{c'}^*(X_i)}_{L_2(\mathcal{P})} + \Norm{\tau_c^*(X_i) -\frac{c}{c'} \tau_{c'}^*(X_i)}_{L_2(\mathcal{P})} \\
		&\leq \p{1-\frac{c}{c'}} \Norm{\tau_{c'}^*}_{L_2(\mathcal{P})} + \frac{1-\eta}{\eta}\Norm{\tau_{c'}^*}_{L_2(\mathcal{P})} \p{1-\frac{c}{c'}} \\
		&= \frac{1}{\eta}\p{1-\frac{c}{c'}}\Norm{\tau_{c'}^*}_{L_2(\mathcal{P})},
		\end{align*}
		where the second inequality follows from \eqref{eq:tau_c_star_final}.
		\endproof  
	\end{lemma}
	
	\begin{lemma}
		\label{lemm:square_tau_tau_c}
		Simultaneously for all $\tau\in \mathcal{H}_c, c \geq 1, \delta\leq 4M$ where $\Norm{\tau - \tau^*_c}_{L_2(\mathcal{P})} \leq \delta$,  we have
		\begin{align*}
		&\frac{1}{n}\sum_{i=1}^n \p{\tau(X_i) - \tau_c^*(X_i)}^2 \nonumber\\
		&\quad\quad= \oo_P\Bigg\{ \delta^2 +  c^{2p} \delta^{2(1-p)} \frac{\log(n)}{\sqrt{n}} +  c^{p}\delta^{2-p}\frac{1}{\sqrt{n}}\sqrt{\log\p{\frac{cn^{1/(1-p)}}{\delta^2}}} + \frac{1}{n}c^{2p}\delta^{2(1-p)}\log\p{\frac{cn^{1/(1-p)}}{\delta^2}} \nonumber\\
		&\quad\quad\quad\quad\quad +\frac{c^{2p}\delta^{2(1-p)}}{n}
		\Bigg\}
		\end{align*}
		
		\begin{proof}
			We proceed by a localization argument by bounding the quantity of interest over sets indexed by $c$ and $\delta$ such that $\Norm{\tau - \tau_c^*}_{L_2(\mathcal{P})} \leq \delta$, i.e. we bound 
			\begin{align*}
			\sup_{\tau \in \hh_{c}} \cb{\frac{1}{n} \sum_{i=1}^n  \p{\tau(X_i) - \tau_c^*(X_i)}^2: \Norm{\tau - \tau^*_c}_{L_2(\mathcal{P})} \leq \delta}.
			\end{align*}
			First we bound the expectation. Let $\varepsilon_i$ be i.i.d. Rademacher random variables. 
			\begin{align}
			&E\Bigg[\sup_{\tau \in \hh_{c}}  \cb{\frac{1}{n}\sum_{i=1}^n \p{\tau(X_i) - \tau_c^*(X_i)}^2: \Norm{\tau - \tau^*_c}_{L_2(\mathcal{P})} \leq \delta}\Bigg]\nonumber\\
			&\quad\quad \stackrel{(a)}{\leq} \sup_{\tau \in \hh_{c}} \cb{ \Norm{\tau-\tau_c^*}_{L_2(\mathcal{P})}^2:\Norm{\tau - \tau^*_c}_{L_2(\mathcal{P})} \leq \delta } \nonumber\\
			& \quad\quad\quad\quad + 8 \sup_{\tau \in \hh_{c}}\cb{ \Norm{\tau - \tau_c^*}_\infty: \Norm{\tau - \tau^*_c}_{L_2(\mathcal{P})} \leq \delta} \cdot \nonumber \\		
			&\quad\quad\quad\quad\quad \EE{\sup_{\tau \in \hh_{c}} \cb{\frac{1}{n}  \abs{\sum_{i=1}^n \varepsilon_i(\tau(X_i) - \tau_c^*(X_i))}:  \Norm{\tau - \tau^*_c}_{L_2(\mathcal{P})} \leq \delta }}\nonumber \\
			&\quad\quad \stackrel{(b)}{\leq}\sup_{\tau \in \hh_{c}} \cb{ \Norm{\tau-\tau_c^*}_{L_2(\mathcal{P})}^2:\Norm{\tau - \tau^*_c}_{L_2(\mathcal{P})} \leq \delta } \nonumber\\
			& \quad\quad\quad\quad + 8 \sup_{\tau \in \hh_{c}}\cb{ \Norm{\tau - \tau_c^*}_\infty: \Norm{\tau - \tau^*_c}_{L_2(\mathcal{P})} \leq \delta} \cdot \nonumber \\
			&\quad\quad\quad\quad\quad \EE{\sup_{\tau \in \hh_{c}}\cb{ \frac{1}{n} \sum_{i=1}^n \varepsilon_i(\tau(X_i) - \tau_c^*(X_i)) :\Norm{\tau - \tau^*_c}_{L_2(\mathcal{P})} \leq \delta}}\nonumber \\
			&\quad\quad\stackrel{(c)}{\leq} \delta^2+ B c^{2p} \delta^{2(1-p)} \frac{\log(n)}{\sqrt{n}},\label{eq:tala-2-apply}
			\end{align}
			where $(a)$ follows from \eqref{eq:talagrand-squared}, $(b)$ follows from the fact that $\varepsilon_i$ are symmetrical around 0, $(c)$ follows from \eqref{eq:chain}, and $B$ is an absolute constant.
			
			Let $f_{\tau,c}(X_i) = \p{\tau(X_i)-\tau_c^*(X_i)}^2 - \EE{\p{\tau(X_i)-\tau_c^*(X_i)}^2}$. Let $G = \sup_{\tau \in \hh_{c} }  \cb{\frac{1}{n}\sum_{i=1}^nf_{\tau,c}(X_i): \Norm{\tau - \tau^*_c}_{L_2(\mathcal{P})} \leq \delta}$. Note that for a different constant $B$,
			\begin{align*}
			\EE{G} 
			&\leq  \EE{\sup_{\tau \in \hh_{c}} \cb{ \frac{1}{n}\sum_{i=1}^n \p{\tau(X_i)-\tau_c^*(X_i)}^2:\Norm{\tau - \tau^*_c}_{L_2(\mathcal{P})} \leq \delta }} \\
			&\quad+ \sup_{\tau \in \hh_{c}}  \cb{\EE{\p{\tau(X_i)-\tau_c^*(X_i)}^2}:  \Norm{\tau - \tau^*_c}_{L_2(\mathcal{P})} \leq \delta} \\
			&\leq B\p{\delta^2 +  c^{2p} \delta^{2(1-p)} \frac{\log(n)}{\sqrt{n}}},
			\end{align*}
			where we note that bounding the first summand on the right-hand side of the first inequality above follows immediately from \eqref{eq:tala-2-apply}.
			
			We also note that by \eqref{eq:MN_bound}, 
			\begin{align*}
			\sup_{\tau \in \hh_{c}   }\cb{ \Norm{f_{\tau,c}}_\infty: \Norm{\tau - \tau^*_c}_{L_2(\mathcal{P})} \leq \delta} \leq B \Norm{\tau-\tau_c^*}_\infty^2
			\leq B c^{2p} \delta^{2(1-p)}
			\end{align*}
			for another different constant $B$, and that 
			\begin{align*}
			\sup_{\tau \in \hh_{c}} \cb{\EE{f_{\tau,c}^2}: \Norm{\tau - \tau^*_c}_{L_2(\mathcal{P})} \leq \delta} 
			&\leq \sup_{\tau \in \hh_{c}}  \cb{\EE{\p{\tau(X_i)-\tau_c^*(X_i)}^4}:  \Norm{\tau - \tau^*_c}_{L_2(\mathcal{P})} \leq \delta}\\
			&\leq \sup_{\tau \in \hh_{c}} \cb{\Norm{\tau-\tau_c^*}_{L_2(\mathcal{P})}^2 \Norm{\tau-\tau_c^*}_\infty^2:\Norm{\tau - \tau^*_c}_{L_2(\mathcal{P})} \leq \delta } \\
			&\leq  c^{2p}\delta^{2(1-p)+2}.
			\end{align*} 
			
			By Talagrand's concentration inequality \eqref{eq:talagrand-1}, for a fixed $c$ and $\delta$, we have that with probability $1-\varepsilon$,
			\begin{align}
			G \leq \Bigg\{\delta^2 + c^{2p} \delta^{2(1-p)} \frac{\log(n)}{\sqrt{n}} +  c^{p}\delta^{2-p}\frac{1}{\sqrt{n}}\sqrt{\log\p{\frac{1}{\varepsilon}}} + \frac{1}{n}c^{2p}\delta^{2(1-p)}\log\p{\frac{1}{\varepsilon}}\Bigg\}.\label{eq:G_tala}
			\end{align}
			
			We conclude that for a fixed $c$ and $\delta$, we have that with probability $1-\varepsilon$, for a different constant $B$,
			\begin{align*}
			&\sup_{\tau \in \hh_{c}} \cb{\frac{1}{n} (\tau(X_i) - \tau_c^*(X_i))^2: \Norm{\tau - \tau^*_c}_{L_2(\mathcal{P})} \leq \delta }\\
			&\quad\quad\quad\quad \leq G + \sup_{\tau \in \hh_{c}} \cb{E\Bigg[\Big\{\tau(X_i)-\tau_c^*(X_i)\Big\}^2\Bigg]:\Norm{\tau - \tau^*_c}_{L_2(\mathcal{P})} 
				\leq \delta}\\
			&\quad\quad\quad\quad\leq B\Bigg\{\delta^2 + c^{2p} \delta^{2(1-p)} \frac{\log(n)}{\sqrt{n}} +  c^{p}\delta^{2-p}\frac{1}{\sqrt{n}}\sqrt{\log\p{\frac{1}{\varepsilon}}} + \frac{1}{n}c^{2p}\delta^{2(1-p)}\log\p{\frac{1}{\varepsilon}}\Bigg\}.
			\end{align*}
			We proceed with bounding the above for all values of $c$ and $\delta$ simultaneously. 
			For a fixed $k = 0, 1, 2, \cdots$, define $\mathcal{C}^{k,\delta} := \{2^k + j n^{-\frac{1}{1-p}}\delta2^k, j=0,1,2,\cdots,   \ceil{\frac{1}{\delta} (n^{1/(1-p)}-1)}\}$. For a fixed $\delta$, and for any $c\geq 1$, let $u(c,\delta) = \min\{d: d > c, d \in \mathcal{C}^{2^{\floor{\log {c}}}, \delta}\}$. 
			Recall that $\Norm{\tau_c^*}_{L_2(\mathcal{P})} \leq 2M$ by definition, and so by Lemma \ref{lemm:tau_c_l2p}, there is a constant $D$ such that 
			\begin{align}
			\Norm{\tau^*_c - \tau^*_{u(c,\delta)}}_{L_2(\mathcal{P})} \leq Dn^{-\frac{1}{1-p}}\delta. \label{eq:c-uc-l2p}
			\end{align}
			
			Thus, for any $c\geq 1$,
			\begin{align*}
			& \sup_{\tau \in \hh_{c}} \cb{\frac{1}{n}\sum_{i=1}^n \p{\tau(X_i) - \tau_c^*(X_i)}^2:\Norm{\tau - \tau^*_c}_{L_2(\mathcal{P})} \leq \delta}\\
			& \quad\quad\leq \sup_{\tau \in \hh_{u(c,\delta)}} \cb{\frac{1}{n}\sum_{i=1}^n \p{\tau(X_i) - \tau_c^*(X_i)}^2:\Norm{\tau - \tau^*_{u(c,\delta)}}_{L_2(\mathcal{P})} \leq \delta + Dn^{-\frac{1}{1-p}}\delta  }\\
			&\quad\quad\leq \sup_{\tau \in \hh_{u(c,\delta)}} \cb{\frac{1}{n}\sum_{i=1}^n \p{\tau(X_i) - \tau_{u(c,\delta)}^*(X_i)}^2:\Norm{\tau - \tau^*_{u(c,\delta)}}_{L_2(\mathcal{P})} \leq \delta +Dn^{-\frac{1}{1-p}}\delta  } \\
			&\quad\quad\quad\quad +\sup_{\tau \in \hh_{u(c,\delta)}}\Bigg\{ \frac{1}{n}\sum_{i=1}^n \p{\tau(X_i) - \tau_c^*(X_i)}^2 - \frac{1}{n}\sum_{i=1}^n \p{\tau(X_i) - \tau_{u(c,\delta)}^*(X_i)}^2: \\
			&\quad\quad\quad\quad\quad\quad\quad\quad\quad\quad \Norm{\tau - \tau^*_{u(c,\delta)}}_{L_2(\mathcal{P})} \leq \delta +Dn^{-\frac{1}{1-p}}\delta\Bigg\}.
			\end{align*}
			Let the two summands be $Z_{1,c,\delta}, Z_{2,c,\delta}$ respectively. Starting with the former, we note that for all $c,\delta > 0$, 
			$$ Z_{1,c,\delta} \leq  \, Z_{1,u(c,\delta),2^{\lceil \log_2(\delta) \rceil}}, $$
			and so it suffices to bound this quantity on a set with $c \in \mathcal{C}^{k_c, \delta}$, with $\delta = 4M \cdot 2^{-k_\delta}$ for $k_c, \, k_\delta = 0, \, 1, \, 2, \, ...$. Applying \eqref{eq:G_tala} unconditionally with probability threshold $\varepsilon \propto 2^{-k_c - k_\delta}n^{-\frac{1}{1-p}}2^{-k_\delta} = 2^{-k_c}2^{-2k_\delta}n^{-\frac{1}{1-p}}$ in \eqref{eq:G_tala}, we can use a union bound to check that
			\begin{align*}
			Z_{1,c,\delta} 
			&= \oo_P\Bigg\{ \delta^2 (1+n^{-\frac{1}{1-p}})^2 + c^{2p} \delta^{2(1-p)} (1+n^{-\frac{1}{1-p}})^{2(1-p)} \frac{\log(n)}{\sqrt{n}}\\
			&\quad\quad\quad\quad +  c^{p}\delta^{2-p}(1+n^{-\frac{1}{1-p}})^{2-p}\frac{1}{\sqrt{n}}\sqrt{\log\p{\frac{cn^{1/(1-p)}}{\delta^2}}}\\
			&\quad\quad\quad\quad + \frac{1}{n}c^{2p}\delta^{2(1-p)}(1+n^{-\frac{1}{1-p}})^{2(1-p)}\log\p{\frac{cn^{1/(1-p)}}{\delta^2}}
			\Bigg\}\\
			&= \oo_P\Bigg\{ \delta^2 + c^{2p} \delta^{2(1-p)} \frac{\log(n)}{\sqrt{n}} +  c^{p}\delta^{2-p}\frac{1}{\sqrt{n}}\sqrt{\log\p{\frac{cn^{1/(1-p)}}{\delta^2}}} + \frac{1}{n}c^{2p}\delta^{2(1-p)}\log\p{\frac{cn^{1/(1-p)}}{\delta^2}}
			\Bigg\}
			\end{align*}
			simultaneously for all $\tau \in\hh_{u(c,\delta)}$ such that $\Norm{\tau - \tau^*_{u(c,\delta)}}_{L_2(\mathcal{P})} \leq \delta $, for all $c > 1$ and $\delta \leq 4M$.
			
			Next, to bound $Z_{2,c,\delta}$, 
			by Cauchy-Schwartz, 
			\begin{align*}
			&\frac{\sum_{\cb{i : q(i) = q}}\p{ \tau(X_i)-\tau_c^*(X_i)}^2 -\p{ \tau(X_i)-\tau_{u(c,\delta)}^*(X_i)}^2} {\abs{\cb{i : q(i) = q}}}  \\
			&\quad\quad\quad\quad = \frac{\sum_{\cb{i : q(i) = q}}  2\p{\tau_{u(c,\delta)}^*(X_i) - \tau(X_i)}\p{\tau_c^*(X_i) - \tau_{u(c,\delta)}^*}}{\abs{\cb{i : q(i) = q}}}  \\
			&\quad\quad\quad\quad\quad\quad + \frac{\sum_{\cb{i : q(i) = q}}  \p{\tau_c^*(X_i) - \tau_{u(c,\delta)}^*(X_i)}^2 }{\abs{\cb{i : q(i) = q}}}  \\
			&\quad\quad\quad\quad \leq2\Norm{\tau_{u(c,\delta)}^*(X_i) - \tau(X_i)}_\infty \Norm{\tau_c^*(X_i) - \tau_{u(c,\delta)}^*}_\infty +  \Norm{\tau_c^*(X_i) - \tau_{u(c,\delta)}^*}_\infty^2  \\
			&\quad\quad\quad\quad = \oo \p{c^p\delta^{1-p}\frac{c^p\delta^{1-p}}{n}+ \frac{c^{2p}\delta^{2-2p}}{n^2}}. \\
			&\quad\quad\quad\quad = \oo \p{\frac{c^{2p}\delta^{2(1-p)}}{n}}. 
			\end{align*}
			where the second to the last equality follows from \eqref{eq:c-uc-l2p} and \eqref{eq:MN_bound}. Note that this is a deterministic bound, so it holds for all $c \geq 1$. The desired result then follows.
		\end{proof}
	\end{lemma}
	
	\begin{lemma}
		\label{lemm:cross-fit}
		Suppose that the propensity estimate $\he(x)$ is uniformly consistent, 
		\begin{equation*}
		\xi_n := \sup_{x \in \mathcal{X}} \abs{\hat{e}(x) - e^*(x)} \to_p 0,
		\end{equation*}
		and the $L_2$ errors converge at rate
		\begin{equation}
		\label{eq:an}
		E\Big[\big\{\hat{m}(X) - m^*(X)\big\}^2\Big], E\Big[\big\{\hat{e}(X) - e^*(X)\big\}^2\Big]= \oo\p{a_n^2}
		\end{equation}
		for some sequence $a_n \rightarrow 0$. Suppose, moreover, that we have overlap, i.e.,
		$\eta < e^*(x) < 1 - \eta$ for some $\eta > 0$, and that
		Assumptions \ref{assu:rkhs} and \ref{assu:approx} hold. 	Then, for any $\varepsilon > 0$, there exists a constant
		$U(\varepsilon)$ such that the regret functions induced by the oracle learner \eqref{eq:oracle_kernel}
		and the feasible learner \eqref{eq:main_kernel} are coupled with probability at least $1 - \varepsilon$ as
		\begin{equation}
		\label{eq:dml_error_shortened}
		\begin{split}
		&\abs{\hR_n(\tau; \, c) - \tR_n(\tau; \, c)} \leq U(\varepsilon) \Bigg\{c^p R(\tau; \, c)^{(1 - p)/2} a_n^2 
		+ c^{2p}R(\tau; \, c)^{1 - p} \frac{1}{\sqrt{n}}\log(n) \\
		&\quad\quad\quad  + c^{2p}R(\tau; \, c)^{1-p}\frac{1}{n} \log\p{\frac{cn^{1/(1-p)}}{R(\tau; \, c)}}
		+  c^{p}R(\tau; \, c)^{1-\frac{ p}{2}}\frac{1}{\sqrt{n}}\sqrt{\log\p{\frac{cn^{1/(1-p)}}{R(\tau; \, c)}}} \\
		&\quad\quad\quad\quad\quad\quad   + c^p R(\tau; \, c)^{(1-p)/2} a_n \frac{1}{\sqrt{n}} \sqrt{\log\p{\frac{cn^{1/(1-p)}}{R(\tau; \, c)}}}   + \xi_n R(\tau; \, c)  
		\Bigg\},
		\end{split}
		\end{equation}
		simultaneously for all $1 \leq c \leq c_n \log(n)$ with $c_n = n^{\alpha/(p+1-2\alpha)}$ and $\tau \in \hh_c$.
		
		\begin{proof}
			
			We start by decomposing the feasible loss function \smash{$\hL(\tau)$} as follows:
			\begin{align*}
			\hL(\tau)
			&= \frac{1}{n} \sum_{i=1}^n \Big[\Big\{Y_i - \hat{m}^{(-q(i))}(X_i)\Big\} - \tau(X_i) \Big\{W_i - \hat{e}^{(-q(i))}(X_i)\Big\}\Big]^2 \\
			&=\frac{1}{n} \sum_{i=1}^n \Big[\Big\{Y_i - m^*(X_i)\Big\} + \Big\{m^*(X_i)-\hat{m}^{(-q(i))}(X_i)\Big\} \\
			&\quad\quad\quad\quad - \tau(X_i) \Big\{W_i - e^*(X_i)\Big\} -\tau(X_i) \Big\{e^*(X_i) - \hat{e}^{(-q(i))}(X_i)\Big\}\Big]^2 \\
			&= \frac{1}{n} \sum_{i=1}^n \Big[\Big\{Y_i - m^*(X_i)\Big\} - \tau(X_i) \Big\{W_i - e^*(X_i)\Big\}\Big]^2 \\
			&\quad\quad+ \frac{1}{n} \sum_{i=1}^n \Big[\Big\{m^*(X_i)-\hat{m}^{(-q(i))}(X_i)\Big\} -\tau(X_i) \Big\{e^*(X_i) - \hat{e}^{(-q(i))}(X_i)\Big\}\Big]^2 \\
			&\quad\quad+ \frac{1}{n} \sum_{i=1}^n2\Big\{Y_i - m^*(X_i)\Big\}\Big\{m^*(X_i)-\hat{m}^{(-q(i))}(X_i)\Big\}\\
			&\quad\quad- \frac{1}{n} \sum_{i=1}^n2\Big\{Y_i - m^*(X_i)\Big\}\Big\{e^*(X_i)-\hat{e}^{(-q(i))}(X_i)\Big\}\tau(X_i)\\
			&\quad\quad- \frac{1}{n} \sum_{i=1}^n2\Big\{W_i - e^*(X_i)\Big\}\Big\{m^*(X_i)-\hat{m}^{(-q(i))}(X_i)\Big\}\tau(X_i)\\
			&\quad\quad+  \frac{1}{n} \sum_{i=1}^n2\Big\{Y_i - m^*(X_i)\Big\}\Big\{e^*(X_i)-\hat{e}^{(-q(i))}(X_i)\Big\}\tau(X_i)\\
			&\quad\quad- \frac{1}{n} \sum_{i=1}^n2\Big\{W_i - e^*(X_i)\Big\}\Big\{e^*(X_i)-\hat{e}^{(-q(i))}(X_i)\Big\}\tau(X_i).
			\end{align*}
			Furthermore, we can verify that some terms cancel out when we restrict attention
			to our main object of interest
			\smash{$\hR(\tau; \, c) - \tR(\tau; \, c) = \hL(\tau) - \hL(\tau^*_c) - \tL(\tau) + \tL(\tau^*_c)$};
			in particular, note that the first summand above is exactly \smash{$\tL(\tau)$}:
			\begin{align*}
			&\hR(\tau; \, c) - \tR(\tau; \, c) \\
			&\quad\quad= \frac{-2}{n} \sum_{i=1}^n \Big\{m^*(X_i)-\hat{m}^{(-q(i))}(X_i)\Big\} \Big\{e^*(X_i) - \hat{e}^{(-q(i))}(X_i)\Big\} \Big\{\tau(X_i) - \tau_c^*(X_i)\Big\} \\
			&\quad\quad\quad\quad + \frac{1}{n} \sum_{i=1}^n \Big\{e^*(X_i) - \hat{e}^{(-q(i))}(X_i)\Big\}^2 \Big\{\tau(X_i)^2 - \tau_c^*(X_i)^2\Big\} \\
			&\quad\quad\quad\quad- \frac{1}{n} \sum_{i=1}^n 2\Big\{Y_i - m^*(X_i)\Big\}\Big\{e^*(X_i)-\hat{e}^{(-q(i))}(X_i)\Big\}  \Big\{\tau(X_i) - \tau_c^*(X_i)\Big\} \\
			&\quad\quad\quad\quad- \frac{1}{n} \sum_{i=1}^n 2\Big\{W_i - e^*(X_i)\Big\}\Big\{m^*(X_i)-\hat{m}^{(-q(i))}(X_i)\Big\}\Big\{\tau(X_i) - \tau_c^*(X_i)\Big\}\\
			&\quad\quad\quad\quad+ \frac{1}{n} \sum_{i=1}^n 2\Big\{W_i - e^*(X_i)\Big\}\Big\{e^*(X_i)-\hat{e}^{(-q(i))}(X_i)\Big\}\Big\{\tau(X_i)^2 - \tau_c^*(X_i)^2\Big\}.
			\end{align*}
			Let $A^c_1(\tau)$, $A^c_2(\tau)$, $B^c_1(\tau)$, $B^c_2(\tau)$ and $B^c_3(\tau)$ denote these 5 summands respectively.
			We now proceed to bound them, each on their own.
			
			Starting with $A^c_1(\tau)$, by Cauchy-Schwarz,
			$$ \abs{A^c_1(\tau)} \leq 2 \sqrt{\frac{1}{n} \sum_{i=1}^n \Big\{m^*(X_i)-\hat{m}^{(-q(i))}(X_i)\Big\}^2}
			\sqrt{\frac{1}{n} \sum_{i=1}^n \Big\{e^*(X_i)-\hat{e}^{(-q(i))}(X_i)\Big\}^2} \Norm{\tau - \tau^*_c}_\infty. $$
			This inequality is deterministic, and so trivially holds simultaneously across all $\tau \in \hh_c$.
			Now, the two square-root terms denote the mean-squared errors of the $m$- and $e$-models respectively, and
			decay at rate $\oo_P(a_n)$ by Assumption \ref{assu:approx} and a direct application of Markov's inequality.
			Thus, applying \eqref{eq:infty_bound} to bound the infinity-norm discrepancy between
			$\tau$ and $\tau^*_c$, we find that simultaneously for all $c \geq 1$,
			\begin{equation*}
			\sup\cb{c^{-p} R(\tau; \, c)^{-\frac{1 - p}{2}} \abs{A^c_1(\tau)}: \tau \in \hh_c, c\geq 1} = \oo_P\p{a_n^2}.
			\end{equation*}
			To bound $A^c_2(\tau)$, note that 
			\begin{align}
			\tau^2(X_i)-\tau_c^*(X_i)^2 
			= 2\tau_c^*(X_i) [\tau(X_i)-\tau_c^*(X_i)] + (\tau(X_i)-\tau_c^*(X_i))^2  \label{eq:square-decomp} 
			\end{align}
			and so,
			\begin{align*}
			\abs{A_2^c(\tau)}
			&\leq  2 \Norm{\tau - \tau^*_c}_\infty\Norm{\tau_c^*}_\infty {\frac{1}{n} \sum_{i=1}^n \p{e^*(X_i)-\hat{e}^{(-q(i))}(X_i)}^2} \\
			&\quad\quad\quad\quad\quad\quad + \Norm{\tau-\tau_c^*}_\infty^2 \frac{1}{n} \sum_{i=1}^n \p{e^*(X_i)-\hat{e}^{(-q(i))}(X_i)}^2 \\
			&= A_{2,1}^c(\tau) + A_{2,3}^c(\tau).
			\end{align*}
			To bound the two terms above, we can use a similar argument to the one used to bound $A_1^c(\tau)$. Specifically, $\frac{1}{n} \sum_{i=1}^n \p{e^*(X_i)-\hat{e}^{(-q(i))}(X_i)}^2$ is bounded with high probability and does not depend on $c$ or $\tau$, whereas terms that depend on $c$ or $\tau$ are deterministically bounded via \eqref{eq:infty_bound}; also, recall that $\Norm{\tau^*_c}_\infty \leq 2M$ by \eqref{eq:Hc}. We thus find that
			\begin{align*}
			&\sup\cb{c^{-p} R(\tau; \, c)^{-\frac{1 - p}{2}} \abs{A^c_{2,1}(\tau)}: \tau \in \hh_c, c\geq 1} = \oo_P\p{a_n^2}.\\
			&\sup\cb{c^{-2p} R(\tau; \, c)^{-(1 - p)} \abs{A^c_{2,2}(\tau)}: \tau \in \hh_c, c\geq 1} = \oo_P\p{a_n^2},
			\end{align*}
			which all in fact decay at the desired rate.
			
			We now move to bounding $B_1^c(\tau)$. To do so, first define
			\begin{align*}
			&B_{1,q}^c(\tau) = \frac{\sum_{\cb{i : q(i) = q}} 2\Big\{Y_i - m^*(X_i)\Big\}\Big\{e^*(X_i)-\hat{e}^{(-q(i))}(X_i)\Big\}  \Big\{\tau(X_i) - \tau_c^*(X_i)\Big\}}{\abs{\cb{i : q(i) = q}}},
			\end{align*}
			and note that $\abs{B_1^c(\tau)} \leq \sum_{q=1}^Q \abs{B_{1,q}^{c}(\tau)}$. We first bound $\sup B_{1,q}^c(\tau)$. To proceed, we bound this quantity over sets indexed by $c$ and $\delta$ such that \smash{$\Norm{\tau - \tau^*_c}_{L_2(\mathcal{P})} \leq \delta$},
			i.e., we bound
			\begin{align*}
			\sup_{\tau \in \hh_{c} } \cb{ B_{1,q}^c(\tau) : \Norm{\tau - \tau^*_c}_{L_2(\mathcal{P})} \leq \delta }.
			\end{align*}
			Let $\mathcal{I}^{(-q)} = \cb{X_i, \, W_i, \, Y_i : q(i) \neq q}$ be the set of data points excluded in the $q$-th fold.
			By cross-fitting, 
			\begin{align*}
			&E\{B^{c}_{1,q}(\tau) \cond \mathcal{I}^{(-q)}\} \\
			&\ \ \ \ = \sum_{\cb{i : q(i) = q}} E\Bigg[\frac{2\{Y_i - m^*(X_i)\}\{e^*(X_i)-\hat{e}^{(-q(i))}(X_i)\}  \{\tau(X_i) - \tau_c^*(X_i)\}}{\abs{\cb{i : q(i) = q}}}\cond \mathcal{I}^{(-q)}\Bigg] \\
			&\ \ \ \ = \sum_{\cb{i : q(i) = q}} \EE{E\Bigg[\frac{2\{Y_i - m^*(X_i)\}\{e^*(X_i)-\hat{e}^{(-q(i))}(X_i)\}  \{\tau(X_i) - \tau_{c}^*(X_i)\}}{\abs{\cb{i : q(i) = q}}}\cond \mathcal{I}^{(-q)}, X_i\Bigg] \cond \mathcal{I}^{(-q)}}\\
			&\ \ \ \ = \sum_{\cb{i : q(i) = q}} E\Bigg[\frac{2\{e^*(X_i)-\hat{e}^{(-q(i))}(X_i)\}  \{\tau(X_i) - \tau_c^*(X_i)\}}{\abs{\cb{i : q(i) = q}}}E[\{Y_i - m^*(X_i)\}\cond X_i] \cond \mathcal{I}^{(-q)}\Bigg]\\
			&\ \ \ \ =  0, 
			\end{align*}
			where the last equation follows because $\EE{\p{Y_i - m^*(X_i)}\cond X_i} = 0$ by definition.
			Moreover, by conditioning on $\mathcal{I}^{(-q)}$, the summands in $B_{1,q}^c(\tau)$ become independent, as $\hat{e}^{(-q(i))}(X_i)$ is now only random in $X_i$. 
			By Lemma \ref{lemm:chaining} and \eqref{eq:an},
			we can bound the expectation of the supremum of this term as 
			\begin{align*}
			\frac{\EE{\sup_{\tau \in \hh_{c} } \cb{ B_{1,q}^c(\tau) : \Norm{\tau - \tau^*_c}_{L_2(\mathcal{P})} \leq \delta } \cond \mathcal{I}^{(-q)}}}{E\Big[\Big\{e^*(X)-\hat{e}^{(-q)}(X)\Big\}^2\Big]^{1/2}}
			= \oo \Bigg\{c^p \delta^{1-p} \frac{\log(n)}{\sqrt{n}}\Bigg\},
			\end{align*}
			and so, in particular,
			\begin{equation}
			\label{eq:B_1_chaining}
			\EE{\sup_{\tau \in \hh_{c} } \cb{ B_{1,q}^c(\tau) : \Norm{\tau - \tau^*_c}_{L_2(\mathcal{P})} \leq \delta } \cond \mathcal{I}^{(-q)}}
			= \oo_P\Bigg\{a_n c^p \delta^{1-p} \frac{\log(n)}{\sqrt{n}}\Bigg\}.
			\end{equation}
			It now remains to bound stochastic fluctuations of this supremum; and we do so using
			Talagrand's concentration inequality \eqref{eq:talagrand-1}. To proceed, first note that for an absolute constant $B$, 
			\begin{align*}
			\sup_{\tau \in \hh_{c} } \cb{\Norm{2\{Y_i - m^*(\cdot)\}\{e^*(\cdot)-\hat{e}(\cdot)\}  \{\tau(\cdot) - \tau_c^*(\cdot)\}}_\infty : \Norm{\tau - \tau^*_c}_{L_2(\mathcal{P})} \leq \delta} \leq Bc^p\delta^{1-p},
			\end{align*}
			and for a different constant $B$,
			\begin{align*}
			&\sup_{\tau \in \hh_{c} } \cb{\EE{\Bigg[2\Big\{Y_i - m^*(X_i)\Big\}\Big\{e^*(X_i)-\hat{e}^{(-q(i))}(X_i)\Big\} \Big\{\tau(X_i) - \tau_c^*(X_i)\Big\}\Bigg]^2} : \Norm{\tau - \tau^*_c}_{L_2(\mathcal{P})} \leq \delta}\\
			&\quad\quad\quad\quad \leq Bc^{2p}\delta^{2(1-p)}a_n^2.
			\end{align*}
			Following from \eqref{eq:talagrand-1} and \eqref{eq:B_1_chaining}, for any fixed $c, \, \delta, \, \varepsilon > 0$, there exists an (again, different) absolute constant $B$ such that, with probability at least $1-\varepsilon$, 
			\begin{equation}
			\begin{split}
			&\sup_{\tau \in \hh_{c} } \cb{B_{1,q}^c(\tau)\cond \mathcal{I}^{(-q)} : \Norm{\tau - \tau^*_c}_{L_2(\mathcal{P})} \leq \delta} \\
			&\quad\quad\quad\quad < B\Bigg\{c^p \delta^{1-p}a_n \frac{\log(n)}{\sqrt{n}} + \frac{c^p\delta^{1-p}a_n}{\sqrt{n}} \sqrt{\log\p{\frac{1}{\varepsilon}}} + \frac{1}{n} c^p \delta^{1-p}\log\p{\frac{1}{\varepsilon}}\Bigg\} 
			\end{split}
			\label{eq:B_1_c_q_tala}
			\end{equation}
			Because the right-hand side does not depend on \smash{$\mathcal{I}^{(-q)}$}, this bound also holds unconditionally.
			
			Our next step is to establish a bound that holds for all values of $c$ and $\delta$ simultaneously,
			as opposed to single values only as in \eqref{eq:B_1_c_q_tala}. 
			For $k = 0, 1, 2, \cdots$, define
			$$\mathcal{C}^{k,\delta} := \cb{2^k + jn^{-\frac{1}{1-p}}\delta 2^k, j=0,1,2,\cdots, \ceil{ (n^{1/(1-p)}-1)/\delta } }.$$
			For any $c\geq 1$, let $u(c,\delta) = \min\{d: d > c, d \in \mathcal{C}^{2^{\floor{\log_2 {c}}},\delta}\}$. 
			Recall that $\Norm{\tau_c^*}_{L_2(\mathcal{P})} \leq 2M$ by definition \eqref{eq:Hc},
			and so by Lemma \ref{lemm:tau_c_l2p}, there is a constant $D$ such that 
			\begin{align}
			\Norm{\tau^*_c - \tau^*_{u(c,\delta)}}_{L_2(\mathcal{P})} \leq Dn^{-\frac{1}{1-p}}\delta.
			\label{eq:tau_c_u_c_l2p}
			\end{align}
			Thus, for any $c\geq 1$, 
			\begin{align*}
			&\sup_{\tau \in \hh_{c} } \cb{B_{1,q}^c(\tau) : \Norm{\tau - \tau^*_c}_{L_2(\mathcal{P})} \leq \delta} \\
			&\quad\quad\quad\quad\leq \sup_{\tau \in \hh_{u(c,\delta)}  } \cb{ B_{1,q}^c(\tau) : \Norm{\tau - \tau^*_{u(c,\delta)}}_{L_2(\mathcal{P})} \leq \delta + Dn^{-\frac{1}{1-p}}\delta}\\
			&\quad\quad\quad\quad\leq \sup_{\tau \in \hh_{u(c,\delta)} } \cb{ B_{1,q}^{u(c,\delta)}(\tau) :  \Norm{\tau - \tau^*_{u(c,\delta)}}_{L_2(\mathcal{P})} \leq \delta +Dn^{-\frac{1}{1-p}}\delta}\\
			&\quad\quad\quad\quad\quad\quad +\sup_{\tau \in \hh_{u(c,\delta)}} \cb{ B_{1,q}^{c}(\tau) - B_{1,q}^{u(c,\delta)}(\tau) :  \Norm{\tau - \tau^*_{u(c,\delta)}}_{L_2(\mathcal{P})} \leq \delta +Dn^{-\frac{1}{1-p}}\delta}.
			\end{align*}
			Let the two summands be $Z_{1,c,\delta}^{B_1}$ and $Z_{2,c,\delta}^{B_1}$ respectively.
			Starting with the former, we note that for all $c, \, \delta > 0$,
			$$ Z_{1,c,\delta}^{B_1} \leq  \, Z_{1,u(c,\delta),2^{\lceil \log_2(\delta) \rceil}}^{B_1}, $$ 
			and so it suffices to bound this quantity on a set with
			$c\in \mathcal{C}^{k_c,\delta}$ with $\delta = 4M \cdot 2^{-k_\delta}$, for $k_c, \, k_\delta = 0, \, 1, \, 2, \, ...$.
			Applying \eqref{eq:B_1_c_q_tala} unconditionally with probability threshold
			$\varepsilon \propto 2^{-k_c - k_\delta}n^{-\frac{1}{1-p}}2^{-k_\delta} = 2^{-k_c}2^{-2k_\delta} n^{-\frac{1}{1-p}}$,
			we can use a union bound to check that
			\begin{align*}
			Z_{1,c,\delta}^{B_1} 
			&= \oo_P\Bigg[ c^p \Big\{\delta + Dn^{-\frac{1}{1-p}}\delta\Big\}^{1-p}a_n \frac{\log(n)}{\sqrt{n}} + \frac{\Big\{c^p  (\delta + Dn^{-\frac{1}{1-p}}\delta)^{1-p}a_n\Big\}}{\sqrt{n}} \sqrt{\log\p{\frac{cn^{1/(1-p)}}{\delta^2}}} \nonumber\\
			&\quad\quad\quad\quad + \frac{1}{n} c^p \p{\delta+ Dn^{-\frac{1}{1-p}}\delta}^{1-p}\log\p{\frac{cn^{1/(1-p)}}{\delta^2}}\Bigg] \\
			&= \oo_P\Bigg\{c^p \delta^{1-p}a_n \frac{\log(n)}{\sqrt{n}} +  \frac{c^p\delta^{1-p}a_n}{\sqrt{n}} \sqrt{\log\p{\frac{cn^{1/(1-p)}}{\delta^2}}} \nonumber + \frac{1}{n} c^p \delta^{1-p}\log\p{\frac{cn^{1/(1-p)}}{\delta^2}}
			\Bigg\}
			\end{align*}
			simultaneously for all $c > 1$ and $\delta \leq 4M$.
			Next, to bound $Z_{2,c,\delta}^{B_1}$, 
			we use Cauchy-Schwartz to check that
			\begin{align*}
			&\frac{\sum_{\cb{i : q(i) = q}} 2\Big\{Y_i - m^*(X_i)\Big\}\Big\{e^*(X_i)-\hat{e}^{(-q(i))}(X_i)\Big\} \Big \{ \tau_{u(c,\delta)}^*(X_i)-\tau_c^*(X_i)\Big\}}{\abs{\cb{i : q(i) = q}}}  \\
			&\quad\quad\quad\quad \leq D \sqrt{\frac{\sum_{\cb{i : q(i) = q}}\Big\{e^*(X_i)-\hat{e}^{(-q(i))}(X_i)\Big\}^2 }{\abs{\cb{i : q(i) = q}}}} \sqrt{\frac{\sum_{\cb{i : q(i) = q}} \Big\{ \tau_{u(c,\delta)}^*(X_i)-\tau_c^*(X_i)\Big\}^2 }{\abs{\cb{i : q(i) = q}}}} \\
			&\quad\quad\quad\quad \leq  D \sqrt{\frac{\sum_{\cb{i : q(i) = q}}\Big\{e^*(X_i)-\hat{e}^{(-q(i))}(X_i)\Big\}^2 }{\abs{\cb{i : q(i) = q}}}}\Norm{\tau_{u(c,\delta)}^* - \tau_c^*}_\infty\\
			&\quad\quad\quad\quad = \oo_p\p{\frac{a_nc^p\delta^{1-p}}{n}}.
			\end{align*}
			where the last equality follows from \eqref{eq:tau_c_u_c_l2p}, \eqref{eq:MN_bound}  and \eqref{eq:an} with a direct application of Markov's inequality. Note that the term that depends on $c$ is deterministically bounded, so the above bound holds for all $c \geq 1$. We can similarly bound $-B_1^c(\tau)$. For any $T$, $\PP{\sup_{\tau\in\hh_c} \abs{B_1^c(\tau)} \geq T }\leq \PP{\sup_{\tau\in\hh_c} B_1^c(\tau) \geq T} + \PP{\sup_{\tau\in\hh_c} -B_1^c(\tau) \geq T}$, the desired result then follows.
			Similar arguments apply to bounding $B_2^c(\tau)$ as well, and the same bound (up to constants) suffices. 
			
			Now moving to bounding $B_3^c(\tau)$, note that by \eqref{eq:square-decomp},
			\begin{align*}
			&B_3^c(\tau) \leq \frac{4}{n} \sum_{i=1}^n \Big\{W_i - e^*(X_i)\Big\}\Big\{e^*(X_i)-\hat{e}^{(-q(i))}(X_i)\Big\} \Big\{\tau(X_i) - \tau_c^*(X_i)\Big\}\tau_c^*(X_i) \\
			&\quad\quad\quad\quad+  \frac{2}{n} \sum_{i=1}^n \Big\{W_i - e^*(X_i)\Big\}\Big\{e^*(X_i)-\hat{e}^{(-q(i))}(X_i)\Big\}\Big\{\tau(X_i) - \tau_c^*(X_i)\Big\}^2.
			\end{align*}
			Denote the two summands by $D_{1}^{B_3,c}(\tau)$ and $D_{2}^{B_3,c}(\tau)$ respectively. Note that since $\Norm{\tau_c^*}_\infty \leq 2M$, we can use a similar argument to the one used to bound $\sup B_1^c(\tau)$, and the same bound suffices for bounding $\sup_{\tau \in \hh_c} D_1^{B_3, c}(\tau)$ .
			
			We now proceed to bound $D_2^{B_3,c}(\tau)$. First, we note that 
			\begin{align*}
			D_{2}^{B_3,c} (\tau)
			&\leq \frac{\sum_{i=1}^n 2\Norm{Y_i - m^*(\cdot)}_\infty \Norm{e^*(\cdot)-\hat{e}^{(-q(i))}(\cdot)}_\infty  \p{\tau(X_i) - \tau_c^*(X_i)}^2}{n} \\
			&\leq B  \Norm{e^*(\cdot)-\hat{e}^{(-q(i))}(\cdot)}_\infty  \frac{\sum_{i=1}^n  \{\tau(X_i) - \tau_c^*(X_i)\}^2}{n},
			\end{align*}
			where $B$ is an absolute constant. By Lemma \ref{lemm:square_tau_tau_c}, uniformly  for all $\tau\in \mathcal{H}_c, c \geq 1, \delta\leq 4M$ where $\Norm{\tau - \tau^*_c}_{L_2(\mathcal{P})} \leq \delta$,  we have
			\begin{align*}
			\sup_{\tau \in \hh_c} D_{2}^{B_3,c} (\tau) &= \oo_P\Bigg\{ \xi_n\delta^2 +  c^{2p} \delta^{2(1-p)} \frac{\log(n)}{\sqrt{n}} +  c^{p}\delta^{2-p}\frac{1}{\sqrt{n}}\sqrt{\log\p{\frac{cn^{1/(1-p)}}{\delta^2}}}  \nonumber\\
			&\quad\quad\quad\quad\quad + \frac{1}{n}c^{2p}\delta^{2(1-p)}\log\p{\frac{cn^{1/(1-p)}}{\delta^2}}
			+\frac{c^{2p}\delta^{2-2p}}{n}
			\Bigg\},
			\end{align*}
			where $\xi_n =  \Norm{e^*(\cdot)-\hat{e}^{(-q(i))}(\cdot)}_\infty = o(1)$.
			Finally, recalling that from \eqref{eq:R_tau_l2p},
			$R(\tau; \, c)$ is within a constant factor of $\Norm{\tau - \tau^*_c}^2_{L_2(\mathcal{P})}$
			given overlap, we obtain the following:
			Then, for any $\varepsilon > 0$, there is a constant
			$U(\varepsilon)$ such that the regret functions induced by the oracle learner \eqref{eq:oracle_kernel}
			and the feasible learner \eqref{eq:main_kernel} are coupled as
			\begin{equation}
			\label{eq:dml_error}
			\begin{split}
			&\abs{\hR_n(\tau; \, c) - \tR_n(\tau; \, c)} \\
			&\quad\quad \leq U(\varepsilon) \Bigg\{c^p R(\tau; \, c)^{(1 - p)/2} a_n^2 + c^{2p}R(\tau; \, c)^{1 - p} a_n^2+ \frac{1}{n} c^p R(\tau; \, c)^{(1 - p)/2}\log\p{\frac{cn^{1/(1-p)}}{\delta^2}} \\
			&\quad\quad\quad\quad\quad\quad + \frac{1}{n}a_nc^p R(\tau; \, c)^{(1 - p/2} +  c^p R(\tau; \, c)^{(1 - p)/2}a_n \frac{\log(n)}{\sqrt{n}}  +  c^{2p}R(\tau; \, c)^{1-p}\frac{1}{n} \\
			&\quad\quad\quad\quad\quad\quad  
			+ \frac{a_nc^pR(\tau; \, c)^{(1 - p)/2}}{\sqrt{n}} \sqrt{\log\p{\frac{cn^{1/(1-p)}}{\delta^2}}}+ \frac{1}{n}c^{2p}R(\tau; \, c)^{1 - p}\log\p{\frac{cn^{1/(1-p)}}{\delta^2}} \\
			&\quad\quad\quad\quad\quad\quad   + \xi_n R(\tau; \, c)  +  c^{2p} R(\tau; \, c)^{1 - p} \frac{\log(n)}{\sqrt{n}}+  c^{p}R(\tau; \, c)^{1-\frac{ p}{2}}\frac{1}{\sqrt{n}}\sqrt{\log\p{\frac{cn^{1/(1-p)}}{\delta^2}}} 
			\Bigg\},
			\end{split}
			\end{equation}
			simultaneously for all $c \geq 1$ and $\tau \in \hh_c$, with probability at least $1 - \varepsilon$.
			
			Comparing \eqref{eq:dml_error} with \eqref{eq:dml_error_shortened}, we note that given the conditions, all other terms that are omitted in \eqref{eq:dml_error_shortened} are on lower order to the first leading term in \eqref{eq:dml_error_shortened}, and so the desired result follows.
		\end{proof}
	\end{lemma}
	\subsection{Proof of Lemma \ref{lemm:cross-fit-shortened}}
	
	\begin{proof}
		The key step in proving this result is to establish a high probability bound for $\smash{\abs{\hR_n(\tau; \, c) - \tR_n(\tau; \, c)}}$, which was proven in Lemma \ref{lemm:cross-fit} above. Given our result from Lemma \ref{lemm:cross-fit}, the main remaining task is to transform this bound into the desired form shown in \eqref{eq:R_h_R_t_inequality}. In order to do so, we linearize all terms that involve $R(\tau; \,c)$ in \eqref{eq:dml_error_shortened} in Lemma \ref{lemm:cross-fit}, and show that the linear coefficients of $R(\tau; \, c)$ are small, and that the remaining terms are lower order to $\rho_n(c)$ for all $1 \leq c \leq c_n \log(n)$. 
		
		We use the following facts to linearize the terms that involve $R(\tau; \,c)$ in \eqref{eq:dml_error_shortened} in Lemma \ref{lemm:cross-fit}. For any $\gamma_{n}, \, \zeta_{n} > 0$, and $ 0 \leq \nu_\gamma, \, \nu_\zeta < 1-p$, by concavity, 
		\begin{align}
		&R(\tau, \, c)^{(1 - p - \nu_\gamma)/2} \leq \gamma_{n}^{(1 - p - \nu_\gamma)/2} + \frac{1 - p - \nu_\gamma}{2} \gamma_{n}^{-(1 + p + \nu_\gamma)/2}\p{R(\tau; \, c) - \gamma_{n}} \nonumber\\
		&\quad\quad\quad\quad\quad\quad= \frac{1 + p + \nu_\gamma}{2} \gamma_{n}^{(1 - p - \nu_\gamma)/2} +  \frac{1 - p - \nu_\gamma}{2} \gamma_{n}^{-(1 + p + \nu_\gamma)/2} R(\tau; \, c),\label{eq:gamma} \\
		&R(\tau, \, c)^{1 - p - \nu_\zeta} \leq \zeta_{n}^{1 - p -\nu_\zeta} + (1 - p - \nu_\zeta) \zeta_{n}^{-p - \nu_\zeta}\p{R(\tau; \, c) - \zeta_{n}} \nonumber\\
		&\quad\quad\quad\quad\quad\quad= (p + \nu_\zeta) \zeta_{n}^{1 - p - \nu_\zeta} +  (1 - p - \nu_\zeta)\zeta_{n}^{-p - \nu_\zeta} R(\tau; \, c). \label{eq:zeta}
		\end{align}
		
		We now proceed to show \eqref{eq:R_h_R_t_inequality} holds by applying the above bounds with specific choices of $\gamma_n,\, \zeta_n$. First, following from \eqref{eq:R_tau_l2p}, $R(\tau; \, c) < (1-\eta)^2 4M^2 = \oo(1)$. Let $J$ be a constant such that $R(\tau; \, c) < J$. Now we bound each term in \eqref{eq:dml_error_shortened} as follows:
		
		To bound the terms $c^p R(\tau; \, c)^{(1 - p)/2} a_n^2$, 
		let $\gamma_n = \{U(\varepsilon)\}^{2/(1+p)}\p{\frac{1-p}{0.04}}^{2/(1+p)} c^{(2p)/(1+p)}a_n^{4/(1+p)}$. 
		Note that since $a_n = o(n^{-1/4})$, $\gamma_n = o(\rho_n(c))$ for all $c \geq 1$. 
		
		Following from \eqref{eq:gamma}, 
		\begin{align*}
		c^p R(\tau; \, c)^{(1 - p)/2}a_n^2 \leq \frac{1}{U(\varepsilon)}\Big[0.02 R(\tau; \, c) + o\{\rho_n(c)\}\Big].
		\end{align*}
		
		To bound the term $c^{2p}R(\tau; \, c)^{1 - p} \frac{1}{\sqrt{n}}\log(n)$, let $\zeta_n = U(\varepsilon)^{1/p}\p{\frac{1-p}{0.02}}^{1/p}c^2 n^{1/(2p)}\log(n)^{1/p}$. When $c= c_n\log(n)$, 
		\begin{align*}
		\zeta_n 
		&= U(\varepsilon)^{1/p}\p{\frac{1-p}{0.02}}^{1/p} c_n^2\{\log(n)\}^{2+1/p}n^{-1/(2p)}\\
		&= \oo\p{n^{(2\alpha)/(p+1-2\alpha)-1/(2p)}\{\log(n)\}^{2+1/p}} \\
		&\stackrel{(a)}{=} o\p{n^{(2\alpha p)/\{(p+1-2\alpha)(1+p)\} - 1/(1+p)}}\\
		&= o\p{c_n^{(2p)/(1+p)}n^{-1/(1+p)}} = o\Big(\rho_n\{c_n\log(n)\}\Big),
		\end{align*}
		where $(a)$ follows from a few lines of algebra and the assumption that \smash{$2\alpha < 1-p$}.
		Since the exponent on $c$ in $\zeta_n$ is greater than that in $\rho_n(c)$, we can verify that for any \smash{$c \leq c_n\log(n)$}, $\frac{\zeta_n(c)}{\rho_n(c)} \leq \frac{\zeta_n(c_n)}{\rho_n(c_n)} = o(1)$. Following from \eqref{eq:zeta},
		\begin{align*}
		c^{2p}R(\tau; \, c)^{1 - p} \frac{1}{\sqrt{n}}\log(n) \leq \frac{1}{U(\varepsilon)}(0.02 R(\tau; \, c) + o(\rho_n(c))).
		\end{align*}

		To bound the term $c^{2p}R(\tau; \, c)^{1-p}\frac{1}{n} \log\p{\frac{cn^{1/(1-p)}}{R(\tau; \, c)}}$, since $R(\tau; \, c)^{\nu_\zeta}\log(1/R(\tau; \, c)) < R(\tau; \, c)^{\nu_\zeta} < J^{\nu_\zeta} = \oo(1)$, and $\log(cn^{1/(1-p)}) =\log(c) + \log(n^{1/(1-p)}) < \frac{2}{1-p} \log(c) \log(n)$, it is sufficient to bound $c^{2p}R(\tau; \, c)^{1-p-\nu_\zeta}\frac{1}{n} \log(n)\log(c)$ for some $0 < \nu_\zeta < 1-p$.
		Let a different $\zeta_n = \Big\{\frac{2}{1-p}J^{\nu_\zeta}U(\varepsilon)\Big\}^{1/(p+\nu_\zeta)}\p{\frac{1-p-\nu_\zeta}{0.02}}^{1/(p+\nu_\zeta)}c^{(2p)/(p+\nu_\zeta)} n^{-\frac{1}{p+\nu_\zeta}}\log(n)^{1/(p+\nu_\zeta)}\log(c)^{1/(p+\nu_\zeta)}$. 
		When $c= c_n\log(n)$, 
		\begin{align*}
		\zeta_n 
		&= \oo\p{n^{(2\alpha p)/\{(p+1-2\alpha)(p+\nu_\zeta)\}} n^{-1/(p+\nu_\zeta)}\log(n)^{2/(p+\nu_\zeta)+(2p)/(p+\nu_\zeta)}}\\
		&\stackrel{(a)}{=} o\p{n^{(2\alpha p)/\{(p+1-2\alpha)(1+p)\} - 1/(1+p)}}\\
		&= o\p{c_n^{(2p)/(1+p)}n^{-1/(1+p)}} = o\Big(\rho_n\{c_n\log(n)\}\Big),
		\end{align*}
		where $(a)$ follows from \smash{$2\alpha < 1$}.
		Since the exponent on $c$ in $\zeta_n$ is greater than that in $\rho_n(c)$, we can verify that for any \smash{$c \leq c_n\log(n)$}, $\frac{\zeta_n(c)}{\rho_n(c)} \leq \frac{\zeta_n(c_n)}{\rho_n(c_n)} = o(1)$. Following from \eqref{eq:zeta},
		\begin{align*}
		c^{2p}R(\tau; \, c)^{1-p-\nu_\zeta}\frac{1}{n} \log(n) \log(c) \leq \frac{1-p}{2J^{\nu_\zeta}U(\varepsilon)}\Big[0.02 R(\tau; \, c) + o\{\rho_n(c)\}\Big].
		\end{align*}
		Thus, 
		\begin{align*}
		c^{2p}R(\tau; \, c)^{1-p}\frac{1}{n} \log\Bigg\{\frac{cn^{1/(1-p)}}{R(\tau; \, c)}\Bigg\} \leq \frac{1}{U(\varepsilon)}\Big[0.02 R(\tau; \, c) + o\{\rho_n(c)\}\Big].
		\end{align*}

		To proceed, let a different $$\gamma_n = \Big\{\frac{2}{\sqrt{1-p}}J^{1/2}U(\varepsilon)\Big\}^{2/(1+p)}\p{\frac{1-p}{0.04}}^{2/(1+p)} c^{(2p)/(1+p)}n^{-1/(1+p)}\{\log(n)\}^{1/(1+p)}\{\log(c)\}^{1/(1+p)}.$$
		Note that for $1 \leq c \leq c_n\log(n)$, 
		$\{\log(c)\}^{1/(1+p)} \leq [\log\{c_n \log(n)\}]^{1/(1+p)}$. For a different constant $D$ and $D'$,
		\begin{align*} 
		\gamma_n &\leq D c^{(2p)/(1+p)}n^{-1/(1+p)} \{\log(n)\}^{1/(1+p)} [\log\{c_n \log(n)\}]^{1/(1+p)}\\
		&\leq D' c^{(2p)/(1+p)}n^{1/(1+p)} \{\log(n)\}^{2/(1+p)}\\
		&= o\{\rho_n(c)\}.
		\end{align*}
		Following from \eqref{eq:gamma}, 
		\begin{align*}
		c^p R(\tau; \, c)^{(1 - p)/2} \frac{1}{\sqrt{n}} \sqrt{\log(n)} \sqrt{\log(c)} \leq \frac{\sqrt{1-p}}{2J^{1/2}U(\varepsilon)}\{0.02 R(\tau; \, c) + o(\rho_n(c))\}.
		\end{align*}
		Thus, 
		\begin{align*}
		c^{p}R(\tau; \, c)^{1-p/2}\frac{1}{\sqrt{n}}\sqrt{\log\p{\frac{cn^{1/(1-p)}}{R(\tau; \, c)}}} \leq \frac{1}{U(\varepsilon)}\p{0.02 R(\tau; \, c) + o(\rho_n(c))}.
		\end{align*}
		
		To bound the term $c^p R(\tau; \, c)^{(1-p)/2} a_n \frac{1}{\sqrt{n}} \sqrt{\log\p{\frac{cn^{1/(1-p)}}{R(\tau; \, c)}}}$, since 
		\begin{align*}
		R(\tau; \, c)^{\nu_\gamma/2}\sqrt{\log(1/R(\tau; \, c))} &< R(\tau; \, c)^{\nu_\gamma/2} \\
		&< J^{\nu_\gamma/2} = \oo(1),
		\end{align*}
		and 
		\begin{align*}
		\sqrt{\log(cn^{1/(1-p)})} 
		&= \sqrt{\log(c) + \frac{1}{1-p}\log(n)} \\
		&< \sqrt{\log(c)} + \frac{1}{\sqrt{1-p}}\sqrt{\log(n)}\\
		&< \frac{2}{\sqrt{1-p}}\sqrt{\log(c)}\sqrt{\log(n)},
		\end{align*}
		and $a_n=o(n^{-1/4})$, it is sufficient to bound $c^p R(\tau; \, c)^{(1-p-\nu_\gamma)/2} n^{-3/4} \sqrt{\log(n)}\sqrt{\log(c)}$ for some $\nu_\gamma$ such that $0 < \nu_\gamma < 1-p$. 
		Let a different 
		\begin{align*}
		\gamma_n &= \Big\{\frac{2}{\sqrt{1-p}}J^{\nu_\gamma/2}U(\varepsilon)\Big\}^{2/(1+p+\nu_\gamma)}\p{\frac{1-p-\nu_\gamma}{0.04}}^{2/(1+p+\nu_\gamma)}\\
		&\ \ \ \ \ \ \ \ c^{(2p)/(1+p+\nu_\gamma)}n^{-3/\{2(1+p+\nu_\gamma)\}}\{\log(n)\log(c)\}^{1/(1+p+\nu_\gamma)}.
		\end{align*}
		Let $\nu_\gamma = (1-p)/2$, it is straightforward to check that $\gamma_n = o(\rho_n(c))$ for all $c\geq 1$. 
		Following from \eqref{eq:gamma}, 
		\begin{align*}
		c^p R(\tau; \, c)^{(1-p-\nu_\gamma)/2} n^{-3/4} \sqrt{\log(n)}\sqrt{\log(c)} \leq \frac{\sqrt{1-p}}{2J^{\nu_\gamma/2}U(\varepsilon)}\{0.02 R(\tau; \, c) + o(\rho_n(c))\}.
		\end{align*}
		Thus,
		\begin{align*}
		c^p R(\tau; \, c)^{(1-p)/2} a_n \frac{1}{\sqrt{n}} \sqrt{\log\p{\frac{cn^{1/(1-p)}}{R(\tau; \, c)}}} \leq \frac{1}{U(\varepsilon)}\{0.02 R(\tau; \, c) + o(\rho_n(c))\}.
		\end{align*}

		Finally, to bound the term $\xi_n R(\tau; \, c)$, note that since $\xi_n \to 0$, for $n$ large enough, $\xi_n R(\tau; \, c) \leq \frac{1}{U(\varepsilon)}0.025 R(\tau; \, c)$.
		
		Given the above derivations, \eqref{eq:R_h_R_t_inequality} is now immediate.
	\end{proof}

	\subsection{Proof of Theorem \ref{theo:rlearn}}
	As discussed earlier, the arguments of \citet{mendelson2010regularization} can be used to get regret bounds
	for the oracle learner. In particular, we note that the $R-$learning objective can be written as a weighted regression problem: $\hat{\tau}(x) = \argmin_{\tau \in \hh_c}\frac{1}{n} \sum_{i=1}^n \{W_i- e^{-(i)}(X_i)\}^2 \Big\{\frac{Y_i - m^{(-i)}(X_i)}{W_i - e^{-(i)}(X_i)} - \tau(X_i)\Big\}^2$. To adapt the setting in \citet{mendelson2010regularization} to our setting, note that we weight the data generating distribution of \smash{$\{X_i, Y_i, W_i\}$} by the weights \smash{$\{W_i - e^{(-i)}(X_i)\}^2$}. In addition, by Lemma \ref{lemm:hiercharchy}, the class of functions we consider $\hh_c$ with capped infinity norm is also an ordered, parameterized hierarchy, thus their results follow. In order to extend their results, we first review their analysis briefly.
	Their results imply the following facts (details see Theorem A and the proof of Theorem 2.5 in the Appendix section in \citet{mendelson2010regularization}). For any $\varepsilon > 0$, there is a constant $U(\varepsilon)$ such that
	\begin{equation}
	\label{eq:trho}
	\rho_n(c) = U(\varepsilon) \{1 + \log(n) + \log\log\p{c + e^1}\} \Bigg\{\frac{\p{c + 1}^p \log(n)}{\sqrt{n}}\Bigg\}^{2/(1 + p)}
	\end{equation}
	satisfies, for large enough $n$ with probability at least $1 - \varepsilon$, simultaneously for all $c \geq 1$, the condition
	\begin{equation}
	\label{eq:oracle_isomoprhism2}
	0.5 \tR_{n}(\tau; \, c) - \rho_n(c) \leq R(\tau; \, c) \leq 2 \tR_n(\tau; \, c) + \rho_n(c).
	\end{equation}
	Thus, thanks to Lemma \ref{lemm:bmn} and \eqref{eq:approx_err}, we know that
	\begin{equation}
	\label{eq:bmn_plgin}
	R(\ttau) \leq \oo_P[\{L(\tau_{c_n}^*) - L(\tau^*)\} + \rho_n(c_n)] \ \with \ c_n = n^{\alpha/\{p + (1 - 2\alpha)\}},
	\end{equation}
	and then pairing \eqref{eq:approx} with the form of \smash{$\rho_n(c)$} in \eqref{eq:trho},
	we conclude that 
	\begin{equation}
	\label{eq:regret_target}
	R(\ttau) \lesssim_P \max\cb{L(\tau_{c_n}^*) - L(\tau^*), \, \rho_n(c_n)} = \too\p{n^{-(1 - 2\alpha)/\{p + (1 - 2\alpha)\}}}.
	\end{equation}
	Our present goal is to extend this argument to get a bound for \smash{$R(\htau)$}.
	
	
	First, Lemma \ref{lemm:cross-fit-shortened} implies that 
	\begin{align*}
	R(\tau; \, c) 
	&\leq 2 \tR_n(\tau; \, c) + \rho_n(c) \nonumber\\
	&\leq 2 \hR_n(\tau; \, c) + 0.25 R(\tau; \, c) + 2\rho_n(c),
	\end{align*}
	which implies that 
	\begin{align*}
	R(\tau; \, c) 
	&\leq \frac{2}{0.75}\hR_n(\tau; \, c) + 2\rho_n(c) \nonumber\\
	&\leq 3\hR_n(\tau; \, c) + 2\rho_n(c)
	\end{align*}
	for large $n$ for all $1 \leq c\leq c_n\log(n)$, with probabilty at least $1-2\varepsilon$. Following a symmetrical argument, \eqref{eq:R_h_R_t_inequality} would imply that 
	\begin{align}
	\frac{1}{3}\hR_n(\tau; \, c) - 2\rho_n(c) \leq R(\tau; \, c) \leq 3\hR_n(\tau; \, c) + 2\rho_n(c) \label{eq:R_hat_isomorphism}
	\end{align}
	for $n$ large enough for all $1 \leq c\leq c_n\log(n)$ with probability at least $1-4\varepsilon$.
	
	Then applying the same argument as above, we use Lemma \ref{lemm:bmn} to check that the constrained estimator defined as 
	\begin{align}
	\bar{\htau} &\in \argmin_{\tau \in \hh} \cb{\hL_n(\tau) + 2 \kappa_1 \rho_n\p{\Norm{\tau}_\hh} : \Norm{\tau}_\hh \leq \log(n) c_n, \, \Norm{\tau}_\infty \leq 2M} \nonumber\\
	&\subseteq \argmin_{\tau \in \hh} \cb{\hR_n(\tau) + 2 \kappa_1 \rho_n\p{\Norm{\tau}_\hh} : \Norm{\tau}_\hh \leq \log(n) c_n, \, \Norm{\tau}_\infty \leq 2M}\label{eq:constr_opt}
	\end{align}
	has regret bounded on the order of
	\begin{equation}
	\label{eq:constr_regret}
	L(\bar{\htau}) - L\p{\tau^*} \lesssim_P \p{\p{L(\tau_{c_n}^*) - L(\tau^*)} + \rho_n(c_n)} \lesssim \rho_n(c_n),
	\end{equation}
	where we note that $\hL_n(\tau) = \hR_n(\tau) + \hL_n(\tau^*)$.
	We see that for some constant $B$ and $B'$, 
	\begin{align}
	&\min_{\tau \in \hh} \cb{\hR(\tau) + 2 \kappa_1 \rho_n\p{\Norm{\tau}_\hh} : \Norm{\tau}_\hh \leq \log(n) c_n, \Norm{\tau}_\infty \leq 2M} \\
	&\quad\quad\quad\quad \leq \hR_n(\tau_{c_n}^*) + 2\kappa_1 \rho_n(c_n) \nonumber\\
	&\quad\quad\quad\quad\stackrel{(a)}{\leq} 3R(\tau_{c_n}^*) + (2 \kappa_1+6) \rho_n(c_n)\ \  w.p.\ 1-4\varepsilon \nonumber\\
	&\quad\quad\quad\quad\stackrel{(b)}{\leq} B c_n^{(2\alpha-1)/\alpha} + (2 \kappa_1+6) \rho_n(c_n)\nonumber \\
	&\quad\quad\quad\quad\stackrel{(c)}{=} B'\rho_n(c_n).\label{eq:r_min}
	\end{align}
	where $(a)$ follows from \eqref{eq:R_hat_isomorphism}, $(b)$
	follows from \eqref{eq:R_tau_l2p} and \eqref{eq:approx}, and $(c)$ follows from \eqref{eq:bmn_plgin} and \eqref{eq:regret_target}. In addition, we see that 
	$$ \inf_{\tau \in \hh} \cb{\hR_n(\tau) + 2 \kappa_1 \rho_n\p{\Norm{\tau}_\hh} : \Norm{\htau}_\hh = \log(n) c_n, \Norm{\tau}_\infty \leq 2M} \gtrsim_P \rho_n\{c_n \log(n)\}$$
	which, combined with \eqref{eq:r_min}, implies that the optimum of the problem
	\eqref{eq:constr_opt} occurs in the interior of its domain (i.e., the constraint is not active).
	Thus, the solution \smash{$\htau$} to the unconstrained problem
	matches \smash{$\bar{\htau}$}, and so \smash{$\htau$} also satisfies \eqref{eq:constr_regret} and hence the regret bound \eqref{eq:regret_target}.
	
	\section{Detailed Simulation Results}
	\label{sec:extrasimu}
	
	For completeness, we include the mean-squared error numbers behind Figure \ref{fig:lasso}-\ref{fig:boost} for the simulations based on lasso, kernel ridge regression and boosting in Section \ref{sec:simu}.
	
	\input{tables/simulation_results_setup_A_lasso.tex}
	\input{tables/simulation_results_setup_B_lasso.tex}
	\input{tables/simulation_results_setup_C_lasso.tex}
	\input{tables/simulation_results_setup_D_lasso.tex}

	\input{tables/simulation_results_setup_A_kernel.tex}
	\input{tables/simulation_results_setup_B_kernel.tex}
	\input{tables/simulation_results_setup_C_kernel.tex}
	\input{tables/simulation_results_setup_D_kernel.tex}

	\input{tables/simulation_results_setup_A_boost.tex}
	\input{tables/simulation_results_setup_B_boost.tex}
	\input{tables/simulation_results_setup_C_boost.tex}
	\input{tables/simulation_results_setup_D_boost.tex}

	\end{appendix}
\end{document}

%% file: tables/simulation_results_setup_A_lasso.tex
\begin{table}[ht]
\centering
\begin{tabular}{cccccccccc}
  \hline
n & d & $\sigma$ & S & T & X & U & R & RS & oracle \\ 
  \hline
500 & 6 & 0.5 & 0.13 & 0.19 & 0.10 & 0.12 & \bf 0.06 & \bf 0.06 & 0.05 \\ 
  500 & 6 & 1 & 0.21 & 0.27 & 0.16 & 0.37 & 0.10 & \bf 0.07 & 0.07 \\ 
  500 & 6 & 2 & 0.27 & 0.35 & 0.25 & 1.25 & 0.21 & \bf 0.12 & 0.19 \\ 
  500 & 6 & 4 & 0.51 & 0.66 & 0.41 & 1.95 & 0.55 & \bf 0.26 & 0.61 \\ 
  500 & 12 & 0.5 & 0.15 & 0.20 & 0.12 & 0.17 & 0.07 & \bf 0.06 & 0.05 \\ 
  500 & 12 & 1 & 0.22 & 0.26 & 0.18 & 0.46 & 0.11 & \bf 0.09 & 0.08 \\ 
  500 & 12 & 2 & 0.30 & 0.35 & 0.26 & 1.18 & 0.23 & \bf 0.14 & 0.23 \\ 
  500 & 12 & 4 & 0.47 & 0.56 & 0.43 & 1.98 & 0.59 & \bf 0.28 & 0.63 \\ 
  1000 & 6 & 0.5 & 0.09 & 0.13 & 0.06 & 0.06 & \bf 0.04 & 0.05 & 0.04 \\ 
  1000 & 6 & 1 & 0.15 & 0.21 & 0.11 & 0.25 & 0.07 & \bf 0.06 & 0.06 \\ 
  1000 & 6 & 2 & 0.23 & 0.29 & 0.20 & 0.85 & 0.13 & \bf 0.08 & 0.11 \\ 
  1000 & 6 & 4 & 0.34 & 0.43 & 0.31 & 2.40 & 0.34 & \bf 0.16 & 0.32 \\ 
  1000 & 12 & 0.5 & 0.11 & 0.14 & 0.08 & 0.11 & \bf 0.05 & \bf 0.05 & 0.04 \\ 
  1000 & 12 & 1 & 0.18 & 0.22 & 0.14 & 0.34 & 0.08 & \bf 0.07 & 0.06 \\ 
  1000 & 12 & 2 & 0.25 & 0.30 & 0.21 & 0.94 & 0.14 & \bf 0.09 & 0.12 \\ 
  1000 & 12 & 4 & 0.33 & 0.40 & 0.29 & 1.95 & 0.35 & \bf 0.18 & 0.33 \\ 
   \hline
\end{tabular}
\caption{\tt Mean-squared error running \texttt{lasso} from Setup A. Results are averaged across 500 runs, rounded to two decimal places, and reported on an independent test set of size $n$.} 
\label{table:setup1-lasso}
\end{table}

%% file: tables/simulation_results_setup_B_lasso.tex
\begin{table}[ht]
\centering
\begin{tabular}{cccccccccc}
  \hline
n & d & $\sigma$ & S & T & X & U & R & RS & oracle \\ 
  \hline
500 & 6 & 0.5 & 0.26 & 0.43 & \bf 0.22 & 0.46 & 0.28 & 0.29 & 0.16 \\ 
  500 & 6 & 1 & 0.44 & 0.66 & \bf 0.38 & 0.83 & 0.43 & 0.72 & 0.33 \\ 
  500 & 6 & 2 & 0.84 & 1.12 & \bf 0.71 & 1.27 & 0.85 & 1.26 & 0.75 \\ 
  500 & 6 & 4 & 1.52 & 1.73 & \bf 1.29 & 1.40 & 1.51 & 1.41 & 1.46 \\ 
  500 & 12 & 0.5 & 0.30 & 0.46 & \bf 0.25 & 0.54 & 0.33 & 0.41 & 0.18 \\ 
  500 & 12 & 1 & 0.52 & 0.71 & \bf 0.43 & 0.90 & 0.50 & 0.95 & 0.38 \\ 
  500 & 12 & 2 & 0.93 & 1.12 & \bf 0.78 & 1.28 & 0.96 & 1.31 & 0.84 \\ 
  500 & 12 & 4 & 1.62 & 1.77 & \bf 1.33 & 1.42 & 1.55 & 1.40 & 1.54 \\ 
  1000 & 6 & 0.5 & 0.14 & 0.24 & \bf 0.13 & 0.24 & 0.15 & 0.15 & 0.10 \\ 
  1000 & 6 & 1 & 0.27 & 0.43 & \bf 0.23 & 0.46 & 0.25 & 0.36 & 0.20 \\ 
  1000 & 6 & 2 & 0.54 & 0.73 & \bf 0.45 & 1.12 & 0.52 & 0.92 & 0.47 \\ 
  1000 & 6 & 4 & 1.06 & 1.31 & \bf 0.92 & 1.34 & 1.07 & 1.34 & 1.06 \\ 
  1000 & 12 & 0.5 & 0.17 & 0.28 & \bf 0.15 & 0.29 & 0.18 & 0.18 & 0.11 \\ 
  1000 & 12 & 1 & 0.30 & 0.45 & \bf 0.26 & 0.55 & 0.30 & 0.52 & 0.23 \\ 
  1000 & 12 & 2 & 0.61 & 0.76 & \bf 0.50 & 1.19 & 0.59 & 1.14 & 0.54 \\ 
  1000 & 12 & 4 & 1.15 & 1.30 & \bf 1.01 & 1.33 & 1.19 & 1.34 & 1.13 \\ 
   \hline
\end{tabular}
\caption{\tt Mean-squared error running \texttt{lasso} from Setup B. Results are averaged across 500 runs, rounded to two decimal places, and reported on an independent test set of size $n$.} 
\label{table:setup2-lasso}
\end{table}

%% file: tables/simulation_results_setup_C_lasso.tex
\begin{table}[ht]
\centering
\begin{tabular}{cccccccccc}
  \hline
n & d & $\sigma$ & S & T & X & U & R & RS & oracle \\ 
  \hline
500 & 6 & 0.5 & 0.18 & 0.80 & 0.18 & 0.53 & 0.05 & \bf 0.02 & 0.01 \\ 
  500 & 6 & 1 & 0.33 & 1.18 & 0.29 & 0.66 & 0.10 & \bf 0.03 & 0.03 \\ 
  500 & 6 & 2 & 0.75 & 1.95 & 0.58 & 1.42 & 0.21 & \bf 0.09 & 0.12 \\ 
  500 & 6 & 4 & 1.68 & 3.13 & 1.24 & 3.56 & 0.64 & \bf 0.26 & 0.51 \\ 
  500 & 12 & 0.5 & 0.18 & 0.88 & 0.19 & 0.55 & 0.08 & \bf 0.03 & 0.01 \\ 
  500 & 12 & 1 & 0.34 & 1.29 & 0.31 & 0.86 & 0.12 & \bf 0.06 & 0.04 \\ 
  500 & 12 & 2 & 0.81 & 2.08 & 0.65 & 1.82 & 0.24 & \bf 0.13 & 0.14 \\ 
  500 & 12 & 4 & 1.79 & 3.28 & 1.43 & 4.02 & 0.62 & \bf 0.33 & 0.58 \\ 
  1000 & 6 & 0.5 & 0.10 & 0.49 & 0.10 & 0.23 & 0.02 & \bf 0.01 & 0.00 \\ 
  1000 & 6 & 1 & 0.19 & 0.73 & 0.17 & 0.34 & 0.03 & \bf 0.01 & 0.01 \\ 
  1000 & 6 & 2 & 0.41 & 1.29 & 0.35 & 0.82 & 0.08 & \bf 0.04 & 0.07 \\ 
  1000 & 6 & 4 & 0.97 & 2.38 & 0.82 & 2.31 & 0.27 & \bf 0.11 & 0.22 \\ 
  1000 & 12 & 0.5 & 0.09 & 0.58 & 0.10 & 0.41 & 0.03 & \bf 0.01 & 0.00 \\ 
  1000 & 12 & 1 & 0.18 & 0.82 & 0.18 & 0.54 & 0.04 & \bf 0.02 & 0.01 \\ 
  1000 & 12 & 2 & 0.43 & 1.40 & 0.37 & 1.21 & 0.11 & \bf 0.05 & 0.05 \\ 
  1000 & 12 & 4 & 1.10 & 2.43 & 0.87 & 3.20 & 0.29 & \bf 0.14 & 0.21 \\ 
   \hline
\end{tabular}
\caption{\tt Mean-squared error running \texttt{lasso} from Setup C. Results are averaged across 500 runs, rounded to two decimal places, and reported on an independent test set of size $n$.} 
\label{table:setup3-lasso}
\end{table}

%% file: tables/simulation_results_setup_D_lasso.tex
\begin{table}[ht]
\centering
\begin{tabular}{cccccccccc}
  \hline
n & d & $\sigma$ & S & T & X & U & R & RS & oracle \\ 
  \hline
500 & 6 & 0.5 & 0.46 & \bf 0.37 & 0.45 & 1.20 & 0.51 & 0.72 & 0.47 \\ 
  500 & 6 & 1 & 0.77 & \bf 0.66 & 0.75 & 1.68 & 0.81 & 1.57 & 0.80 \\ 
  500 & 6 & 2 & 1.32 & \bf 1.23 & 1.29 & 1.81 & 1.43 & 1.79 & 1.42 \\ 
  500 & 6 & 4 & 2.02 & 2.20 & 1.97 & 2.10 & 2.20 & \bf 1.91 & 2.19 \\ 
  500 & 12 & 0.5 & 0.59 & \bf 0.44 & 0.56 & 1.19 & 0.63 & 1.08 & 0.57 \\ 
  500 & 12 & 1 & 0.94 & \bf 0.77 & 0.88 & 1.70 & 0.96 & 1.74 & 0.95 \\ 
  500 & 12 & 2 & 1.47 & \bf 1.38 & 1.45 & 1.84 & 1.59 & 1.81 & 1.59 \\ 
  500 & 12 & 4 & 2.06 & 2.21 & 1.98 & 2.12 & 2.28 & \bf 1.94 & 2.17 \\ 
  1000 & 6 & 0.5 & 0.27 & \bf 0.21 & 0.27 & 0.74 & 0.30 & 0.41 & 0.28 \\ 
  1000 & 6 & 1 & 0.50 & \bf 0.41 & 0.48 & 1.57 & 0.54 & 0.87 & 0.53 \\ 
  1000 & 6 & 2 & 0.93 & \bf 0.79 & 0.91 & 1.76 & 0.97 & 1.74 & 0.99 \\ 
  1000 & 6 & 4 & 1.61 & 1.58 & \bf 1.56 & 1.95 & 1.73 & 1.83 & 1.70 \\ 
  1000 & 12 & 0.5 & 0.35 & \bf 0.26 & 0.34 & 0.76 & 0.38 & 0.55 & 0.36 \\ 
  1000 & 12 & 1 & 0.61 & \bf 0.48 & 0.57 & 1.54 & 0.63 & 1.28 & 0.64 \\ 
  1000 & 12 & 2 & 1.10 & \bf 0.93 & 1.05 & 1.78 & 1.11 & 1.76 & 1.17 \\ 
  1000 & 12 & 4 & 1.76 & 1.73 & \bf 1.68 & 1.94 & 1.82 & 1.83 & 1.84 \\ 
   \hline
\end{tabular}
\caption{\tt Mean-squared error running \texttt{lasso} from Setup D. Results are averaged across 500 runs, rounded to two decimal places, and reported on an independent test set of size $n$.} 
\label{table:setup4-lasso}
\end{table}

%% file: tables/simulation_results_setup_A_kernel.tex
\begin{table}[ht]
\centering
\begin{tabular}{ccccccccc}
  \hline
n & d & $\sigma$ & S & T & X & U & R & oracle \\ 
  \hline
500 & 6 & 0.5 & 0.06 & 0.08 & 0.05 &  0.11 & \bf 0.03 & 0.03 \\ 
  500 & 6 & 1 & 0.09 & 0.15 & 0.10 &  0.58 & \bf 0.08 & 0.07 \\ 
  500 & 6 & 2 & \bf 0.14 & 0.32 & 0.19 &  2.21 & 0.22 & 0.22 \\ 
  500 & 6 & 4 & \bf 0.31 & 0.65 & 0.39 &  8.10 & 0.78 & 0.76 \\ 
  500 & 12 & 0.5 & 0.07 & 0.09 & 0.06 &  0.10 & \bf 0.05 & 0.04 \\ 
  500 & 12 & 1 & 0.10 & 0.16 & 0.10 &  0.33 & \bf 0.09 & 0.07 \\ 
  500 & 12 & 2 & \bf 0.16 & 0.32 & 0.22 &  1.42 & 0.25 & 0.18 \\ 
  500 & 12 & 4 & \bf 0.30 & 0.61 & 0.41 &  5.09 & 0.61 & 0.68 \\ 
  1000 & 6 & 0.5 & 0.05 & 0.06 & 0.03 &  0.07 & \bf 0.02 & 0.02 \\ 
  1000 & 6 & 1 & 0.08 & 0.10 & 0.07 &  0.38 & \bf 0.04 & 0.05 \\ 
  1000 & 6 & 2 & \bf 0.11 & 0.20 & 0.13 &  1.80 & \bf 0.11 & 0.11 \\ 
  1000 & 6 & 4 & \bf 0.18 & 0.45 & 0.28 &  6.12 & 0.35 & 0.38 \\ 
  1000 & 12 & 0.5 & 0.06 & 0.07 & 0.05 &  0.09 & \bf 0.03 & 0.02 \\ 
  1000 & 12 & 1 & 0.08 & 0.10 & \bf 0.07 &  0.22 & \bf 0.07 & 0.05 \\ 
  1000 & 12 & 2 & \bf 0.13 & 0.23 & \bf 0.13 &  0.84 & 0.14 & 0.11 \\ 
  1000 & 12 & 4 & \bf 0.20 & 0.41 & 0.30 &  3.30 & 0.53 & 0.36 \\ 
   \hline
\end{tabular}
\caption{\tt Mean-squared error running \texttt{kernel ridge regression} from Setup A. Results are averaged across 200 runs, rounded to two decimal places, and reported on an independent test set of size $n$.} 
\label{table:setup1-kernel}
\end{table}

%% file: tables/simulation_results_setup_B_kernel.tex
\begin{table}[ht]
\centering
\begin{tabular}{ccccccccc}
  \hline
n & d & $\sigma$ & S & T & X & U & R & oracle \\ 
  \hline
500 & 6 & 0.5 & 0.09 & 0.10 & \bf 0.05 & \bf  0.05 & \bf 0.05 & 0.04 \\ 
  500 & 6 & 1 & 0.21 & 0.23 & 0.13 & \bf  0.10 & 0.11 & 0.09 \\ 
  500 & 6 & 2 & 0.46 & 0.57 & 0.36 &  0.36 & \bf 0.34 & 0.30 \\ 
  500 & 6 & 4 & \bf 0.89 & 1.33 & 0.94 &  1.21 & 1.15 & 1.12 \\ 
  500 & 12 & 0.5 & 0.21 & 0.22 & 0.10 & \bf  0.09 & \bf 0.09 & 0.05 \\ 
  500 & 12 & 1 & 0.34 & 0.37 & 0.21 & \bf  0.18 & \bf 0.18 & 0.14 \\ 
  500 & 12 & 2 & 0.58 & 0.66 & 0.55 & \bf  0.47 & 0.48 & 0.45 \\ 
  500 & 12 & 4 & 1.21 & 1.38 & \bf 1.11 &  1.30 & 1.24 & 1.24 \\ 
  1000 & 6 & 0.5 & 0.05 & 0.06 & 0.04 & \bf  0.03 & \bf 0.03 & 0.02 \\ 
  1000 & 6 & 1 & 0.12 & 0.13 & 0.08 & \bf  0.06 & \bf 0.06 & 0.05 \\ 
  1000 & 6 & 2 & 0.30 & 0.35 & 0.21 & \bf  0.17 & 0.18 & 0.16 \\ 
  1000 & 6 & 4 & 0.66 & 0.84 & \bf 0.54 &  0.63 & 0.59 & 0.55 \\ 
  1000 & 12 & 0.5 & 0.11 & 0.13 & 0.06 & \bf  0.05 & \bf 0.05 & 0.04 \\ 
  1000 & 12 & 1 & 0.24 & 0.26 & 0.13 & \bf  0.09 & \bf 0.09 & 0.08 \\ 
  1000 & 12 & 2 & 0.41 & 0.44 & 0.30 & \bf  0.26 & 0.27 & 0.23 \\ 
  1000 & 12 & 4 & 0.84 & 0.94 & \bf 0.71 &  0.78 & 0.83 & 0.79 \\ 
   \hline
\end{tabular}
\caption{\tt Mean-squared error running \texttt{kernel ridge regression} from Setup B. Results are averaged across 200 runs, rounded to two decimal places, and reported on an independent test set of size $n$.} 
\label{table:setup2-kernel}
\end{table}

%% file: tables/simulation_results_setup_C_kernel.tex
\begin{table}[ht]
\centering
\begin{tabular}{ccccccccc}
  \hline
n & d & $\sigma$ & S & T & X & U & R & oracle \\ 
  \hline
500 & 6 & 0.5 & 0.11 & 0.11 & 0.04 &  0.39 & \bf 0.02 & 0.01 \\ 
  500 & 6 & 1 & 0.32 & 0.34 & 0.13 &  0.61 & \bf 0.04 & 0.03 \\ 
  500 & 6 & 2 & 0.83 & 0.95 & 0.44 &  2.43 & \bf 0.14 & 0.15 \\ 
  500 & 6 & 4 & 1.62 & 2.23 & 1.13 & 12.53 & \bf 0.71 & 0.55 \\ 
  500 & 12 & 0.5 & 0.32 & 0.32 & 0.12 &  1.02 & \bf 0.02 & 0.01 \\ 
  500 & 12 & 1 & 0.69 & 0.72 & 0.32 &  2.82 & \bf 0.04 & 0.03 \\ 
  500 & 12 & 2 & 1.18 & 1.31 & 0.67 &  6.45 & \bf 0.34 & 0.15 \\ 
  500 & 12 & 4 & 2.10 & 2.71 & \bf 1.13 & 14.87 & 1.51 & 0.57 \\ 
  1000 & 6 & 0.5 & 0.05 & 0.05 & 0.02 &  0.33 & \bf 0.01 & 0.01 \\ 
  1000 & 6 & 1 & 0.17 & 0.17 & 0.06 &  0.40 & \bf 0.02 & 0.02 \\ 
  1000 & 6 & 2 & 0.48 & 0.55 & \bf 0.24 &  1.26 & 0.26 & 0.08 \\ 
  1000 & 6 & 4 & 1.04 & 1.42 & 0.69 &  5.37 & \bf 0.25 & 0.33 \\ 
  1000 & 12 & 0.5 & 0.15 & 0.15 & \bf 0.05 &  0.33 & 0.18 & 0.00 \\ 
  1000 & 12 & 1 & 0.41 & 0.43 & 0.17 &  1.10 & \bf 0.02 & 0.02 \\ 
  1000 & 12 & 2 & 0.85 & 0.93 & 0.44 &  5.16 & \bf 0.08 & 0.06 \\ 
  1000 & 12 & 4 & 1.47 & 1.91 & 0.90 & 11.93 & \bf 0.24 & 0.33 \\ 
   \hline
\end{tabular}
\caption{\tt Mean-squared error running \texttt{kernel ridge regression} from Setup C. Results are averaged across 200 runs, rounded to two decimal places, and reported on an independent test set of size $n$.} 
\label{table:setup3-kernel}
\end{table}

%% file: tables/simulation_results_setup_D_kernel.tex
\begin{table}[ht]
\centering
\begin{tabular}{ccccccccc}
  \hline
n & d & $\sigma$ & S & T & X & U & R & oracle \\ 
  \hline
500 & 6 & 0.5 & \bf 0.15 & \bf 0.15 & \bf 0.15 &  0.53 & 0.18 & 0.15 \\ 
  500 & 6 & 1 & \bf 0.32 & 0.34 & 0.33 &  0.96 & 0.38 & 0.35 \\ 
  500 & 6 & 2 & \bf 0.65 & 0.71 & 0.71 &  2.08 & 0.77 & 0.79 \\ 
  500 & 6 & 4 & \bf 1.27 & 1.75 & 1.45 &  5.26 & 1.81 & 1.68 \\ 
  500 & 12 & 0.5 & 0.31 & \bf 0.30 & 0.31 &  0.63 & 0.37 & 0.30 \\ 
  500 & 12 & 1 & \bf 0.53 & \bf 0.53 & 0.55 &  0.92 & 0.58 & 0.55 \\ 
  500 & 12 & 2 & \bf 0.80 & 0.95 & 0.92 &  1.68 & 0.96 & 0.99 \\ 
  500 & 12 & 4 & \bf 1.41 & 1.77 & 1.69 &  4.83 & 1.90 & 1.89 \\ 
  1000 & 6 & 0.5 & \bf 0.10 & \bf 0.10 & \bf 0.10 &  0.39 & 0.12 & 0.10 \\ 
  1000 & 6 & 1 & \bf 0.20 & \bf 0.20 & \bf 0.20 &  0.53 & 0.24 & 0.21 \\ 
  1000 & 6 & 2 & \bf 0.44 & 0.48 & 0.47 &  1.25 & 0.55 & 0.48 \\ 
  1000 & 6 & 4 & \bf 0.88 & 1.02 & 0.97 &  3.32 & 1.10 & 1.16 \\ 
  1000 & 12 & 0.5 & 0.20 & \bf 0.18 & 0.19 &  0.44 & 0.22 & 0.19 \\ 
  1000 & 12 & 1 & 0.38 & \bf 0.37 & 0.38 &  0.71 & 0.40 & 0.39 \\ 
  1000 & 12 & 2 & \bf 0.64 & 0.65 & 0.65 &  1.25 & 0.69 & 0.71 \\ 
  1000 & 12 & 4 & \bf 1.06 & 1.27 & 1.21 &  3.08 & 1.34 & 1.42 \\ 
   \hline
\end{tabular}
\caption{\tt Mean-squared error running \texttt{kernel ridge regression} from Setup D. Results are averaged across 200 runs, rounded to two decimal places, and reported on an independent test set of size $n$.} 
\label{table:setup4-kernel}
\end{table}

%% file: tables/simulation_results_setup_A_boost.tex
\begin{table}[ht]
\centering
\begin{tabular}{cccccccccc}
  \hline
n & d & $\sigma$ & S & T & X & U & CB & R & oracle \\ 
  \hline
500 & 6 & 0.5 & 0.06 & 0.10 & 0.04 & 0.05 & 0.04 & \bf 0.03 & 0.04 \\ 
  500 & 6 & 1 & 0.12 & 0.20 & 0.08 & 0.11 & 0.09 & \bf 0.06 & 0.06 \\ 
  500 & 6 & 2 & 0.26 & 0.44 & 0.16 & 0.20 & 0.21 & \bf 0.13 & 0.11 \\ 
  500 & 6 & 4 & 0.53 & 0.90 & \bf 0.32 & 1.04 & 0.33 & 0.35 & 0.32 \\ 
  500 & 12 & 0.5 & 0.07 & 0.11 & \bf 0.04 & 0.05 & 0.05 & \bf 0.04 & 0.04 \\ 
  500 & 12 & 1 & 0.13 & 0.23 & 0.08 & 0.12 & 0.10 & \bf 0.06 & 0.05 \\ 
  500 & 12 & 2 & 0.27 & 0.49 & 0.17 & 0.38 & 0.21 & \bf 0.13 & 0.11 \\ 
  500 & 12 & 4 & 0.48 & 0.88 & 0.34 & 1.21 & 0.34 & \bf 0.33 & 0.32 \\ 
  1000 & 6 & 0.5 & 0.05 & 0.07 & \bf 0.02 & 0.05 & 0.03 & \bf 0.02 & 0.03 \\ 
  1000 & 6 & 1 & 0.09 & 0.15 & \bf 0.05 & 0.07 & 0.06 & \bf 0.05 & 0.04 \\ 
  1000 & 6 & 2 & 0.20 & 0.36 & 0.11 & 0.20 & 0.16 & \bf 0.09 & 0.08 \\ 
  1000 & 6 & 4 & 0.38 & 0.68 & 0.23 & 0.50 & 0.27 & \bf 0.19 & 0.19 \\ 
  1000 & 12 & 0.5 & 0.05 & 0.08 & \bf 0.03 & 0.05 & \bf 0.03 & \bf 0.03 & 0.03 \\ 
  1000 & 12 & 1 & 0.09 & 0.16 & \bf 0.05 & 0.10 & 0.06 & \bf 0.05 & 0.05 \\ 
  1000 & 12 & 2 & 0.21 & 0.36 & 0.11 & 0.21 & 0.15 & \bf 0.08 & 0.08 \\ 
  1000 & 12 & 4 & 0.41 & 0.72 & 0.24 & 0.60 & 0.29 & \bf 0.22 & 0.24 \\ 
   \hline
\end{tabular}
\caption{\tt Mean-squared error running \texttt{boosting} from Setup A. Results are averaged across 200 runs, rounded to two decimal places, and reported on an independent test set of size $n$.} 
\label{table:setup1-boost}
\end{table}

%% file: tables/simulation_results_setup_B_boost.tex
\begin{table}[ht]
\centering
\begin{tabular}{cccccccccc}
  \hline
n & d & $\sigma$ & S & T & X & U & CB & R & oracle \\ 
  \hline
500 & 6 & 0.5 & 0.19 & 0.28 & \bf 0.14 & 0.20 & 0.28 & 0.20 & 0.14 \\ 
  500 & 6 & 1 & 0.33 & 0.48 & \bf 0.27 & 0.41 & 0.37 & 0.33 & 0.28 \\ 
  500 & 6 & 2 & 0.67 & 0.89 & \bf 0.56 & 0.84 & 0.67 & 0.68 & 0.62 \\ 
  500 & 6 & 4 & 1.40 & 1.76 & \bf 1.10 & 1.50 & 1.33 & 1.20 & 1.19 \\ 
  500 & 12 & 0.5 & 0.22 & 0.30 & \bf 0.15 & 0.22 & 0.35 & 0.21 & 0.15 \\ 
  500 & 12 & 1 & 0.37 & 0.50 & \bf 0.29 & 0.43 & 0.46 & 0.37 & 0.31 \\ 
  500 & 12 & 2 & 0.77 & 0.95 & \bf 0.58 & 0.89 & 0.79 & 0.74 & 0.68 \\ 
  500 & 12 & 4 & 1.63 & 1.87 & \bf 1.10 & 1.56 & 1.41 & 1.41 & 1.27 \\ 
  1000 & 6 & 0.5 & 0.13 & 0.19 & \bf 0.08 & 0.13 & 0.18 & 0.11 & 0.09 \\ 
  1000 & 6 & 1 & 0.21 & 0.33 & \bf 0.17 & 0.25 & 0.24 & 0.22 & 0.19 \\ 
  1000 & 6 & 2 & 0.45 & 0.65 & \bf 0.39 & 0.58 & 0.43 & 0.46 & 0.43 \\ 
  1000 & 6 & 4 & 1.01 & 1.34 & \bf 0.82 & 1.20 & 0.89 & 1.01 & 1.00 \\ 
  1000 & 12 & 0.5 & 0.14 & 0.21 & \bf 0.09 & 0.13 & 0.20 & 0.13 & 0.09 \\ 
  1000 & 12 & 1 & 0.25 & 0.34 & \bf 0.18 & 0.26 & 0.28 & 0.24 & 0.21 \\ 
  1000 & 12 & 2 & 0.50 & 0.69 & \bf 0.41 & 0.63 & 0.51 & 0.52 & 0.49 \\ 
  1000 & 12 & 4 & 1.16 & 1.33 & \bf 0.84 & 1.24 & 1.01 & 1.12 & 1.08 \\ 
   \hline
\end{tabular}
\caption{\tt Mean-squared error running \texttt{boosting} from Setup B. Results are averaged across 200 runs, rounded to two decimal places, and reported on an independent test set of size $n$.} 
\label{table:setup2-boost}
\end{table}

%% file: tables/simulation_results_setup_C_boost.tex
\begin{table}[ht]
\centering
\begin{tabular}{cccccccccc}
  \hline
n & d & $\sigma$ & S & T & X & U & CB & R & oracle \\ 
  \hline
500 & 6 & 0.5 & 0.30 & 0.65 & 0.13 & 0.97 & 0.65 & \bf 0.08 & 0.03 \\ 
  500 & 6 & 1 & 0.46 & 0.97 & 0.23 & 0.73 & 0.70 & \bf 0.15 & 0.08 \\ 
  500 & 6 & 2 & 0.90 & 1.73 & 0.44 & 0.86 & 0.82 & \bf 0.26 & 0.26 \\ 
  500 & 6 & 4 & 1.65 & 2.91 & 0.91 & 1.74 & 0.96 & \bf 0.57 & 0.43 \\ 
  500 & 12 & 0.5 & 0.32 & 0.68 & 0.15 & 0.90 & 0.69 & \bf 0.09 & 0.03 \\ 
  500 & 12 & 1 & 0.53 & 1.02 & 0.25 & 0.93 & 0.72 & \bf 0.17 & 0.10 \\ 
  500 & 12 & 2 & 0.98 & 1.83 & 0.47 & 0.95 & 0.84 & \bf 0.29 & 0.23 \\ 
  500 & 12 & 4 & 1.73 & 3.36 & 0.91 & 2.02 & 0.97 & \bf 0.53 & 0.56 \\ 
  1000 & 6 & 0.5 & 0.20 & 0.43 & 0.08 & 0.90 & 0.35 & \bf 0.05 & 0.02 \\ 
  1000 & 6 & 1 & 0.31 & 0.67 & 0.14 & 0.82 & 0.41 & \bf 0.11 & 0.07 \\ 
  1000 & 6 & 2 & 0.65 & 1.20 & 0.29 & 0.65 & 0.54 & \bf 0.20 & 0.19 \\ 
  1000 & 6 & 4 & 1.28 & 2.33 & 0.63 & 1.09 & 0.79 & \bf 0.42 & 0.38 \\ 
  1000 & 12 & 0.5 & 0.21 & 0.46 & 0.09 & 1.02 & 0.38 & \bf 0.06 & 0.03 \\ 
  1000 & 12 & 1 & 0.36 & 0.70 & 0.15 & 0.86 & 0.42 & \bf 0.12 & 0.07 \\ 
  1000 & 12 & 2 & 0.74 & 1.28 & 0.31 & 0.84 & 0.61 & \bf 0.23 & 0.20 \\ 
  1000 & 12 & 4 & 1.38 & 2.45 & 0.65 & 1.31 & 0.82 & \bf 0.40 & 0.37 \\ 
   \hline
\end{tabular}
\caption{\tt Mean-squared error running \texttt{boosting} from Setup C. Results are averaged across 200 runs, rounded to two decimal places, and reported on an independent test set of size $n$.} 
\label{table:setup3-boost}
\end{table}

%% file: tables/simulation_results_setup_D_boost.tex
\begin{table}[ht]
\centering
\begin{tabular}{cccccccccc}
  \hline
n & d & $\sigma$ & S & T & X & U & CB & R & oracle \\ 
  \hline
500 & 6 & 0.5 & 0.36 & \bf 0.30 & 0.37 & 0.57 & 0.50 & 0.43 & 0.39 \\ 
  500 & 6 & 1 & 0.55 & \bf 0.53 & 0.57 & 0.96 & 0.76 & 0.66 & 0.65 \\ 
  500 & 6 & 2 & \bf 0.92 & 0.99 & 1.02 & 1.60 & 1.21 & 1.12 & 1.13 \\ 
  500 & 6 & 4 & \bf 1.48 & 1.86 & 1.60 & 2.36 & 1.60 & 1.81 & 1.71 \\ 
  500 & 12 & 0.5 & 0.44 & \bf 0.34 & 0.43 & 0.63 & 0.55 & 0.48 & 0.43 \\ 
  500 & 12 & 1 & 0.65 & \bf 0.57 & 0.64 & 1.04 & 0.84 & 0.74 & 0.74 \\ 
  500 & 12 & 2 & \bf 1.05 & 1.06 & 1.10 & 1.66 & 1.35 & 1.24 & 1.26 \\ 
  500 & 12 & 4 & \bf 1.66 & 1.88 & 1.67 & 2.29 & 1.68 & 1.88 & 1.91 \\ 
  1000 & 6 & 0.5 & 0.24 & \bf 0.20 & 0.25 & 0.42 & 0.41 & 0.29 & 0.26 \\ 
  1000 & 6 & 1 & 0.39 & \bf 0.36 & 0.40 & 0.73 & 0.56 & 0.46 & 0.45 \\ 
  1000 & 6 & 2 & \bf 0.68 & 0.71 & 0.73 & 1.33 & 0.94 & 0.81 & 0.83 \\ 
  1000 & 6 & 4 & \bf 1.23 & 1.45 & 1.34 & 1.98 & 1.41 & 1.44 & 1.51 \\ 
  1000 & 12 & 0.5 & 0.29 & \bf 0.22 & 0.28 & 0.47 & 0.41 & 0.32 & 0.30 \\ 
  1000 & 12 & 1 & 0.45 & \bf 0.38 & 0.45 & 0.78 & 0.62 & 0.52 & 0.51 \\ 
  1000 & 12 & 2 & 0.80 & \bf 0.77 & 0.83 & 1.46 & 1.08 & 0.94 & 0.93 \\ 
  1000 & 12 & 4 & \bf 1.38 & 1.53 & 1.43 & 1.99 & 1.53 & 1.65 & 1.62 \\ 
   \hline
\end{tabular}
\caption{\tt Mean-squared error running \texttt{boosting} from Setup D. Results are averaged across 200 runs, rounded to two decimal places, and reported on an independent test set of size $n$.} 
\label{table:setup4-boost}
\end{table}